%% file: Thesis.tex
\documentclass{article}

\usepackage{arxiv}

\usepackage[utf8]{inputenc}
\usepackage[T1]{fontenc}
\usepackage{array}
\usepackage{amsmath}
\usepackage{amsfonts}
\usepackage{amsthm}

\usepackage{graphicx}
\usepackage{subcaption}
\usepackage{url}
\usepackage[ruled,vlined]{algorithm2e} 
\usepackage{CJKutf8}
\usepackage{bm} 
\usepackage[font=small]{caption}
\usepackage{cite}
\usepackage{subfiles} 

\DeclareMathOperator*{\argmax}{arg\,max}
\DeclareMathOperator*{\argmin}{arg\,min}
\DeclareMathOperator{\Tr}{Tr} 

\newtheorem{thm}{Theorem}
\newtheorem{proposition}[thm]{Proposition}

\title{Practical calibration of the temperature parameter in Gibbs posteriors}

\author{\textbf{Lucie Perrotta}\\
Information Theory Laboratory, EPFL, Lausanne, Switzerland\thanks{Project realized under the supervision of Prof. Emre Telarar}\\
Center for Advanced Intelligence Project (AIP), RIKEN, Tokyo, Japan\thanks{Project realized under the supervision of Dr. Pierre Alquier and Dr. Emtiyaz Khan} \\
\texttt{lucie.perrotta@riken.jp} }

\date{\today}

\begin{document}
\maketitle

\begin{abstract}
    PAC-Bayesian algorithms and Gibbs posteriors are gaining popularity due to their robustness against model misspecification even when Bayesian inference is inconsistent. The PAC-Bayesian $\alpha$-posterior is a generalization of the standard Bayes posterior which can be tempered with a parameter $\alpha$ to handle inconsistency. Data driven methods for tuning $\alpha$ have been proposed but are still few, and are often computationally heavy. Additionally, the adequacy of these methods in cases where we use variational approximations instead of exact $\alpha$-posteriors is not clear. This narrows their usage to simple models and prevents their application to large-scale problems. We hence need fast methods to tune $\alpha$ that work with both exact and variational $\alpha$-posteriors.
    
    First, we propose two data driven methods for tuning $\alpha$, based on \textit{sample splitting} and \textit{bootstrapping} respectively. Second, we formulate the (exact or variational) posteriors of three popular statistical models, and modify them into $\alpha$-posteriors. For each model, we test our strategies and compare them with standard Bayes and Grünwald's \textit{SafeBayes} \cite{Grnwald2012TheSB}. While bootstrapping achieves mixed results, sample splitting and SafeBayes perform well on the exact and variational $\alpha$-posteriors we describe, and achieve better results than standard Bayes in misspecified or complex models. Additionally, sample splitting outperforms SafeBayes in terms of speed. 
    
    Sample splitting offers a fast and easy solution to inconsistency and typically performs similarly or better than Bayesian inference. Our results provide hints on the calibration of $\alpha$ in PAC-Bayesian and Gibbs posteriors, and may facilitate using these methods in large and complex models.
\end{abstract}

\subfile{chapters/introduction}

\subfile{chapters/strategies}

\subfile{chapters/results}

\subfile{chapters/conclusion}

\bibliographystyle{IEEEtranSA}
\bibliography{references}

\subfile{chapters/appendix}

\end{document}

%% file: chapters/introduction.tex
\section{Introduction}
\subsection{Definition of the problem}
Statistics aim to predict the distribution of some observations, and to predict future values. A particular approach to this goal, \textit{Bayesian statistics}, uses the observations and a prior on the parameters of their distribution to build a \textit{posterior distribution} that represents the information on the parameters. Although performing well in general, it has been shown that Bayesian estimation can be inconsistent in some frameworks, especially when the model is complex \cite{barron1999consistency} or misspecified \cite{Grnwald2012TheSB}. A model is here said complex when it has many parameters. A generalization of Bayes estimations have shown to possibly perform better in these setups, by tempering the likelihood with a tunable parameter $\alpha$. Although the theory has been considerably reviewed, practical implementations still are few. We propose new and review existing methods for calibrating $\alpha$, and compare their performances with the Bayesian framework on three statistical models.

\subsubsection{Non-convergence of the Bayesian posterior}
Most problems in supervised learning and statistics can be seen as the estimation of a parameter $\bm{\theta}$, where the bold notation represents vectors and matrices. In Bayesian statistics, and more generally in the PAC-Bayesian framework \cite{McAllester1999, catoni2004, Alquier2008, zhang2006, jiang2008}, the parameters are represented with random variables. The prior information on the parameters is encoded by a probability distribution $\pi_0$, that we simply call the \textit{prior}. For observations $\bm{X} = \{X_1, ..., X_n \}$ i.i.d. in $\mathcal{X}$ from some unknown distribution $P$, a parameter space $\Theta$ of dimension $d$, a temperature parameter $\alpha$, and a risk function $r_n(\bm{\theta})$, the Gibbs posterior is written as
\begin{align}\label{pac_bayes_normal}
    \underbrace{\pi_P(  d \bm{\theta})}_{\text{Gibbs posterior}} \propto \underbrace{  \exp\left[- \alpha \cdot r_n(\bm{\theta}) \right] }_{\text{empirical error}} \cdot \underbrace{  \pi_0(d\bm{\theta}) }_{\text{prior}}.
\end{align}
Note that the risk function $r_n(\bm{\theta})$ is empirical because it depends on the observations $\bm{X}$, although the dependency is here omitted for readability. Alternatively, the computation of the Gibbs posterior can be seen as a minimization of the objective function, itself proportional to the negative evidence lower bound (negative ELBO), over all probability distributions $\mathcal{S}(\Theta)$ (see \cite{catoni2004, Alquier2015OnTP} for examples),
\begin{align}
    \pi_P(  d \bm{\theta}) &\propto \argmin_{\rho \in \mathcal{S}(\Theta)} \Big\{ \underbrace{ \mathcal{KL} (\rho || \pi_P )  }_{\text{objective}} \Big\} \label{bayes_objective} \\
    &= \argmin_{\rho \in \mathcal{S}(\Theta)} \Big\{ \underbrace{ \alpha \cdot \mathbb{E}_{\bm{\theta} \sim \rho} \left[ r_n (\bm{\theta} ) \right] + \mathcal{KL}(\rho || \pi_0) }_{\text{negative ELBO}} \Big\},  \label{bayes_kl}
\end{align}
where $\mathcal{KL}$ denotes the KL (Kullback-Leibler) divergence. The objective function in \eqref{bayes_objective} can be replaced by the negative ELBO in \eqref{bayes_kl} since they only differ by a term proportional to the evidence and independent of $\rho$. When the Gibbs posterior is intractable, many methods are available, and we here use the variational approximation 
\begin{align*}
    \Tilde{\pi}_P(  d \bm{\theta}) \propto \argmin_{\rho \in \mathcal{F}} \Big\{ \alpha \cdot \mathbb{E}_{\bm{\theta} \sim \rho} \left[ r_n (\bm{\theta} ) \right] + \mathcal{KL}(\rho || \pi_0) \Big\}
\end{align*}
where $\mathcal{F}$ is a family of probability distribution that we consider tractable \cite{alquier2017concentration}.\\

The most common use of the formula \eqref{pac_bayes_normal} is a special case in which we define a loss function $\ell:\bm{\theta} \times \mathcal{\bm{X}} \mapsto \mathbb{R}_+$ and the \textit{generalized error} function
\begin{align*}
    R (\bm{\theta} ) = \mathbb{ E}_{X \sim P} \left[ \ell (\bm{\theta}, X ) \right].
\end{align*}
Using the observed data only, the generalized error is estimated by the \textit{empirical error} function as
\begin{align}\label{empirical-risk}
     r_n(\bm{\theta}) =\frac{1}{n}\sum_{i=1}^n \ell(\bm{\theta},X_i),
\end{align}
and can be plugged back into equation \eqref{pac_bayes_normal} to compute the prevision error. In the two equations above, $X$ is one observation vector from $P$, and $X_i$ are columns vectors of the design matrix $\bm{X}$. The vectors are here not written in bold notation which is kept for the design matrix $\bm{X}$ only. A motivation for using the empirical error is that 
\begin{align*}
    \mathbb{E}_{\bm{X} \sim P} \left[ r_n( \bm{\theta} ) \right] = R(\bm{\theta}) .
\end{align*}

Let $(p_{\bm{\theta}})_{\bm{\theta} \in \Theta}$ be a parametric family of probability distribution functions. By choosing the risk as in \eqref{empirical-risk} with $\ell (\bm{\theta}, X_i) = - \log p_{\bm{\theta}}(X_i)$ and subsequently fixing $\alpha=n$, we get a special case of the Gibbs posterior that is the usual Bayesian  \textit{posterior}, and obtain the memorable form of the Bayes' theorem \cite{germain2016pacbayesian}
\begin{align}
    \underbrace{\pi_P(  d \bm{\theta})}_{\text{posterior}} &\propto \prod_{i=1}^n p_{\bm{\theta}}(X_i) \cdot \pi_0 ( d \bm{\theta} ) = \underbrace{  \mathcal{L}(\bm{\theta}, \bm{X}) }_{\text{likelihood}} \cdot \underbrace{  \pi_0(d\bm{\theta}) }_{\text{prior}} . \label{bayes_normal}
\end{align}

It is often taken for granted that the above Bayesian posterior converges in some sense to the delta function centered on the optimal choice $\bm{\theta}^* := \min_{\bm{\theta}} R(\bm{\theta}) $ when $n \rightarrow \infty$. However, \cite{barron1999consistency} has shown that this is not true in general. Some assumptions are necessary to be able to prove the consistency of Bayesian inference, these conditions are stated in \cite{van2000asymptotic}. More specifically, the Bayesian posterior sometimes leads to poor estimations when the model is misspecified with the data, as Bayes does not generally tends to focus on posterior distributions whose KL-divergence to $P$ are minimal \cite{grnwald2014inconsistency, Ramamoorthi_2015, masegosa2019learning}. Even when the usual Bayes approach is consistent, it might be that changing the value of $\alpha$ might improve things. It is hence a focus to explore alternatives to the posterior \eqref{bayes_normal}.

\subsubsection{The $\alpha$-posterior}
In recent years, a different special case of the Gibbs posterior distribution has been discussed \cite{zhang2006, grun999, guedj2019primer}, in which the risk function is taken as in \eqref{empirical-risk} and where the temperature parameter $\alpha$ is this time kept as a tunable parameter. This corresponds to adding an exponent term $\alpha/n$ to the likelihood in \eqref{bayes_normal}. We call this new term the \textit{tempered likelihood}. This allows the computations to adjust the relative weight given to the prior and the likelihood. The Gibbs posterior from \eqref{pac_bayes_normal} and \eqref{bayes_kl} can be rewritten as
\begin{align}
     \underbrace{\pi_\alpha ( d\bm{\theta})}_{\text{$\alpha$-posterior}} &\propto \underbrace{  \mathcal{L}(\bm{\theta}, \bm{X})^{\alpha / n} }_{\text{tempered likelihood}} \cdot \underbrace{  \pi_0(  d \bm{\theta}) }_{\text{prior}}. \label{a-generalized-posterior}
\end{align}
The value $\pi_\alpha (  d \bm{\theta})$ is often denominated as the \textit{$\alpha$-generalized posterior} or simply \textit{$\alpha$-posterior} \cite{grnwald2014inconsistency,jiang2008}, and the parameters of the $\alpha$-posterior are denoted $\Omega_P(\alpha)$ and are themselves functions of $\alpha$. In that sense, the Bayesian posterior \eqref{bayes_normal} is a special case of the $\alpha$-posterior \eqref{a-generalized-posterior} where $\alpha = n$, and the $\alpha$-posterior is itself a special case of the Gibbs posterior \eqref{pac_bayes_normal} where $r_n(\bm{\theta})$ is as in \eqref{empirical-risk}. \cite{grnwald2014inconsistency, grnwald2016safe} show that there exists a scalar $\beta_\text{max} > 0$ such that $\forall \beta \in (0, \beta_\text{max} )$, we have $\pi_{n \beta} ( d\bm{\theta}) {\rightarrow}_{n \rightarrow \infty} \delta ( \bm{\theta}^* ) $. In other words, when the Bayesian posterior does not converge, taking $\alpha = n \beta$ for some $\beta$ small enough will fix things. \\

Additionally, even when the Bayesian posterior does converge to the delta function (when $\beta_\text{max} > 1$), it might still be that taking $\alpha \neq n$ will improve performance. Values of $\alpha \neq n$ may produce an $\alpha$-posterior that converges faster to $P$ than the Bayesian posterior. In this work, we analyse existing and propose new methods that we call \textit{strategies}, aiming to find values of $\alpha$ achieving this goal.\\

\begin{figure}[ht]
    \centering
        \includegraphics[width=.9\textwidth]{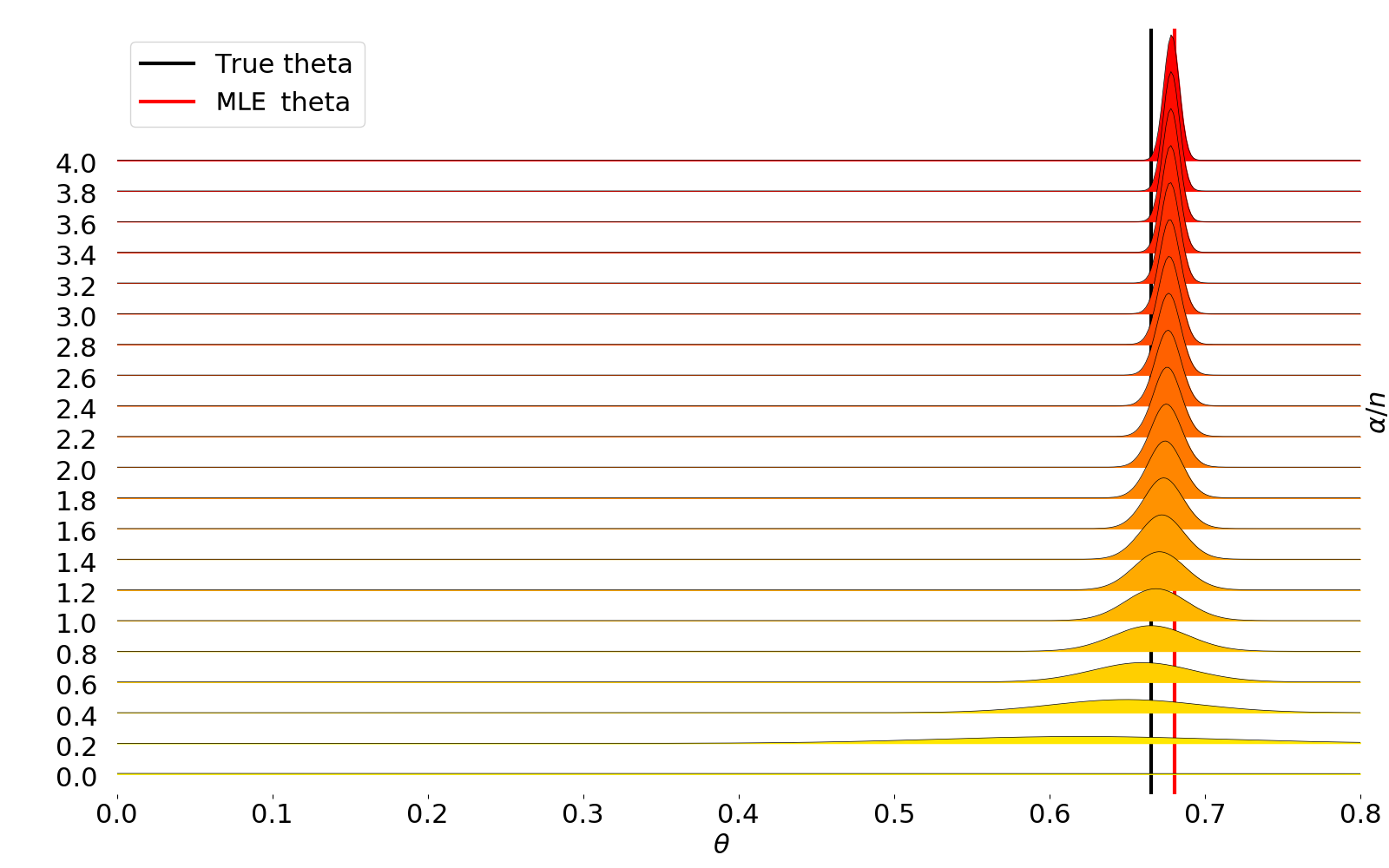}
    \caption{Variation of the $\alpha$-posterior as $\alpha/n$ grows larger, where $\pi_0 \sim \mathcal{N}(0, 1)$, $d=1$ and $n=50$ of observations is available. The Bayesian posterior is the one shown on line $\alpha/n = 1 $.}
    \label{fig:joyplot_linreg1}
\end{figure}

A comprehensive visualization of the influence of $\alpha$ on the $\alpha$-posterior is shown in figure \ref{fig:joyplot_linreg1}. Bell curves representing Gaussian distributions are plotted for increasing values of $\alpha/n$ on the vertical axis, in a so called \textit{joyplot} fashion \cite{jpoyplot}. When $\alpha$ is small, the $\alpha$-posterior is mostly computed from the prior, and hence resembles a flat, poorly informative standard Gaussian distribution. As $\alpha$ gets larger, the observations gain more influence on the $\alpha$-posterior, which starts to concentrate around the maximum likelihood estimate (MLE) of the parameter $\bm{\theta}$. When $\alpha \rightarrow \infty$, the $\alpha$-posterior becomes a zero variance delta function centered on the MLE. More details can be found for each model in the appendices. For a given model $\mathcal{M}$ and observations $\bm{X}$, one can write the bias-variance tradeoff minimization \cite{geman92} as
\begin{align*}
    \argmin_\alpha R(\bm{\theta}) = \argmin_\alpha \left\{ \text{Bias}_{\bm{X}} \left[ \mathcal{M}(\alpha) \right]^2 + \text{Var}_{\bm{X}} \left[ \mathcal{M}(\alpha) \right] \right\} .
\end{align*}
The bias of an $\alpha$-posterior is here the absolute difference between the true $\bm{\theta}$ and the mean of the $\alpha$-posterior. One remarks that when $n$ is small, the MLE may be a poor estimate of the true $\bm{\theta}$, and have a large bias. Small values of $\alpha$, such as in this case $\alpha/n \approx 0.8$, can achieve a zero bias, but instead display a larger variance. The optimal bias-variance tradeoff hence lies between these two values, but is not necessarily equal to the Bayesian choice.

\subsection{Methodology} 
In some models, the $\alpha$-posterior is available explicitly. When this is not the case, several sampling techniques have been proposed \cite{guedj2019primer}, including MCMC methods \cite{lietal}. Other methods include stochastic optimization \cite{chriefabdellatif2019generalization, NIPS2017_6886}, or variational approximation \cite{Alquier2015OnTP}. In this work, we consider models where the $\alpha$-posterior is available explicitly and other models where it is not, in which case we use a variational approximation. We then apply our algorithms on each model regardless of if a variational approximation was used in the $\alpha$-posterior or not.

\subsubsection{Optimizing over $\alpha$}
Before describing the strategies for choosing $\alpha$, we need a measure of the quality of a given $\alpha$ for one model. We hence define the \textit{expected generalization error}, or just \textit{generalization error}, as
\begin{align} \label{generalization_error}
    \mathcal{R} (\alpha) &:= \mathbb{E}_{\bm{\theta} \sim \pi_\alpha} \left[ R(\bm{\theta}) \right]
\end{align}
and our objective is
\begin{align}\label{objective_cost}
    \alpha^* = \min_\alpha \mathcal{R}(\alpha) .
\end{align}
This value is typically non available in practice, as the generalized error $R$ requires to know the distribution $P$. This value is a lower bound to what strategies can achieve in the best case, as it computes the risk of the unknown optimal $\alpha^*$s. The generalization error is lower bounded by the \textit{minimal prediction error}
\begin{align*}
    \min_{\bm{\theta}} R(\bm{\theta})
\end{align*}
which corresponds to the risk computed with smallest possible prediction risk achievable.\\

A naive proposal to compute $\alpha^*$ would have been to alternatively maximize \eqref{bayes_kl} over $\pi_\alpha$ and $\alpha$ as 
\begin{align*}
      \argmax_\alpha \max_{\pi_\alpha} \underbrace{\mathbb{E}_{\bm{\theta} \sim \pi_\alpha} [ - \alpha  r_n (\bm{\theta}) ] - \mathcal{KL} ( \pi_\alpha (\bm{\theta}) || \pi_0(\bm{\theta}) ) }_\text{ELBO} .
\end{align*}
However, such a method would lead to choosing $\alpha = 0$ and $\pi_\alpha = \pi_0$ as the optimal solution each time. Instead, to find the theoretical optimal value of $\alpha$, we run a two-step procedure 
\begin{align}
\begin{cases}  \label{alternating_opti}
    \text{$\max_{\pi_\alpha} \mathrm{ELBO}$ with fixed $\alpha$} \\
    \text{use a strategy to choose $\alpha$}.
\end{cases}
\end{align}
Some strategies to choose $\alpha$ discussed below.

\subsubsection{The overfitting problem}
\begin{figure}[h]
    \centering
    \includegraphics[width=.8\textwidth]{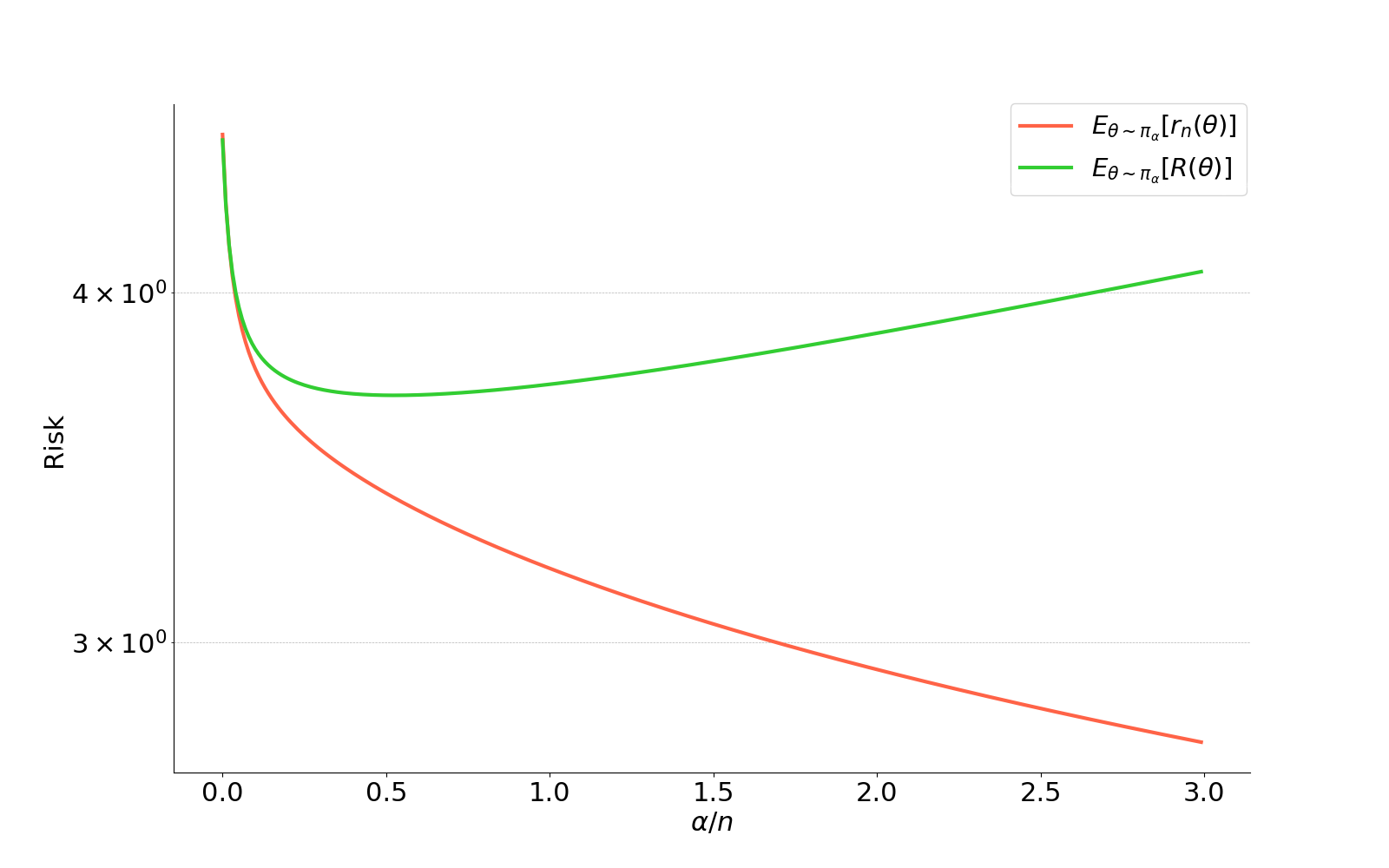}
    \caption{Generalization error (in green), and empirical error (in red). Overfitting here occurs with a model trained on a small dataset of size $n=40$. The Bayesian choice of $\alpha/n=1$ is close yet not equal to the minimum of the generalization error.}
    \label{fig:overfit_example}
\end{figure}

When approximating the generalization error with the empirical error, the size $n$ of the observations dataset is typically small, we must hence be careful with overfitting. Computing the empirical error and the $\alpha$-posterior over the same data will likely lead to overfitting. In that situation, minimizing the empirical error will boil down to choosing $\alpha$ as large as possible, since the empirical error is computed on the same dataset as the $\alpha$-posterior. To illustrate that phenomenon, we plot the curves of the generalization error and the empirical error for all $\alpha$s between 0 and some maximum value. In simple models where overfitting hardly occurs, both curves will typically look similar and decreases as $\alpha$ increases. But as the model becomes more complex, the optimal $\alpha$ will rather lay close to the value chosen by the Bayesian posterior, and the curve will increase on both directions around the minimum. One can observe that effect in figure \ref{fig:overfit_example}. Alternatively, by "training" the $\alpha$-posterior over some data and "testing" the empirical error over some different data, a good strategy can avoid overfitting and find a non-trivial minimal $\alpha$. More details on overfitting can be found in \cite{generalization_overfit}.

In summary, we define a good strategy as being able to
\begin{itemize}
    \item find a value of $\alpha$ associated to a risk as close as possible to the generalization error's minimum $\mathcal{R}(\alpha^*)$,
    \item perform well on exact and variational posteriors.
\end{itemize}
We now summarize existing and new strategies for computing such an $\alpha$.

\subsubsection{Existing strategies}
We use three existing strategies in our comparison: standard Bayes, the naive strategy minimizing the empirical error, as well as the recent SafeBayes strategy which is robust against misspecification.

\paragraph{Bayes} As described above, the standard Bayesian "strategy" chooses $\alpha = n$, regardless of the model nor the data. It gives the same weight to the likelihood and to the prior. 

\paragraph{Naive} The naive strategy uses all the observed data for computing both the $\alpha$-posterior \eqref{a-generalized-posterior} and the empirical error \eqref{empirical-risk}. We minimize
\begin{align*}
    \mathbb{E}_{\bm{\theta} \sim \pi_\alpha} \left[ r_n (\bm{\theta}) \right] .
\end{align*}
This corresponds to minimizing the red curve in figure \ref{fig:overfit_example}. Since the posterior exactly matches the data, this strategy is very confident and always chooses the maximum authorized value for $\alpha$, giving the maximum weight to the likelihood. In simple models, this results are accurate, as typically no overfitting occurs. However, in more complex models, the generalization error may show a clear minimum over $\alpha$ and taking the larger authorized value is no longer viable.

\paragraph{SafeBayes} This strategy was proposed in \cite{Grnwald2012TheSB} as a possible solution to learning from a misspecified model, with which the traditional Bayes strategy typically behaves poorly. 

\subsection{Our contribution}
We propose two strategies to compute $\alpha$ and compare them with the existing strategies presented above. In models where the generalization error and its estimates are tractable, they can be minimized explicitly. When it is not tractable, a gradient descent algorithm is used, in which case we use the closed-form gradient
\begin{align}\label{gradient_introduction}
\frac{\partial }{\partial \alpha} \mathbb{E}_{\bm{\theta}\sim \pi_\alpha^{(\lambda)}} \left[ r_n^{(\nu)}(\bm{\theta}) \right] = -\text{Cov}_{\bm{\theta}\sim \pi_\alpha^{(\lambda)}} \left[ r_n^{(\lambda)}(\bm{\theta}),r_n^{(\nu)}(\bm{\theta}) \right] ,
\end{align}
where the $\alpha$-posterior is computed with a dataset $(\lambda)$ and the empirical error with a dataset $(\nu)$. A proof can be found in the second chapter of this work, along with more details about all strategies.

\subsubsection{Proposed strategies}

\paragraph{Sample splitting} The sample splitting is a two-fold strategy that tries to tackle the overfitting issue by training the $\alpha$-posterior \eqref{a-generalized-posterior} over the first half $(1)$ of the observations only, and to compute the empirical error \eqref{empirical-risk} over the second half $(2)$. We hence minimize the generalization error estimate
\begin{align*}
    \mathbb{E}_{\bm{\theta} \sim \pi_\alpha^{(1)}} \left[ r_n^{(2)} (\bm{\theta}) \right] .
\end{align*}
The strategy is hence expected not to overfit in general and to choose a smaller $\alpha$ than the naive strategy.

\paragraph{Bootstrapping} The second strategy proposed uses the bootstrap theory, whereby a new dataset $\bm{\Tilde{X}}$ is created by uniformly drawing $n$ values from $\bm{X}$ with replacement. We then first compute the $\alpha$-posterior with $\bm{\Tilde{X}}$, and second the empirical error simply with the data $\bm{X}$:
\begin{align*}
    \mathbb{E}_{\bm{\theta} \sim \pi_{\Tilde{\alpha}}} \left[ r_n (\bm{\theta}) \right] , \text{ where } 
    \pi_{\Tilde{\alpha}} ( d\bm{\theta}) &\propto \mathcal{L}(\bm{\theta}, \bm{\Tilde{X}})^{\alpha / n} \cdot   \pi_0(  d \bm{\theta}).
\end{align*}
This generalization error estimate is computed many times with different random draws from the observations, that we average together in order to reduce the noise before minimizing.\\

As said before, all strategies we analyze in this paper estimate the generalization error $\mathcal{R}(\alpha)$ from the observed data, and minimize the estimates.

\subsubsection{Evaluating the strategies on statistical models}
We test the strategies over three statistical models: \textit{the linear regression with known} and \textit{unknown variance}, and \textit{the logistic regression.} For each model, we derive and specify a closed-form formula for the $\alpha$-posterior (exact or variational) and of its parameters, as functions of $\alpha$. We subsequently compute a formula for the generalization error (closed-form or approximated) that we adapt into the strategies in order to obtain formulae for the estimates of the generalization error, as described in the next chapter. We then run the strategies on the models using the computed $\alpha$-posterior and generalization error estimate, and compare their risks.\\

In the linear regression with known and unknown variance, we aim to compare the behaviour of the strategies when different types of misspecification exist between the data and the model. In the logistic regression, we do not have an explicit $\alpha$-posterior available, so we use variational approximations. In this case, there is no reason for \eqref{gradient_introduction} to hold. However, we still use our strategies relying on \eqref{gradient_introduction} and check how they perform. We compare two variational approximations of the $\alpha$-posterior.

\subsubsection{Summary of the results}

\begin{figure}[ht]
\centering
\begin{minipage}{\textwidth}
\makebox[\textwidth]{
\begin{minipage}{.55\textwidth}
  \centering
  \includegraphics[width=\linewidth]{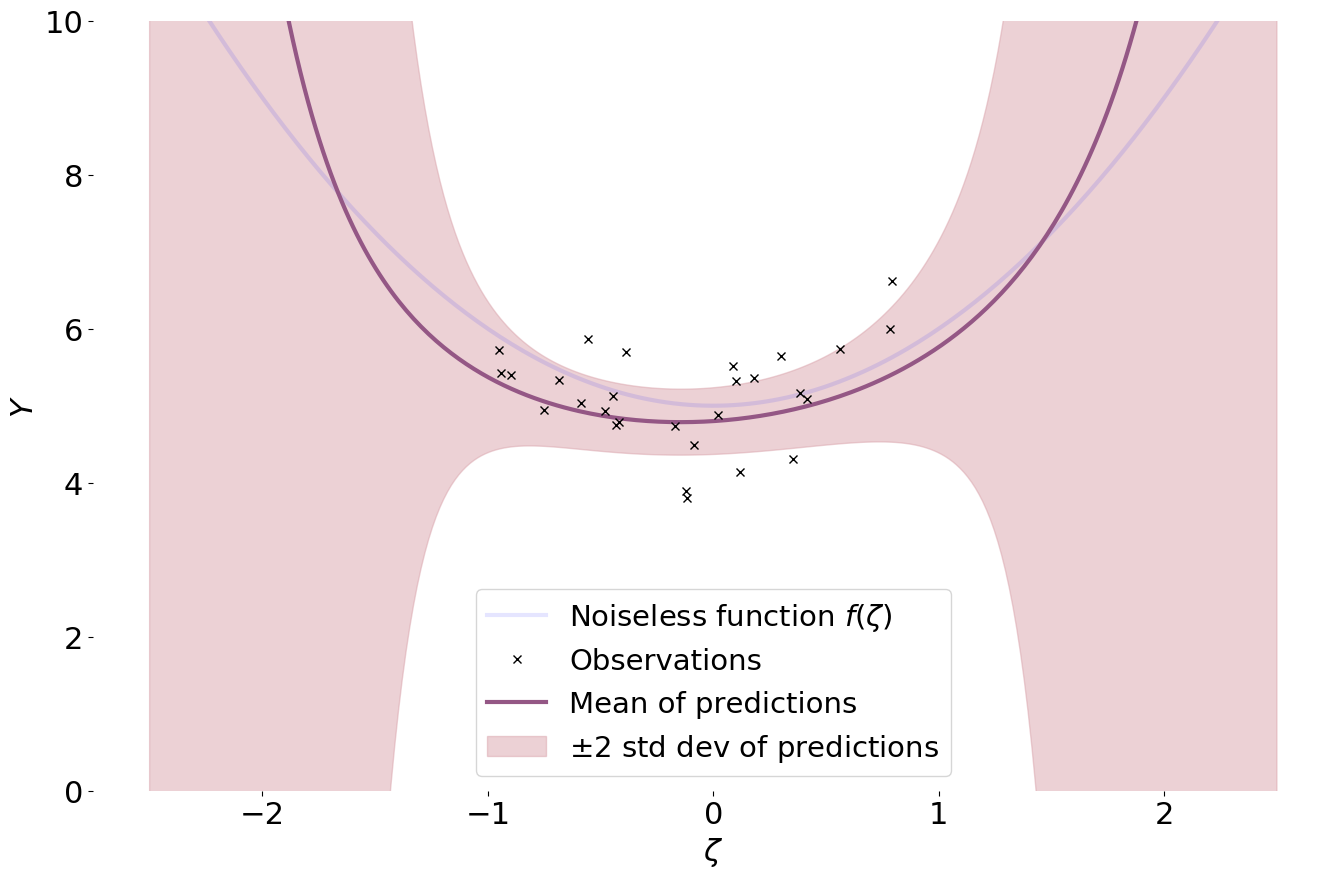}
\end{minipage}%
\begin{minipage}{.55\textwidth}
  \centering
  \includegraphics[width=\linewidth]{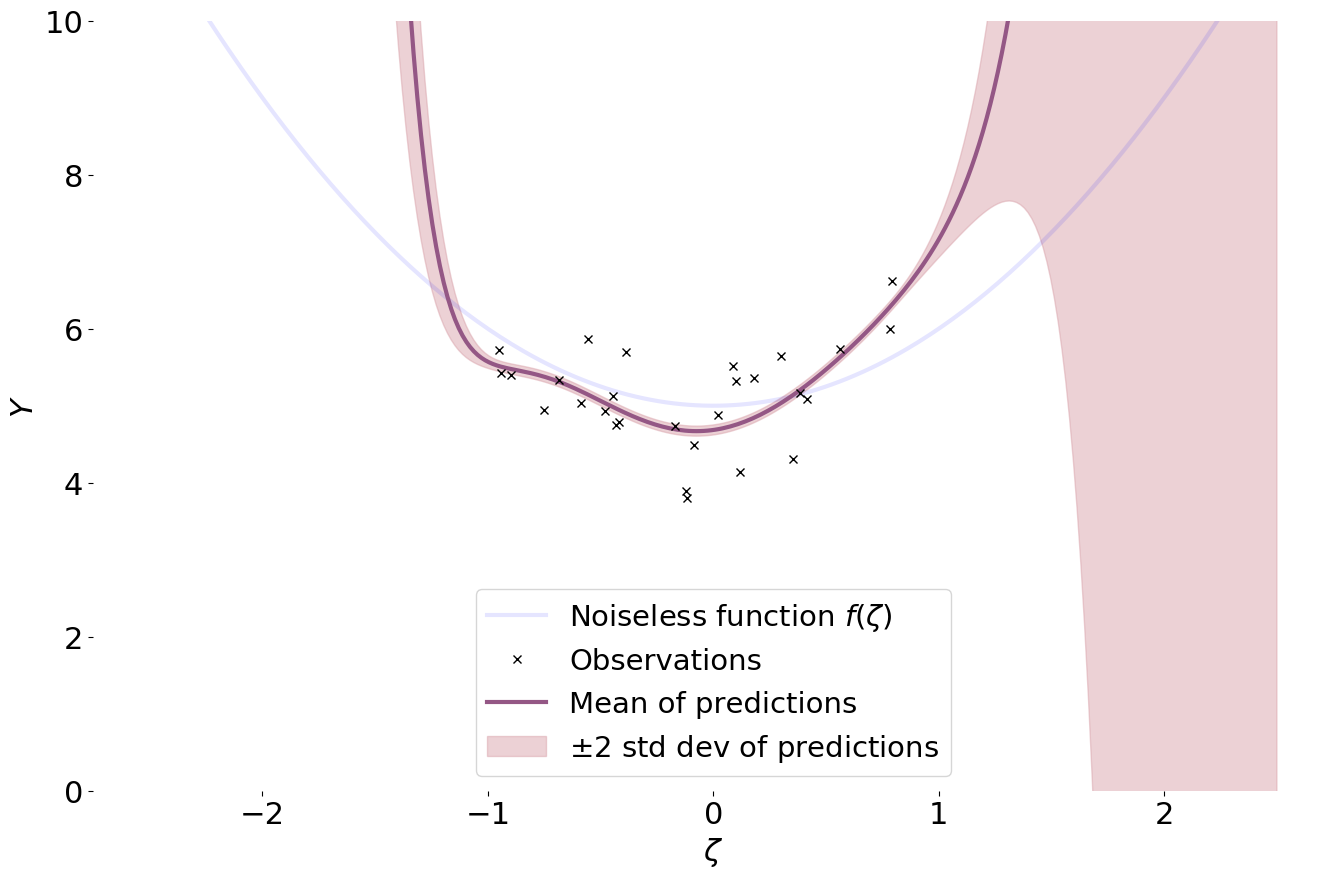}
\end{minipage}
}
\caption{Sample splitting's (left) and Bayes' (right) predictive polynomial curves of a function $f$.}
\label{fig:polyreg_intro}
\end{minipage}

\makebox[\textwidth]{
\begin{minipage}{.7\textwidth}
  \centering
  \includegraphics[width=\linewidth]{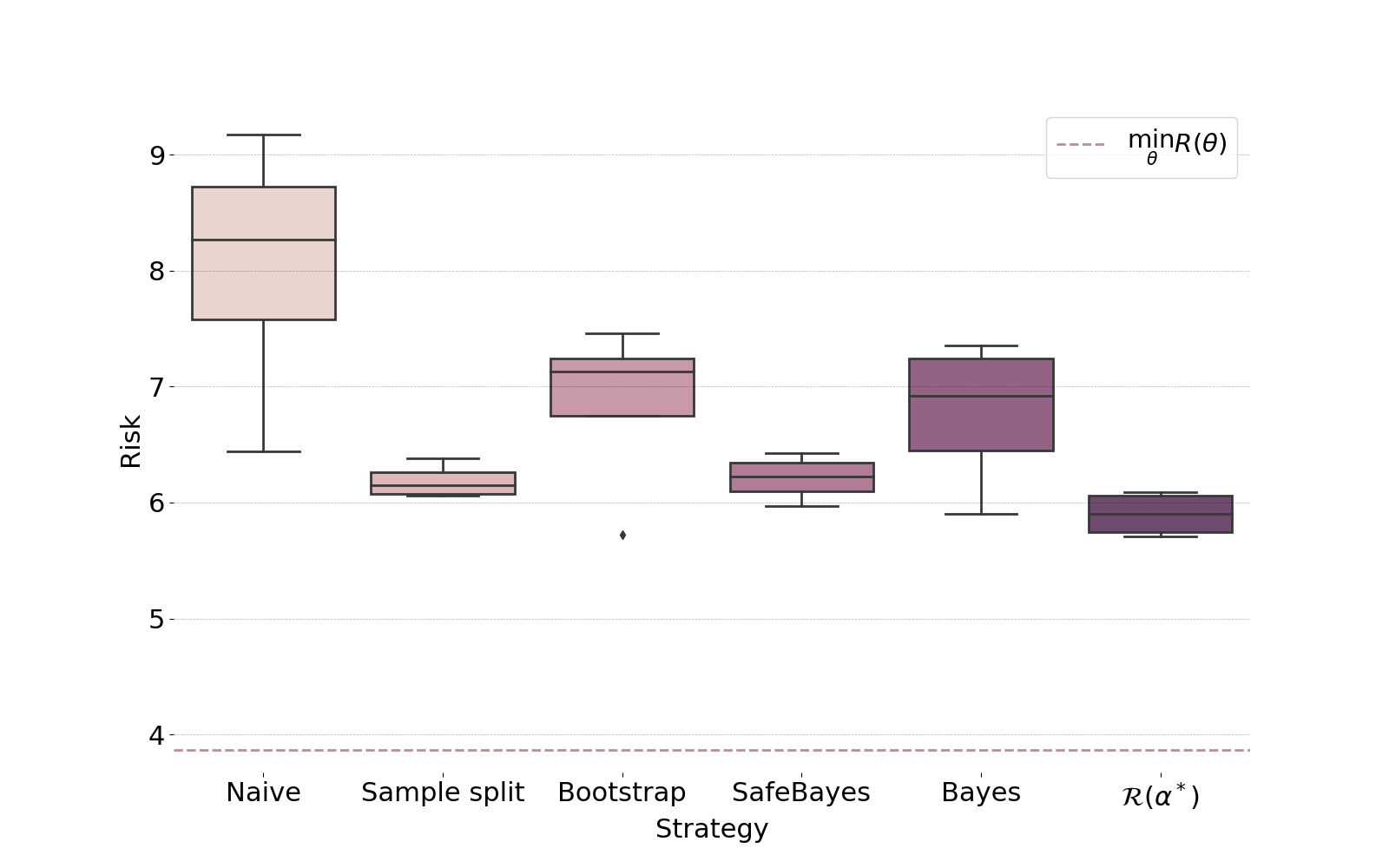}
\caption{Comparison of the achieved risks for each strategy in the linear regression model.}
\label{fig:boxplots_example}
\end{minipage}%
}
\end{figure}

Most of the work done in this thesis was to practically implement the strategies and the statistical models. A general optimization scheme was written in Python to find the best $\alpha$ with each strategy for an arbitrary model. Many simulations were then run on each of the three models with different datasets. Then, box-plots and error curves were created to compare the performances of each strategy. We observe the following main results:\\

\textbf{In complex or misspecified models, the sample splitting and SafeBayes strategies perform the best}. Both strategies usually are similarly successful at estimating $\alpha^*$. The sample splitting is faster to compute than SafeBayes. The bootstrap strategy gives mixed results, performing better when a large number of observations are available. Bayes typically performs worse than sample splitting and SafeBayes. In such models, either $\beta_{\max} < 1$ and Bayes will not converge, either sample splitting and SafeBayes are faster in convergence than Bayes for the given $n$. The naive strategy always chooses the maximal $\alpha$ and performs worse than all other strategies.

\textbf{In simple well-specified models, all strategies perform similarly.} Since overfitting hardly occurs, all strategies choose an $\alpha/n$ close to the maximum allowed value, expect Bayes which takes $\alpha/n=1$. Bayes hence scores almost imperceptibly worse than the other strategies, as the generalization error is very flat and decreases slowly as $\alpha/n$ increases. The naive strategy is here performing well.

\textbf{The strategies are empirically successful on sufficiently accurate logistic regression variational $\alpha$-posteriors.} Although the proposed closed-form gradient formula briefly described above has been proven to work on exact $\alpha$-posteriors only, the strategies perform similarly well on sufficiently accurate variational $\alpha$-posteriors. More generally, the structure of the likelihood does not seem to affect the performance of the strategies, and mostly the number of parameters and the nature of the dataset have an influence. \\

To illustrate our results, in figure \ref{fig:polyreg_intro}, we compare the prediction curves returned by the sample splitting and Bayes in a polynomial regression model. They are computed from a set of noisy observations whose variance is misspecified and assumed to be too small. The sample splitting is better at estimating the function and shows reasonable smoothness and uncertainty, while Bayes is too confident and overfits the data. In figure \ref{fig:boxplots_example}, we compare the risks achieved by the strategies in a linear regression model on 30 repetitions of the experiment. The sample splitting and SafeBayes achieve the lowest risk and fall close to the minimum of the generalization error. \\

The second chapter of this work lists and describes each strategy in more detail, the third chapter describes the experimental statistical models and the results obtained, while the fourth chapter concludes the work. The detailed computations of each formula are explicitly detailed in the appendices of this work. Finally, all the code used to compute the experiments is available at \url{https://github.com/lucieperrotta/temperature_calibration}.

%% file: chapters/strategies.tex
\section{Strategies}
In this chapter, we give a more in-depth explanation of each strategy with both a mathematical description and a pseudo-code implementation. A strategy is a function of the form
\begin{align*}
    \text{Strategy}(\bm{X}, \mathcal{M}) \mapsto \alpha
\end{align*}
where $\bm{X}$ is the dataset generated from an unknown distribution $P$, and $\mathcal{M}$ the model to be fitted on the data (containing the hyperparameters). Note that the value of $\alpha/n$ is bounded in the interval $B$, typically $[0, 3]$, as larger values of $\alpha$ follow a similar behaviour and do not represent an interest to us. In the following, the clipping of the returned $\alpha$ value within the bounds is omitted for readability. Note that each strategy may have a different implementation depending on the complexity of the model, and different approximations may be used accordingly. \\

 We analyze the performance of the strategies by computing the generalization error $\mathcal{R}( \hat{\alpha})$ using the value $\hat{\alpha}$ returned by the strategy. The lower the error the better the performance of the strategy. As mentioned before, all strategies will perform at best as good as the lower bound \eqref{generalization_error}, and the closer they are to that lower bound, to more accurate their generalization error estimator is.

\subsection{Bayes}
The Bayesian strategy is the simplest, as it does not depend on the model nor the values of the data, and bases its choice uniquely on the size of the data. The mathematical function can be simply written as
\begin{align*}
    \text{Bayes}(\bm{X}, \mathcal{M}) =  |\bm{X}|
\end{align*}
where the value $|\bm{X}| = n$ is the size of the observations' dataset.\\

The pseudo-code implementation is as follows:

\begin{algorithm}[H]
\caption{Bayes strategy}
\SetKwFunction{FMain}{Bayes}
\SetKwProg{Fn}{function}{:}{}

\Fn{\FMain{$\bm{X}, \mathcal{M}$}}{
 $\alpha = |\bm{X} |$\\
 \Return $\alpha$}
\end{algorithm}
This strategy is prone to overfitting by taking too large values of $\alpha$ when the model is misspecified.

\subsection{Sample splitting}
In the sample splitting strategy, we first split the dataset $\bm{X}$ into two halves, named $\bm{X}^{(1)}$ and $\bm{X}^{(2)}$ respectively. We then compute the empirical risk of each half as
\begin{align*}
 r_n^{(1)}(\bm{\theta}) & = \frac{1}{(n/2)}\sum_{i=1}^{n/2} \ell(\bm{\theta},X_i), \\
 r_n^{(2)}(\bm{\theta}) & = \frac{1}{(n/2)}\sum_{i=1+n/2}^{n} \ell(\bm{\theta},X_i),
\end{align*}
where $\ell$ is the loss function specific to the model $\mathcal{M}$, typically the negative log likelihood. Then, an $\alpha$-posterior $\pi_{\alpha}^{(1)}$ is computed using the first batch uniquely. When $\pi_{\alpha}^{(1)}$ is computable in closed-form for a given model $\mathcal{M}$, we use
\begin{align*}
    \pi_{\alpha}^{(1)} ( d \bm{\theta}) = \underbrace{  \exp\left[-\alpha r_n^{(1)}(\bm{\theta}) \right] }_{\text{tempered likelihood}} \cdot \underbrace{  \pi_0(d\bm{\theta}) }_{\text{prior}}
\end{align*}
where the superscript $\cdot^{(1)}$ has been kept to emphasize the use of the first half only, and where the prior $\pi_0$ has tunable hyperparameters. When no closed-form is available, we use a variational approximation.\\

Next, we define the following estimate of the generalization error
\begin{align*}
    \mathcal{\hat{R}} (\alpha) := \mathbb{E}_{\bm{\theta}\sim \pi_\alpha^{(1)}} \left[ r_n^{(2)}(\bm{\theta}) \right]
\end{align*}
which we have to minimize. $ \mathcal{\hat{R}}$ is available in closed-form when the empirical risk computed for the model $\mathcal{M}$ is simple enough to explicitly compute the expectation of all the occurrences of $\bm{\theta}$ in it. In this case, $ \mathcal{\hat{R}}$ becomes a function of the parameters $\Omega(\alpha)$ of the $\alpha$-posterior and we can simply run an automatic minimization algorithm over $ \mathcal{\hat{R}}$ to find its minimum (where no gradient function is needed). When no closed-form is available, we approximate $ \mathcal{\hat{R}}$ using Monte-Carlo (MC) as
\begin{align*}
      \mathcal{\hat{R}} \stackrel{\text{MC}}{\approx}  \mathcal{\hat{R}}_\text{MC}(\alpha) := \frac{1}{mc} \sum_{i=1}^{mc} r_n^{(2)}(\bm{\theta}_i), \qquad \bm{\theta}_i \sim \pi_\alpha^{(1)} \forall i
\end{align*}
where $mc$ is the number of MC samples and is chosen large, and where a new value of $\bm{\theta}$ is sampled for each index of the sum. In this case, directly optimizing over $ \mathcal{\hat{R}}_\text{MC}$ is hard, as the function is now noisy because of the MC approximation. Having a closed-form of the gradient of $ \mathcal{\hat{R}}$ instead allows us to run a gradient descent algorithm to optimize the function. We hence propose
\begin{proposition}\label{proposition_covariance}
For the exact $\alpha$-posterior $\pi_\alpha^{(1)}$, we have
$$ \frac{\partial }{\partial \alpha} \mathbb{E}_{\bm{\theta}\sim \pi_\alpha^{(1)}} \left[ r_n^{(2)}(\bm{\theta}) \right] = -{\rm Cov}_{\bm{\theta}\sim \pi_\alpha^{(1)}} \left[ r_n^{(1)}(\bm{\theta}),r_n^{(2)}(\bm{\theta}) \right] . $$
\end{proposition}
The proof of this proposition can be found in the appendix \eqref{proof_of_covariance}. We can again approximate this value using MC,
\begin{align*}
\frac{\partial  \mathcal{\hat{R}}}{\partial \alpha} \stackrel{\text{MC}}{\approx} \left( \frac{\partial  \mathcal{\hat{R}}}{\partial \alpha} \right)_\text{MC}  &= - \frac{1}{mc} \sum_{i=1}^{mc} [r_n^{(2)}(\bm{\theta}_i) \cdot r_n^{(1)}(\bm{\theta}_i) ] \\
&\qquad \qquad + \frac{1}{(mc)^2} \sum_{i=1}^{m_c} [r_n^{(2)}(\bm{\theta}_i)] \cdot \sum_{i=1}^{mc} [ r_n^{(1)}(\bm{\theta}_i) ],  \bm{\theta}_i \sim \pi_\alpha^{(1)} \forall i.
\end{align*}
We then run a SGD algorithm using $ \left( \frac{\partial  \mathcal{\hat{R}}}{\partial \alpha} \right)_\text{MC}$ as the gradient.\\

Thus, we obtain the following mathematical function
\begin{align*}
    \text{SampleSplit}(\bm{X}, \mathcal{M}) = \begin{cases}
        \argmin_\alpha   \mathcal{\hat{R}}(\alpha) \\
        \qquad \text{when }  \mathcal{\hat{R}} \text{ is available in closed-form for model $\mathcal{M}$,} \\
        \alpha^* \text{, the output of the SGD with } \alpha := \alpha - \eta \cdot \left( \frac{\partial  \mathcal{\hat{R}}}{\partial \alpha} \right)_\text{MC} \\
        \qquad \text{when }  \mathcal{\hat{R}} \text{ is not available in closed-form for model $\mathcal{M}$.}
    \end{cases}
\end{align*}
where $\eta$ is an adaptive learning rate parameter. This translates the to pseudocode

\begin{algorithm}[H]
\caption{Sample splitting strategy}
\SetKwFunction{FMain}{SampleSplit}
\SetKwProg{Fn}{function}{:}{}

\Fn{\FMain{$\bm{X}, \mathcal{M} $}}{
 $r_n^{(1)}(\bm{\theta}) = \frac{1}{(n/2)}\sum_{i=1}^{n/2} \ell(\bm{\theta},X_i); \quad r_n^{(2)}(\bm{\theta}) = \frac{1}{(n/2)}\sum_{i=n/2+1}^{n} \ell(\bm{\theta},X_i)$\\
 compute the parameters $ \Omega_P(\alpha) $ of the posterior\\
 compute $\pi_{\alpha}^{(1)} =  \exp\left[-\alpha r_n^{(1)}(\bm{\theta}) \right]\cdot \pi_0(\rm d\bm{\theta}) $ using $\Omega_P (\alpha)$ \\
 \If{$\mathcal{M}$ has an available closed-form function $ \mathcal{\hat{R}}$}{
 compute $ \mathcal{\hat{R}}(\alpha) = \mathbb{E}_{\bm{\theta}\sim \pi_\alpha^{(1)}} \left[ r_n^{(2)}(\bm{\theta}) \right] $ using $\Omega_P(\alpha)$\\
 \Return $ \min_\alpha \mathcal{\hat{R}}$ using any optimizer
 }
 \Else{
 \While{SGD has not converged}{
 sample $mc$ values of $\bm{\theta} $ from $ \pi_\alpha^{(1)}$\\
 update $ \alpha := \alpha - \eta \cdot \left( \frac{\partial  \mathcal{\hat{R}}}{\partial \alpha} \right)_\text{MC} $
 }
 \Return $\alpha$
  }
 }
\end{algorithm}

\subsection{Naive}
The pipeline of the computation is essentially the same, with the only difference being that the whole dataset $\bm{X}$ is used two times instead of $\bm{X}^{(1)}$ and $\bm{X}^{(2)}$ respectively, in all the computations. The estimate $\mathcal{\hat{R}}(\alpha)$ hence becomes
\begin{align*}
    \mathcal{\hat{R}}(\alpha) := \mathbb{E}_{\bm{\theta}\sim \pi_\alpha} \left[ r_n(\bm{\theta}) \right]
\end{align*}
and its derivative
\begin{align}\label{variance_naive}
    \frac{\partial \mathcal{\hat{R}}}{\partial \alpha}(\alpha) = -{\rm Cov}_{\bm{\theta}\sim \pi_\alpha} \left[ r_n(\bm{\theta}),r_n(\bm{\theta}) \right] = - \text{Var}_{\bm{\theta}\sim \pi_\alpha} \left[ r_n(\bm{\theta}) \right]
\end{align}
which is obviously negative. This means that the naive strategy will take $\alpha$ as large as possible. The formulation of the mathematical function and the pseudocode are simply computed by applying the same replacement and are hence omitted here. As mentioned in the introduction, the value of $ \mathcal{\hat{R}}$ is both computed and sampled from $\bm{X}$, which can be seen as training and testing the model $\mathcal{M}$ on the same data. Hence, the data is a perfect predictor of itself and the model tends to overfit. The naive strategy gives importance to the data through the likelihood rather than to the prior, and returned values of $\alpha$ will always be equal to the upper bound $B$. 

\subsection{Bootstrapping}
The bootstrapping strategy uses the bootstrap theory presented in \cite{10.2307/2958830}. Assuming that we have no information about $P$, we use the MC-bootstrap algorithm for case resampling to generate a new bootstrap dataset as
\begin{align*}
    \bm{\Tilde{X}} := (\tilde{X}_1,\dots,\tilde{X}_n)
\end{align*}
where each $\Tilde{X}_i$ is drawn uniformly from $ \bm{X} $ with replacement. The pipeline is similar to the sample splitting, except that each computation is done $boot$ different times, using iteratively $boot$ different $\bm{\Tilde{X}}^{(b)}$ datasets where $b \in (1, boot)$, with dataset $\bm{X}$, and are averaged together in a MC fashion. We typically set $boot \sim 1000$.\\

The pipeline is as follows. First, the empirical risk $r_n$ is computed for the dataset $\bm{X}$, as well as the empirical risks $\tilde{r}_n^\text{(b)}$ for each of the $boot$ datasets $\bm{\Tilde{X}}^{(b)}$.
\begin{align*}
 r_n(\bm{\theta}) & = \frac{1}{n}\sum_{i=1}^{n} \ell(\bm{\theta},X_i), \\
 \Tilde{r}^{(b)}_n(\bm{\theta}) & = \frac{1}{n}\sum_{i=1}^{n} \ell(\bm{\theta},\Tilde{X}_i^{(b)}).
\end{align*}
Then, the $\alpha$-posterior
\begin{align*}
    \Tilde{\pi}_{\alpha}^{(b)} ( d \bm{\theta}) = \underbrace{  \exp\left[-\alpha \tilde{r}_n^{(b)}(\bm{\theta}) \right] }_{\text{tempered likelihood}} \cdot \underbrace{  \pi_0(d\bm{\theta}) }_{\text{prior}}
\end{align*}
is computed for each bootstrap dataset $\bm{\Tilde{X}}^{(b)}$. The generalization error estimate $ \mathcal{\hat{R}}$ is then computed $boot$ times with the dataset $\bm{X}$ as
\begin{align*}
     \mathcal{\hat{R}}^{(b)}(\alpha) := \mathbb{E}_{\bm{\theta}\sim \tilde{\pi}_{\alpha}^{(b)}} \left[ r_n(\bm{\theta}) \right]
\end{align*}
as well as its derivative, either in explicit form when available, or using the MC approximation otherwise. The values obtained from the derivative (there is a number $boot$ of them) are then averaged together, and this new value $\Psi$ is used in an SGD algorithm,
\begin{align*}
    \Psi  &= \frac{1}{boot} \sum_{b=1}^{boot} \frac{\partial }{\partial \alpha} \mathcal{\hat{R}}^{(b)}(\alpha) \\
    \Psi_\text{MC} &= \frac{1}{boot \cdot mc} \sum_{b=1}^{boot} \sum_{i=1}^{mc}  \frac{\partial }{\partial \alpha} r_n(\bm{\theta_i}) , \quad \bm{\theta_i} \sim \tilde{\pi}_\alpha^{(b)} .
\end{align*}
Indeed, due to the noisy nature of the MC-bootstrap averaging, a direct optimization over one noisy computation of $ \mathcal{\hat{R}}^{(b)}(\alpha)$ would be biased, and hence the SGD alternative is always preferred, unlike the sample splitting strategy.\\

The mathematical function therefore reads
\begin{align*}
    \text{Bootstrap}(\bm{X}, \mathcal{M}) = \begin{cases}
        \alpha^* \text{, output of the SGD with } \alpha := \alpha -  \eta \cdot\Psi \\
        \qquad \text{when } \frac{\partial  \mathcal{\hat{R}}^{(b)}}{\partial \alpha} \text{ is available in closed-form for model $\mathcal{M}$,} \\
        \alpha^* \text{, output of the SGD with } \alpha := \alpha -  \eta \cdot\Psi_\text{MC} \\
        \qquad \text{when } \frac{\partial  \mathcal{\hat{R}}^{(b)}}{\partial \alpha} \text{ is not available in closed-form for model $\mathcal{M}$.}
    \end{cases}
\end{align*}

The equivalent pseudocode is

\begin{algorithm}[H]
\caption{Bootstrapping strategy}
\SetKwFunction{FMain}{Bootstrap}
\SetKwProg{Fn}{function}{:}{}

\Fn{\FMain{$\bm{X}, \mathcal{M} $}}{
 $r_n(\bm{\theta}) = \frac{1}{n}\sum_{i=1}^{n} \ell(\bm{\theta},X_i)$\\
 \While{SGD has not converged}{
 \For{$b = 1$ to $boot$}{
 draw a new $\bm{\Tilde{X}}^{(b)} = (\tilde{X}_1,\dots,\tilde{X}_n)$ uniformly from $\bm{X}$ \\
  $ \Tilde{r}_n^{(b)}(\bm{\theta}) = \frac{1}{n}\sum_{i=1}^{n} \ell(\bm{\theta},\Tilde{X}^{(b)}_i)$\\
  compute $\Tilde{\pi}^{(b)}_{\alpha} =  \exp\left[-\alpha \Tilde{r}_n^{(b)}(\bm{\theta}) \right]\cdot \pi_0(\rm d\bm{\theta}) $ \\
  
 \If{$\mathcal{M}$ has an available closed-form function $\frac{\partial  \mathcal{\hat{R}}^{(b)}}{\partial \alpha} $}{
 compute $ \frac{\partial  \mathcal{\hat{R}}^{(b)}}{\partial \alpha}$ 
 }
 \Else{
 sample $mc$ values of $\bm{\theta} $ from $ \tilde{\pi}^{(b)}_\alpha $\\
 compute numerically $ \left( \frac{\partial  \mathcal{\hat{R}}^{(b)}}{\partial \alpha} \right)_\text{MC}$
 }
  }
  compute the average $\Psi$ or $\Psi_\text{MC}$ \\
  update $ \alpha := \alpha -  \eta \cdot \Psi $ or  $ \alpha := \alpha -  \eta \cdot\Psi_\text{MC}$ \\
  }
  \Return $\alpha$
 }
\end{algorithm}

\subsection{SafeBayes}
The SafeBayes strategy is implemented as it was proposed by Peter Grünwald in \cite{Grnwald2012TheSB, grnwald2014inconsistency} as a robust strategy against misspecification in models where an exact posterior is available. The approach of this strategy is to compare the score of many subsets of the dataset $\bm{X}$. We first consider the empirical risk function \textit{up to observation $ t$}
\begin{align*}
    r_n^{(t)} (\bm{\theta}) = \frac{1}{t} \sum_{i=1}^t \ell (\bm{\theta}, X_i)
\end{align*}
where the dataset $\bm{X}$ is only evaluated from its first up to its $t$-th observation. Note that the superscript $\cdot^{(t)}$ is here a scalar parameter of the risk function. Similarly, we compute the $\alpha$-posterior \textit{up to observation $t$ }as
\begin{align*}
    \pi_\alpha^{(t)}  ( d \bm{\theta}) = \exp \left[ - \alpha r_n^{(t)} (\bm{\theta}) \right] \pi( \rm d \bm{\theta}) .
\end{align*}
We now define the \textit{expected loss} up to observation $t$, where the observation $t+1$ is predicted by a posterior trained on observations $1$ to $t$ as
\begin{align*}
    \mathcal{E}(\alpha, t) := \mathbb{E}_{\bm{\theta}\sim \pi_\alpha^{(t)} } \left[ \ell(\bm{\theta}, X_{t+1} ) \right].
\end{align*}
All expected losses for values of $t \in \{1, ... , n-1\}$ are finally summed up together to obtain a function of $\alpha$ only:
\begin{align*}
    \mathcal{S}(\alpha) := \sum_{t=1}^{n-1} \mathcal{E}( \alpha, t) .
\end{align*}
Grünwald denotes this function as the \textit{posterior-expected posterior-randomized loss (PEPRL)} of predicting the next observation. He observes that this strategy tends to select small values of $\alpha$, and we may expect it to underestimate $\alpha^*$ rather than the contrary.\\

While Grünwald proposes this strategy for models where all computations are available in closed-form, we extend the strategy to more complex models where $\mathcal{E}$ may not be available in explicit form and hence approximated with MC, as well as the $\alpha$-posterior which we can estimate using variational inference. As stated before, complex models lead to noisy functions which are hard to optimize, and we instead compute the derivative of the expected loss in a very similar fashion to the sample splitting strategy by slightly modifying proposition \eqref{proposition_covariance} into the proposition
\begin{proposition}
For the exact $\alpha$-posterior $\pi_\alpha^{(t)}$, we have
$$ \frac{\partial \mathcal{E}}{\partial \alpha}(\alpha, t) = -{\rm Cov}_{\bm{\theta}\sim \pi_\alpha^{(t)}} \left[ \ell(\bm{\theta}, X_{t+1} ) ,r_n^{(t)}(\bm{\theta}) \right] . $$
\end{proposition}
The proof is very similar to that of proposition $\eqref{proposition_covariance}$ and is hence omitted. This following formula follows by linearity of the derivation
\begin{align*}
    \frac{\partial \mathcal{S}}{\partial \alpha}(\alpha) := \sum_{t=1}^{n-1} \frac{\partial \mathcal{E}}{\partial \alpha}(\alpha, t) .
\end{align*}
Finally, a MC approximation $\left( \frac{\partial \mathcal{S}}{\partial \alpha} \right)_\text{MC}$ is used inside an SGD optimizer. \\

The strategy can be summarized as
\begin{align*}
    \text{SafeBayes}(\bm{X}, \mathcal{M}) = \begin{cases}
        \argmin_\alpha  \mathcal{S}(\alpha) \\
        \qquad \text{when } \mathcal{E} \text{ is available in closed-form for model $\mathcal{M}$,} \\
        \alpha^* \text{, output of the SGD with } \alpha := \alpha -  \eta \cdot \left( \frac{\partial \mathcal{S}}{\partial \alpha} \right)_\text{MC} \\
        \qquad \text{when } \mathcal{E} \text{ is not available in closed-form for model $\mathcal{M}$.}
    \end{cases}
\end{align*}

\begin{algorithm}[H]
\caption{SafeBayes strategy}
\SetKwFunction{FMain}{SafeBayes}
\SetKwProg{Fn}{function}{:}{}

\Fn{\FMain{$\bm{X}, \mathcal{M} $}}{
\For{ $t = 1$ to $n-1$ }{
 $r_n^{(t)}(\bm{\theta}) = \frac{1}{t}\sum_{i=1}^{t} \ell(\bm{\theta},X_i)$\\
 compute the parameters $ \Omega_P(\alpha) $ of the posterior\\
 compute $\pi_{\alpha}^{(t)} =  \exp\left[-\alpha r_n^{(t)}(\bm{\theta}) \right]\cdot \pi_0(d\bm{\theta}) $ using $\Omega_P(\alpha)$ \\
 compute $ \mathcal{E}(\alpha, t ) = \mathbb{E}_{\bm{\theta}\sim \pi_\alpha^{(t)}} \left[ \ell ( \bm{\theta}, X_{t+1}) \right] $ exactly or using MC \\
 }
 \If{$\mathcal{M}$ has an available closed-form function $\mathcal{E}$}{
 compute $ \mathcal{S}(\alpha, t ) =\sum_{t=1}^{n-1} \mathcal{E}( \alpha, t)$\\
 \Return $ \argmin_\alpha \mathcal{S} (\alpha) $ using any minimizer
 }
 \Else{
 \While{SGD has not converged}{
 sample $mc$ values of $\bm{\theta} $ from $ \pi_\alpha^{(t)}$\\
 update $ \alpha := \alpha -  \eta \cdot \left( \frac{\partial \mathcal{S}}{\partial \alpha} \right)_\text{MC} $
 }
 \Return $\alpha$
  }
  }
\end{algorithm}

%% file: chapters/results.tex
\section{Experimental results}
In this chapter, we compare the performance of the strategies over three statistical models: the linear regression with known and unknown variance, and the logistic regression. For each model, we first explicitly compute the different functions needed to run each of the strategies. We then run the strategies over the models with different data settings, noise distributions, and number of parameters, in order to analyze the effects of misspecification and model size on each strategy. We draw boxplots summarizing the results of 30 repetitions of the experiment. Next to the five strategies boxplots, we also create a sixth boxplot for the minimal value of the optimal generalization error $\mathcal{R}(\alpha^*)$ on the right, used as a lower bound to what strategies can perform.\\

As explained in the previous chapter, each strategy is implemented differently according to the statistical model. In the linear regression with known and unknown variance, the posterior is available in closed-form. For naive, sample splitting and SafeBayes strategies, the estimate of the generalization error is computed exactly and is a smooth function of $\alpha$ that can be optimized efficiently using a automatic minimization algorithm such as \texttt{scipy.optimize} \cite{2020SciPy-NMeth}, which does not require to explicitly compute the gradients of the function. The bootstrapping strategy, in contrast, needs to average many computations of the generalization error in a MC fashion, giving a noisy estimate of the generalization error. Automatic minimization algorithms typically perform poorly on noisy functions, so we instead use the proposition \eqref{proposition_covariance} to compute an explicit gradient that we use in a SGD algorithm. In the logistic regression, the $\alpha$-posterior is approximated with variational inference, and the generalization error has no closed-form. It is hence approximated with MC and becomes noisy for all strategies, and the SGD algorithm is always used. The implementation choices are summarized in table \ref{table-backbone}. \\

\begin{table}[ht]
\centering
\begin{tabular}{>{\hspace{0pt}}p{0.3\linewidth}>{\hspace{0pt}}p{0.3\linewidth}>{\hspace{0pt}}p{0.3\linewidth}} 
\hline
 & Exact $\alpha$-posterior  & Variational $\alpha$-posterior  \\ 
\hline
Exact $\mathcal{\hat{R}}$ \par \texttt{scipy.optimize} & Linear regression: \par \textit{Naive},  \par \textit{sample splitting,} \par  \textit{SafeBayes} & \\ 
\hline
MC approximated $\mathcal{\hat{R}}$ \par SGD with \eqref{proposition_covariance}  & Linear regression: \par  \textit{Bootstrapping}  &  Logistic regression:  \par \textit{Naive},  \par \textit{sample splitting,}  \par \textit{bootstrapping}, \par  \textit{SafeBayes} \\
\hline
\end{tabular}
\caption{Backbone implementation of the strategies depending on the model.}
\label{table-backbone}
\end{table}

\subsection{Linear regression with known variance}
\subsubsection{Model setup}
The first model we analyze is the Bayesian linear regression with known fixed variance. In this model, the dataset $\bm{X}$ is composed of input-output tuples $X_i = (Z_i, Y_i)$ generated as follows:
\begin{align*}
    Y_i =  Z_i^\top \bm{\theta} + \varepsilon_i, \quad \varepsilon_i \sim \mathcal{N}(0, \sigma^2)
\end{align*}
where $Y_i$ is a scalar, $Z_i$ is a vector of size $d \times 1$, and $\bm{\bm{\theta}}$ is a parameter vector of dimension $d \times 1$. $\varepsilon_i$ is a scalar Gaussian additive noise with a known fixed variance $\sigma^2$. This can be rewritten in vector notation as
\begin{align*}
    \bm{Y} = \bm{\theta}\bm{Z} + \bm{\varepsilon} 
\end{align*}
where $\bm{Z}$ is a matrix of dimension $n \times d$ whose lines are transposed versions of  $Z_i$ of size $d \times 1$, and $\bm{Y}$ and $\bm{\varepsilon}$ are vectors of dimension $n \times 1$.\\

The likelihood of this model is
\begin{align*} \mathcal{L}(\bm{\theta}, \bm{X}) &= \left( \frac{1}{\sqrt{2 \pi \sigma^2}} \right)^{|\bm{X}|} \exp \left\{ -\frac{1}{2\sigma^2}  (\bm{Y} - \bm{Z} \bm{\theta})^\top (\bm{Y} - \bm{Z} \bm{\theta}) \right\} \\
&\propto \exp \left\{ -\frac{1}{2\sigma^2}  (\bm{Y} - \bm{Z} \bm{\theta})^\top (\bm{Y} - \bm{Z} \bm{\theta}) \right\}
\end{align*}
and we define the loss function to be the negative log likelihood
\begin{align*}
     \ell (\bm{\theta}, X_i) = \frac{(Y_i - Z_i^\top \bm{\theta})^2}{2 \sigma^2}.
\end{align*}
Additionally, we choose the prior to be Gaussian, 
\begin{align*}
    \pi (d \bm{\theta}) \sim \mathcal{N} (\bm{\mu}_0, \bm{S}_0 )
\end{align*}
where $\Omega_0 = (\bm{\mu}_0 , \bm{S}_0)$ are tunable hyperparameters, the mean vector and the full covariance matrix of the Gaussian distribution. We typically set them to
\begin{align*}
    \bm{\mu_0} &= \bm{0} \\
    \bm{S}_0 &= \bm{I}_d .
\end{align*}
Using this as well as the closed-form formula for the linear regression's Bayesian posterior \cite{bishop_book, ahmedLinReg, stupido}, we can find an explicit conjugate Gaussian $\alpha$-posterior
\begin{align}\label{linear_regression_posterior_gaussian}
    \pi_{\alpha} ( d \bm{\theta}) \sim \mathcal{N} \left( \bm{\mu}_P, \bm{S}_P \right),
\end{align}
where the parameters $\Omega_P (\alpha)$ are
\begin{align*}
     \bm{S}_P &= \left( \frac{\alpha}{\sigma^2 |\bm{X}|}  \bm{Z}^{ \top} \bm{Z} + \bm{S}_0^{-1} \right)^{-1} \\
     \bm{\mu}_P &= \bm{S}_P \left( \frac{ \alpha}{\sigma^2 |\bm{X}|} \bm{Z}^{ \top} \bm{Y} + \bm{S}_0^{-1} \bm{\mu}_0 \right).
\end{align*}
All the complete derivations of the results can be found in the appendix \ref{appendix_linreg1}. Note that the other strategies' posteriors $\pi_\alpha^{(1)}$ and $\pi_\alpha^{(t)}$ are computed similarly, using their respective datasets $\bm{X}^{(1)}$ and $\bm{X}^{(t)}$ in the calculations.

\subsubsection{Datasets generation}
We analyze the performance of the strategies of the linear regression model with three different settings:
\begin{enumerate}
    \item Well specified linear regression
    $$ Y_i = \bm{\theta} Z_i + \varepsilon_i , \quad \varepsilon_i \sim \mathcal{N} (\theta, \sigma^2)$$
    \item Gaussian mean estimation
    $$ Y_i = \theta + \varepsilon_i, \quad \varepsilon_i \sim \mathcal{N} (\theta, 1) $$
    where $\theta$ is scalar
    \item Polynomial regression
    $$ Y_i = f(\zeta_i) + \varepsilon_i  , \quad \varepsilon_i \sim \mathcal{N} (\theta, \sigma^2)$$
    where $f$ is a smooth function, $\bm{\zeta}$ is a $n \times 1$ vector, and $\sigma^2$ is misspecified.
\end{enumerate}
The Gaussian mean estimation case is described in more details in appendix \ref{gaussian_setup}, and the polynomial regression in appendix \ref{polynomial_setup}.

\subsubsection{Strategies performances}

\paragraph{Well specified linear regression}

\begin{figure}[ht]
    \centering
    \includegraphics[width=.9\textwidth]{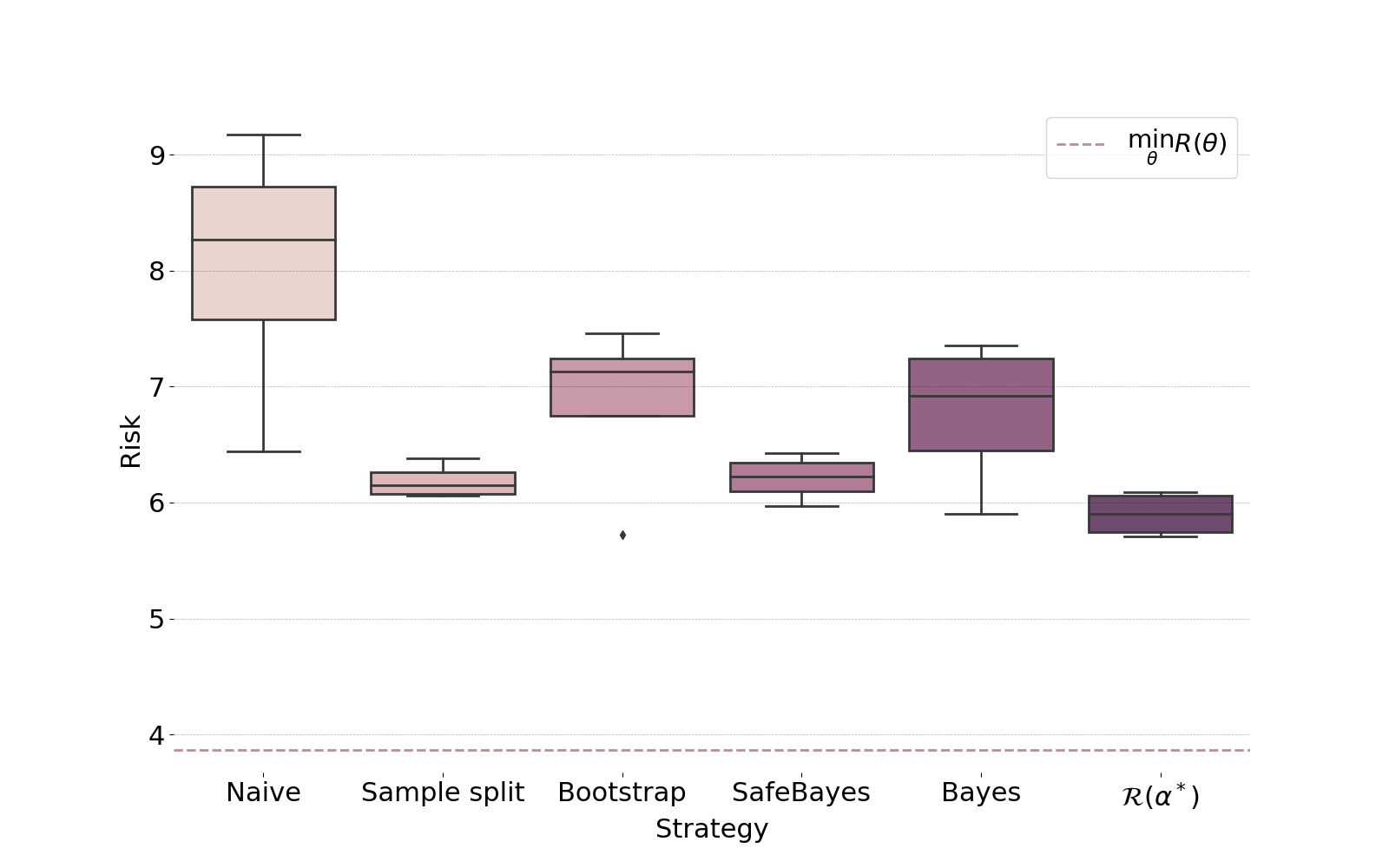}
    \caption{Boxplots of the linear regression with known variance when $ n = d = 40$, and $\sigma^2=8$. The sample splitting and SafeBayes strategies perform the best.}
    \label{fig:linreg1_boxplot}
\end{figure}

\begin{figure}[pt]
\centering

\begin{minipage}{.5\textwidth}
  \centering
  \includegraphics[width=\linewidth]{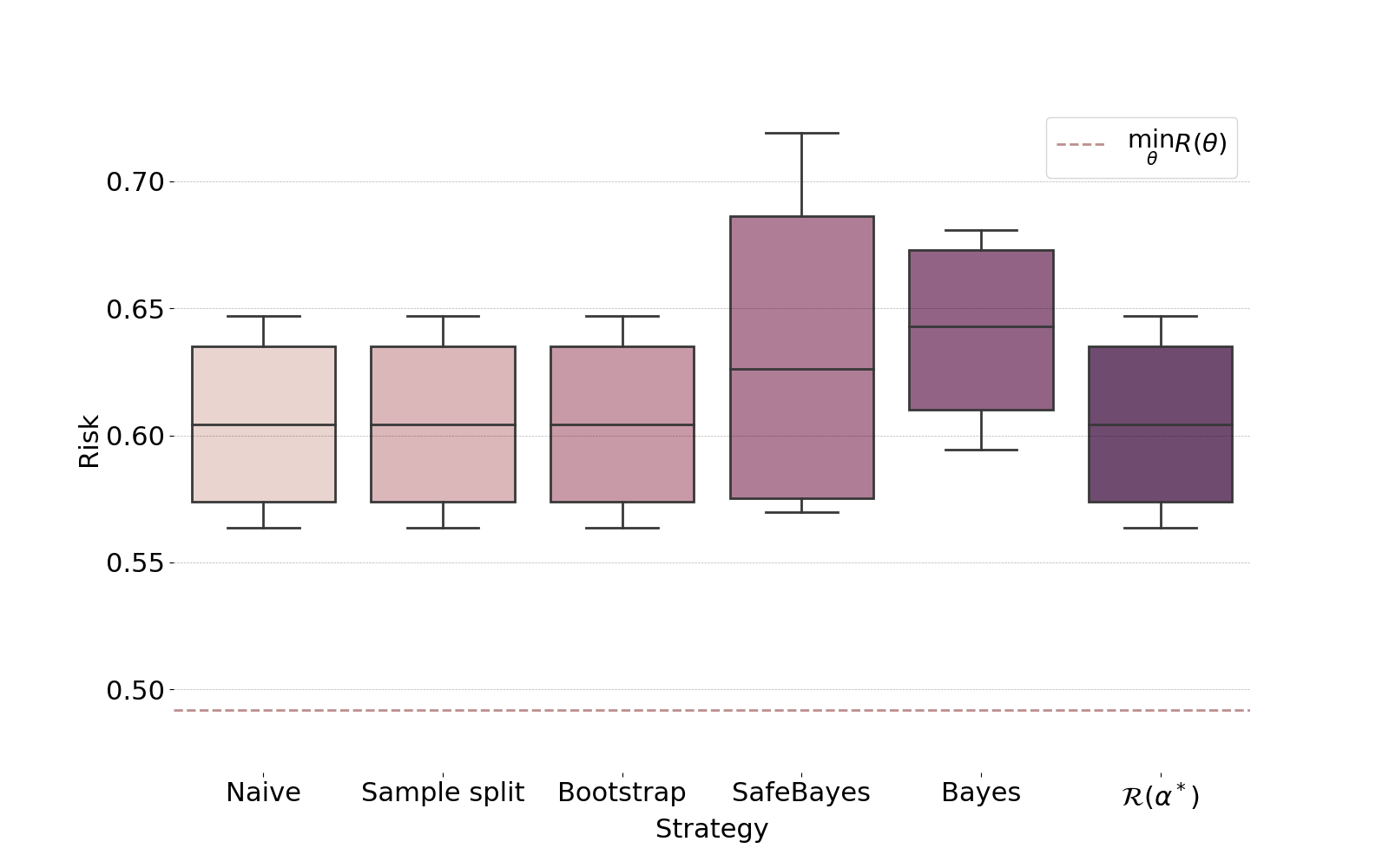}
\end{minipage}%
\begin{minipage}{.5\textwidth}
  \centering
  \includegraphics[width=\linewidth]{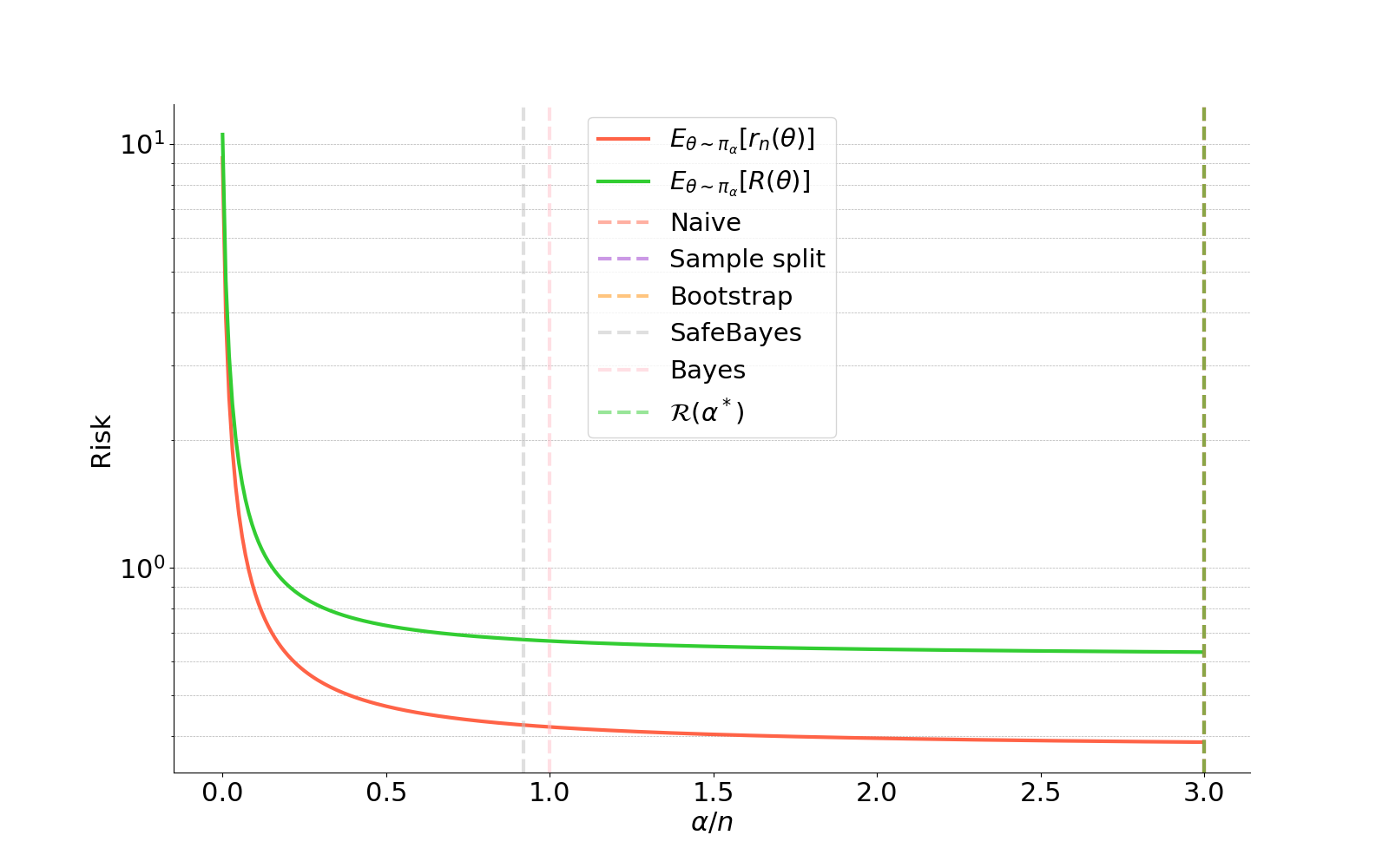}
\end{minipage}
\caption{Low noise and many observations, when $n=100, d=20, \sigma^2=4$.}
\label{fig:linreg1_additional_plots1}

\begin{minipage}{.5\textwidth}
  \centering
  \includegraphics[width=\linewidth]{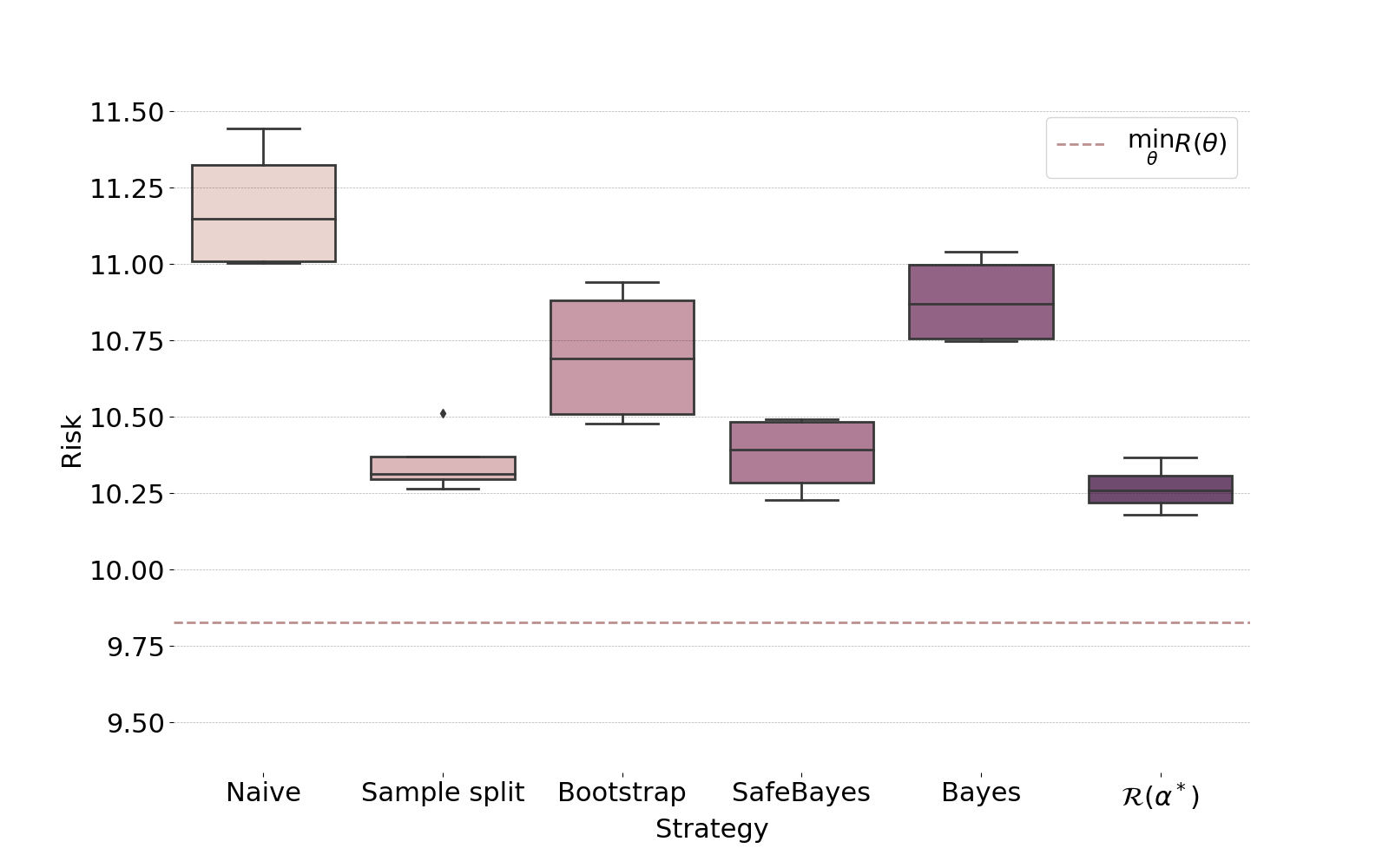}
\end{minipage}%
\begin{minipage}{.5\textwidth}
  \centering
  \includegraphics[width=\linewidth]{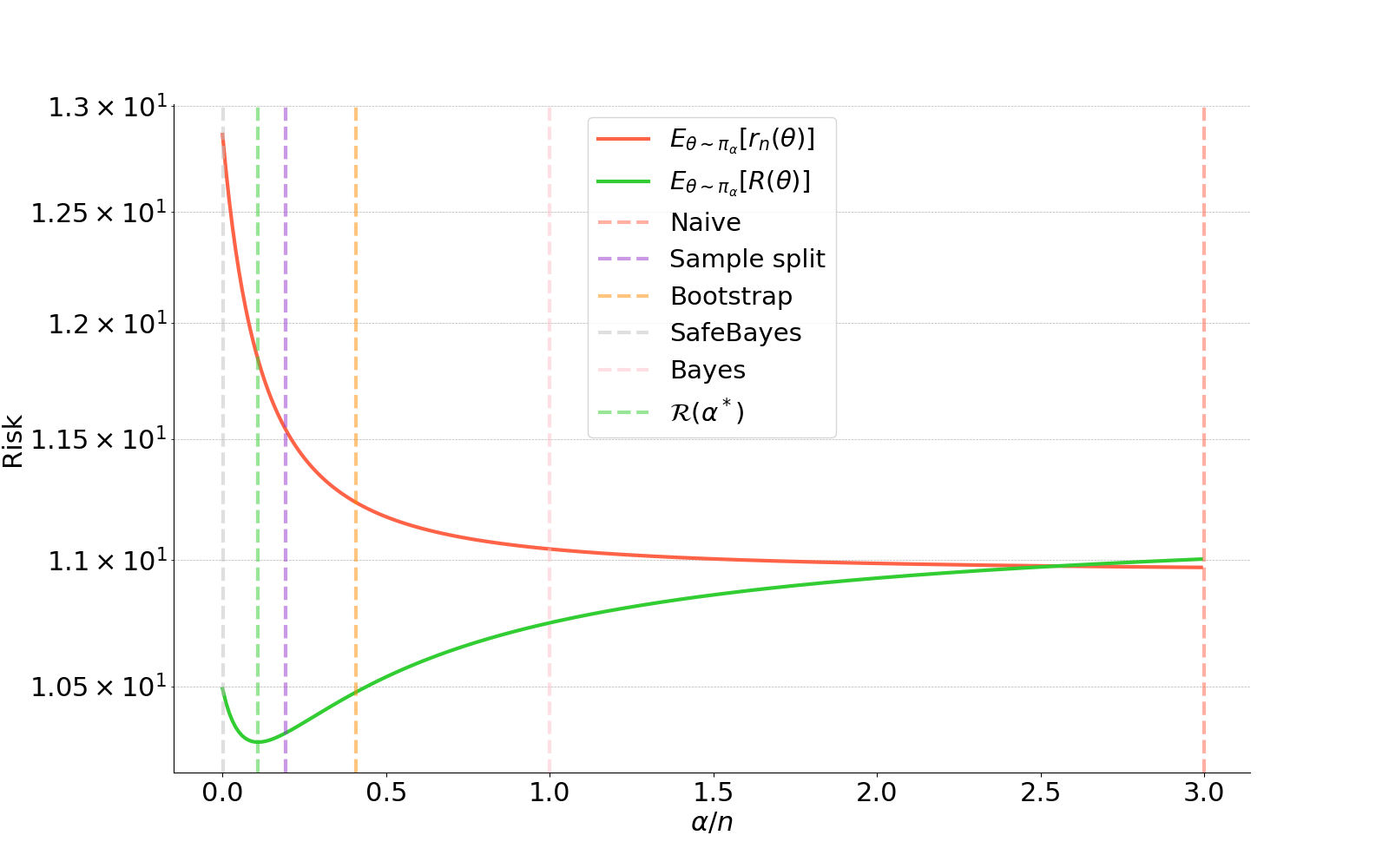}
\end{minipage}
\caption{High noise and many observations, when $n=100, d=20, \sigma^2=20$.}
\label{fig:linreg1_additional_plots2}

\begin{minipage}{.5\textwidth}
  \centering
  \includegraphics[width=\linewidth]{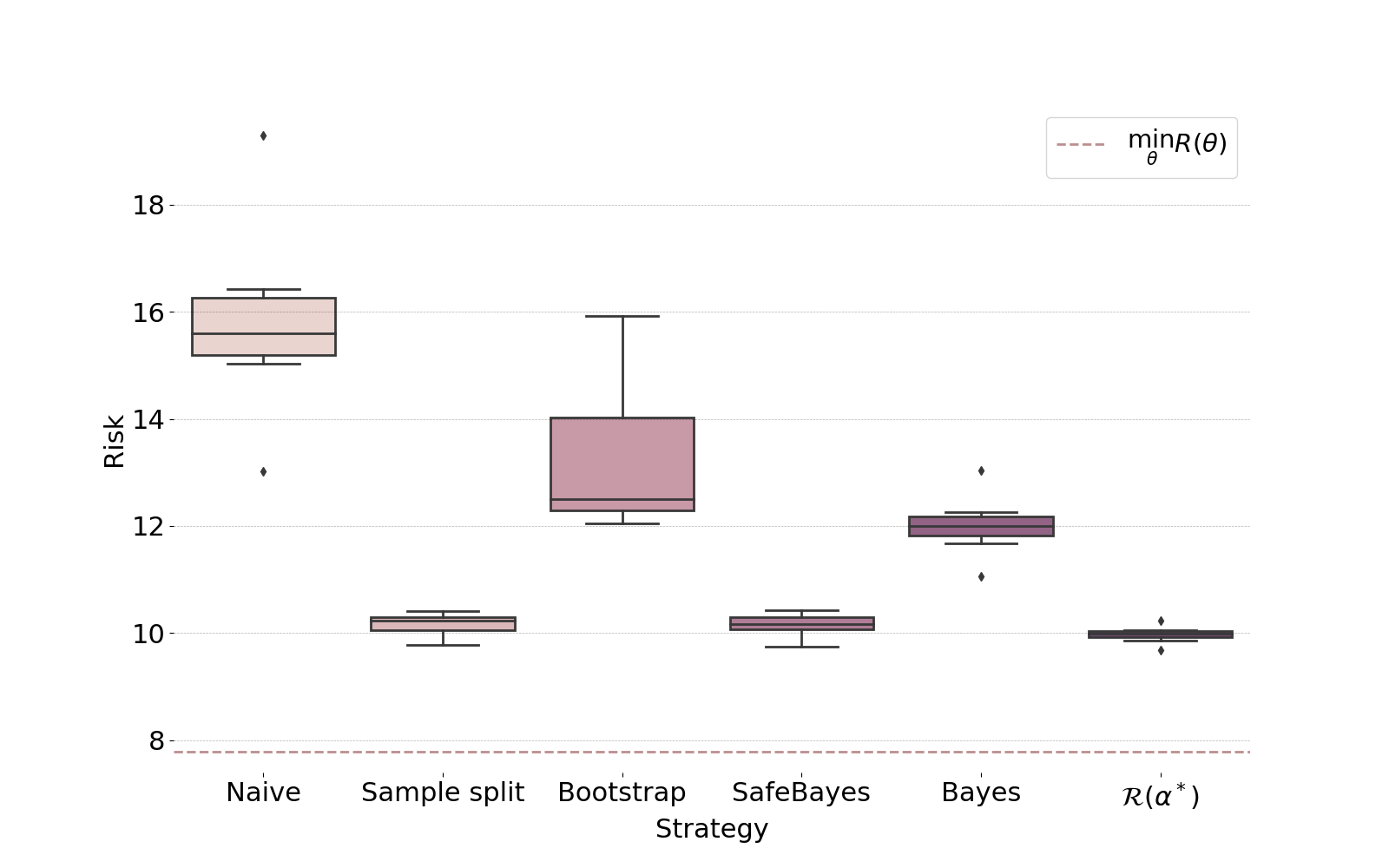}
\end{minipage}%
\begin{minipage}{.5\textwidth}
  \centering
  \includegraphics[width=\linewidth]{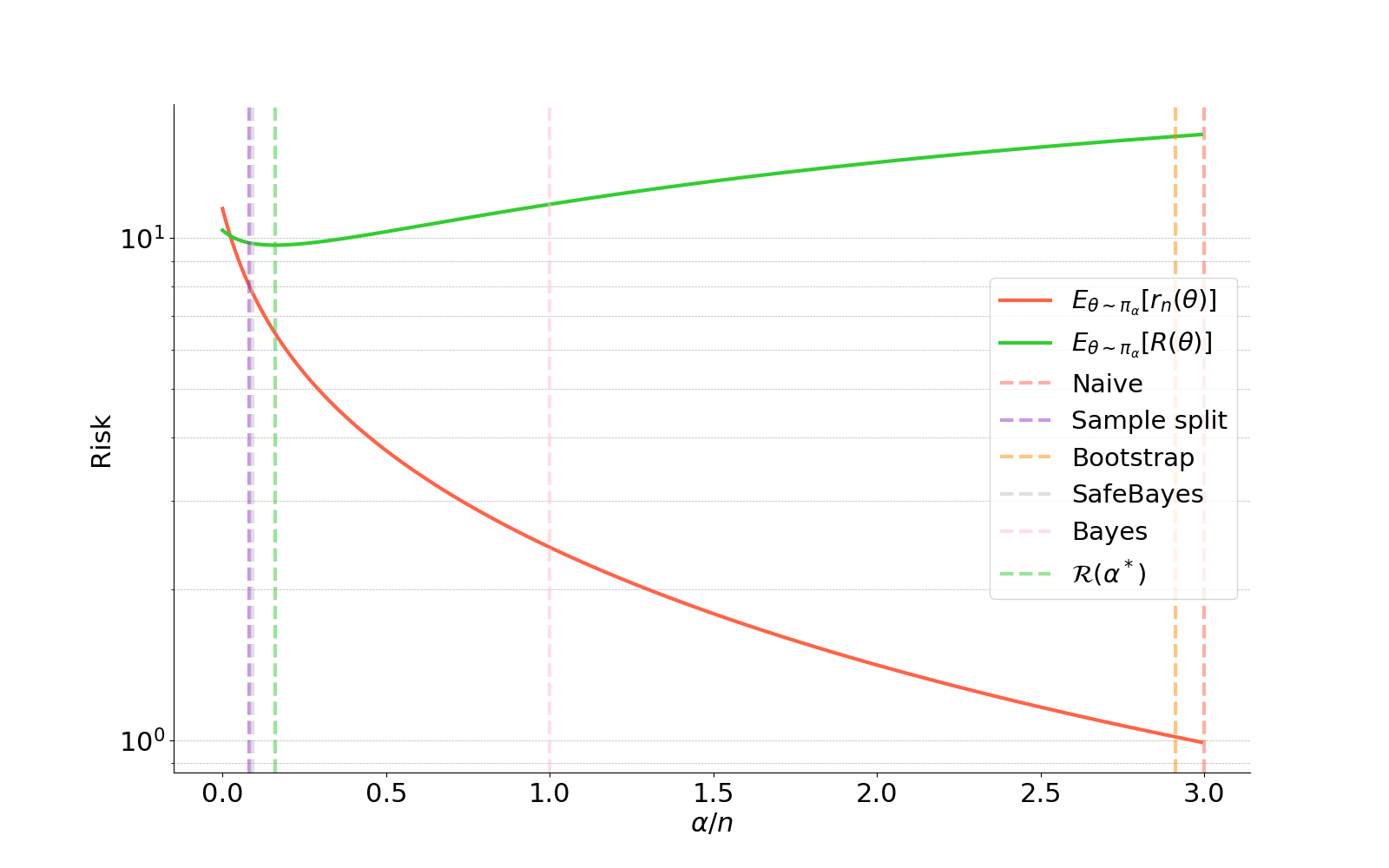}
\end{minipage}
\caption{High noise and few observations, $n=40, d=40, \sigma^2=16$.}
\label{fig:linreg1_additional_plots3}

\begin{minipage}{.5\textwidth}
  \centering
  \includegraphics[width=\linewidth]{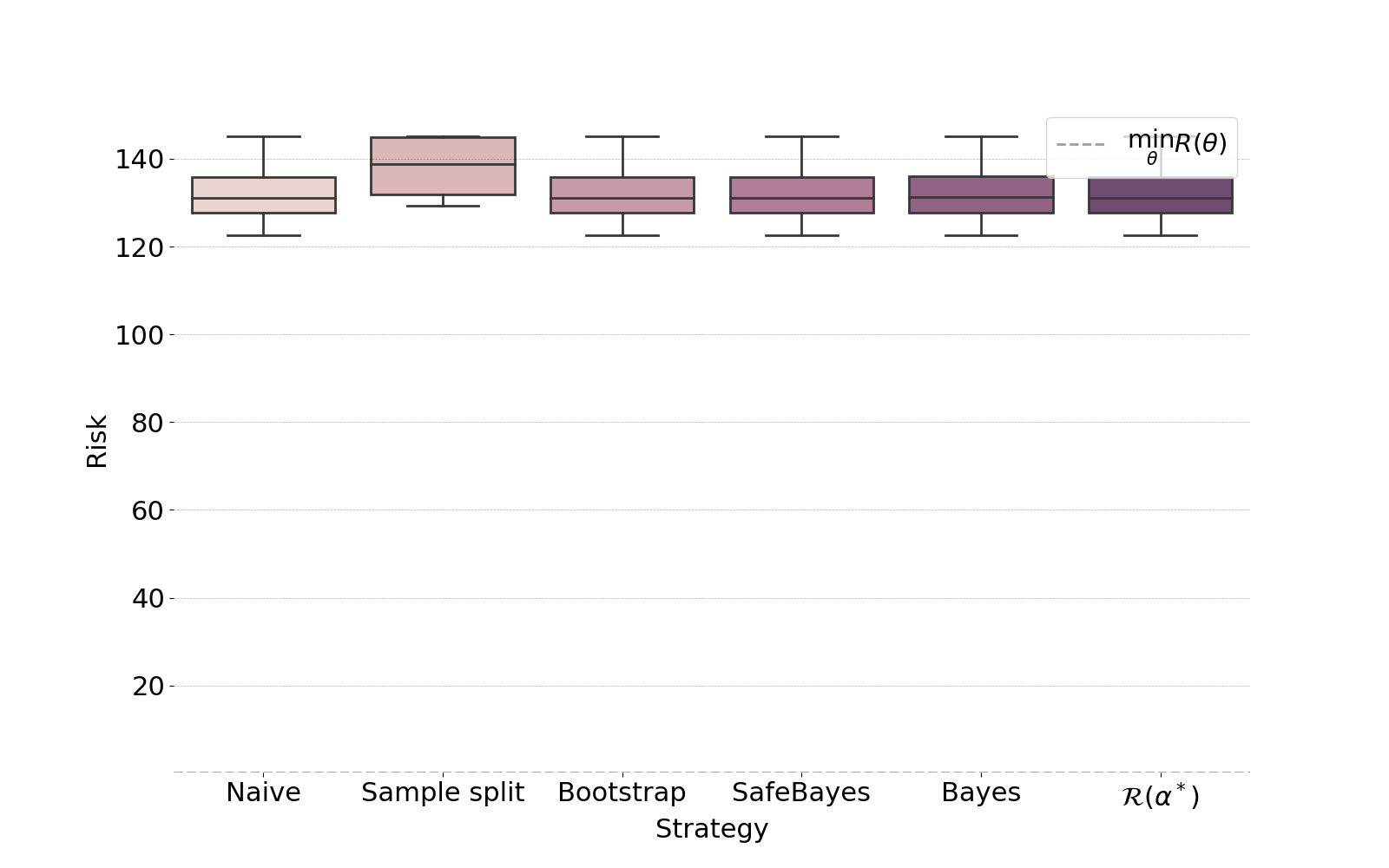}
\end{minipage}%
\begin{minipage}{.5\textwidth}
  \centering
  \includegraphics[width=\linewidth]{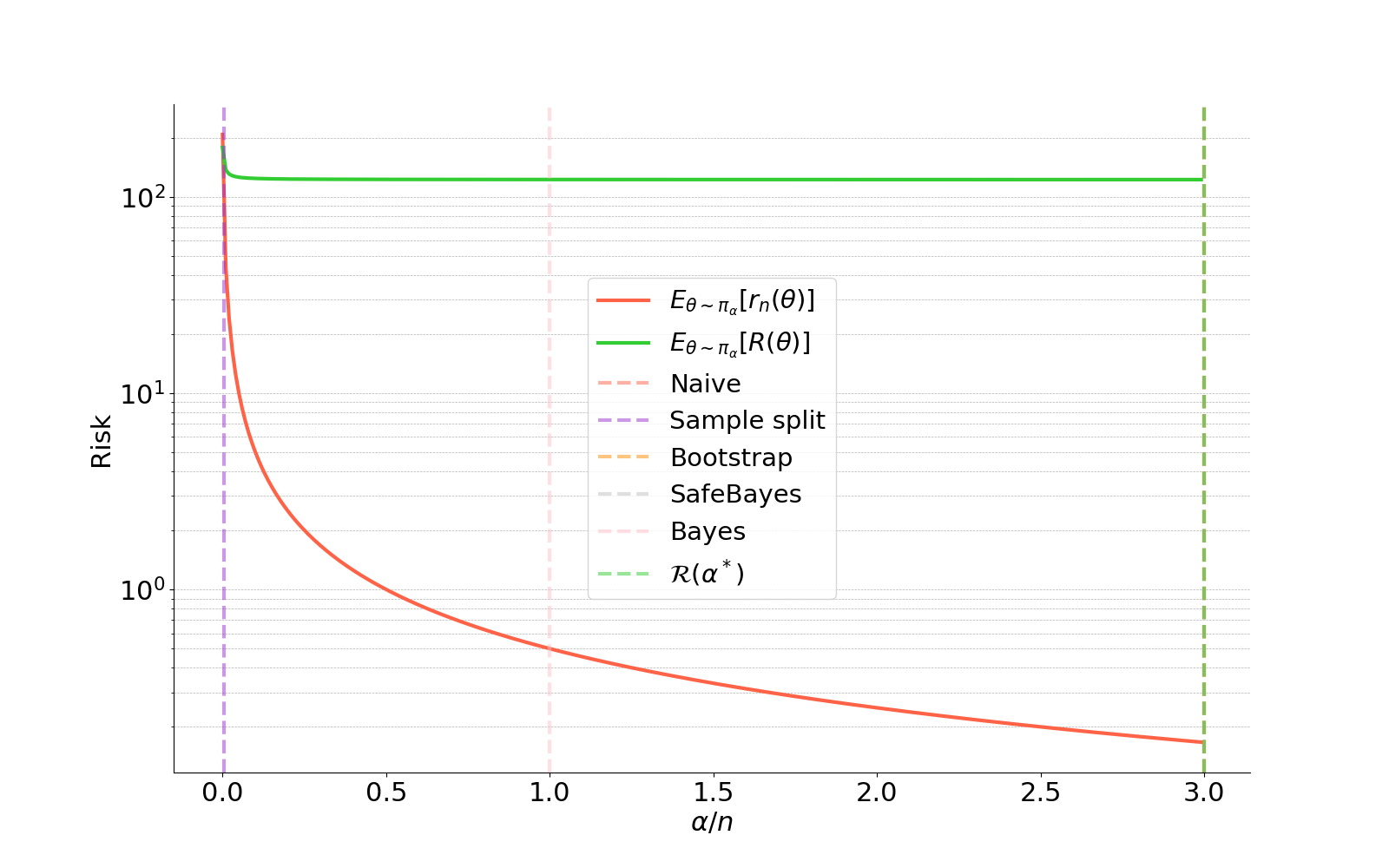}
\end{minipage}
\caption{Low noise and few observations, when $n=16, d=40, \sigma^2=.2$.}
\label{fig:linreg1_additional_plots4}

\end{figure}

In the well specified linear regression case, the proposed strategies perform the better when the number of observations is sufficiently large relatively to the dimension, and when the noise is small compared to the number of observations. We call these values the observations-to-dimension (Otd) ratio, and the observations-to-noise (Otn) ratio. When both ratios are not extreme, the model shows a behaviour where the generalization error is minimized for a value of $\alpha$ slightly smaller or larger than $n$. Example boxplots are shown in figure \ref{fig:linreg1_boxplot}. The sample splitting and SafeBayes strategies are good at estimating the generalization error and give lower risks than Bayes, almost as good as $\mathcal{R}(\alpha)$. Bootstrap generally performs a little bit worse than sample splitting and SafeBayes but still better than Bayes. The naive strategy, performs the worst as it selects the maximum value of $\alpha$ that lays up the "overfitting hill". One also remarks that due to the limited number of observations, the minimal prediction error (plotted on dashed) $\min_{\bm{\theta}} (\bm{\theta}) $ is significantly smaller than the optimal results of any strategy. \\

As a comparison, we also analyze four extreme cases of the well-specified linear regression, where the Otd and Otn ratios are alternatively very small, or large. 
\begin{itemize}
    \item \textbf{Otd large, Otn large.} This is the ideal case of the linear regression. When both ratios are large, the problem becomes very easy as the observations are both accurate and redundant. Overfitting hardly occurs and most strategies will take the maximum value of $\alpha$. The SafeBayes strategy tends to choose small values of $\alpha/n$, being initially designed for finding values of $\alpha/n \in [0,1]$, and may not converge to a large value of $\alpha$ \cite{grnwald2014inconsistency}. The associated risk is, however, very similar to the risk obtained by the other strategies, while all strategies perform very similarly well by achieving a low risk. The generalization and empirical error curves are similar. An example is shown in figure $\ref{fig:linreg1_additional_plots1}$.
    \item \textbf{Otd large, Otn small.} This case is the most similar to the non-extreme case presented in figure \ref{fig:linreg1_boxplot}. When both ratios are high, most strategies perform well as the number of observations achieves compensating the high noise. Overfitting occurs for $\alpha/n$ in the region of $1$, and the achieved risk is relatively high, but close to the minimal prediction error. Both sample splitting and SafeBayes lie close to $\alpha^*$. The large number of observations allows the bootstrapping strategy to perform well by having many observations to draw from. By the law of large numbers, the empirical error does not underestimate the risk and gets close to the generalization error when $\alpha$ increases. One can observe this in figure \ref{fig:linreg1_additional_plots2}.
    \item \textbf{Otd small, Otn small.} When the noise is large and the observations are few, the data carries very little information about the process. The generalization error will typically be minimized in values of $\alpha/n$ tending to $0$ as the ratios become smaller. Hence, choosing values close to $\Omega_P = \Omega_0$ are considered a safe choice by the sample splitting and SafeBayes strategies. Bootstrapping on the other hand performs poorly with the limited number of observations and has a limited set of bootstrap draws to choose from. It hence behaves similarly to the naive strategy. The strategies' risks are overall large, as the minimal prediction error is itself large. The generalization error is similar as in the previous case, however, due to the small number of observations, the empirical error is this time too confident as it underestimates the generalization error. This can be observed in figure \ref{fig:linreg1_additional_plots3}.
    \item \textbf{Otd small, Otn large.} In this last case, the observations are accurate but very limited. The number of observations on which the $\alpha$-posterior is trained is crucial. One can observe that the sample splitting strategy, using only the first half of the observations, performs worse than all the other strategies using the whole batch. However, in such an extreme setup, all strategies perform poorly and entail large risks. Contrarily to the second case, the empirical error is quite low compared to the generalization error and is hence too confident, as the law of large numbers applies to a smaller extent in this case. This is shown in figure \ref{fig:linreg1_additional_plots4}.
\end{itemize}

\paragraph{Gaussian mean estimation}

\begin{figure}[t]
\begin{minipage}{.5\textwidth}
    \centering
    \includegraphics[width=\linewidth]{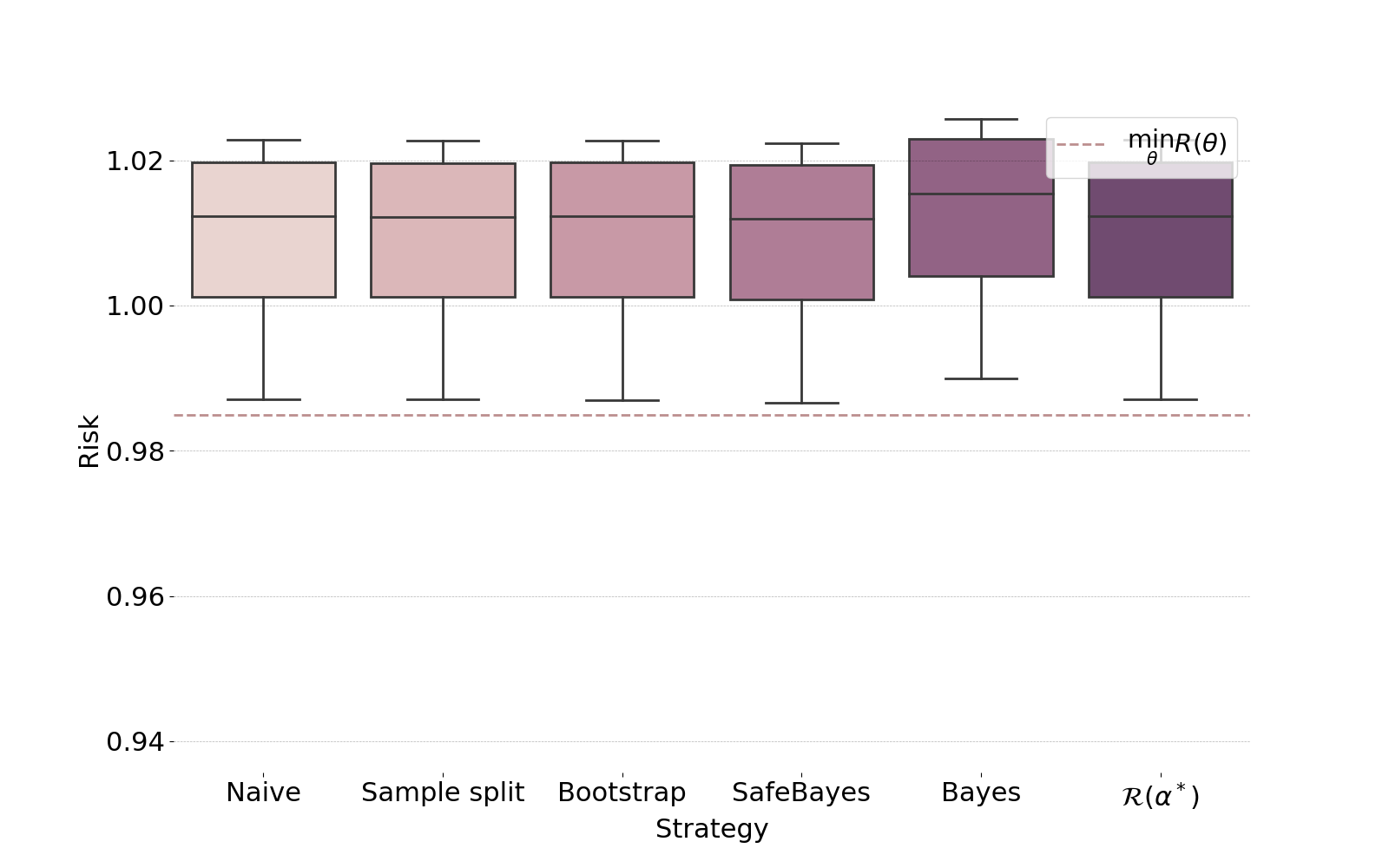}
    \end{minipage}%
\begin{minipage}{.5\textwidth}
    \includegraphics[width=\linewidth]{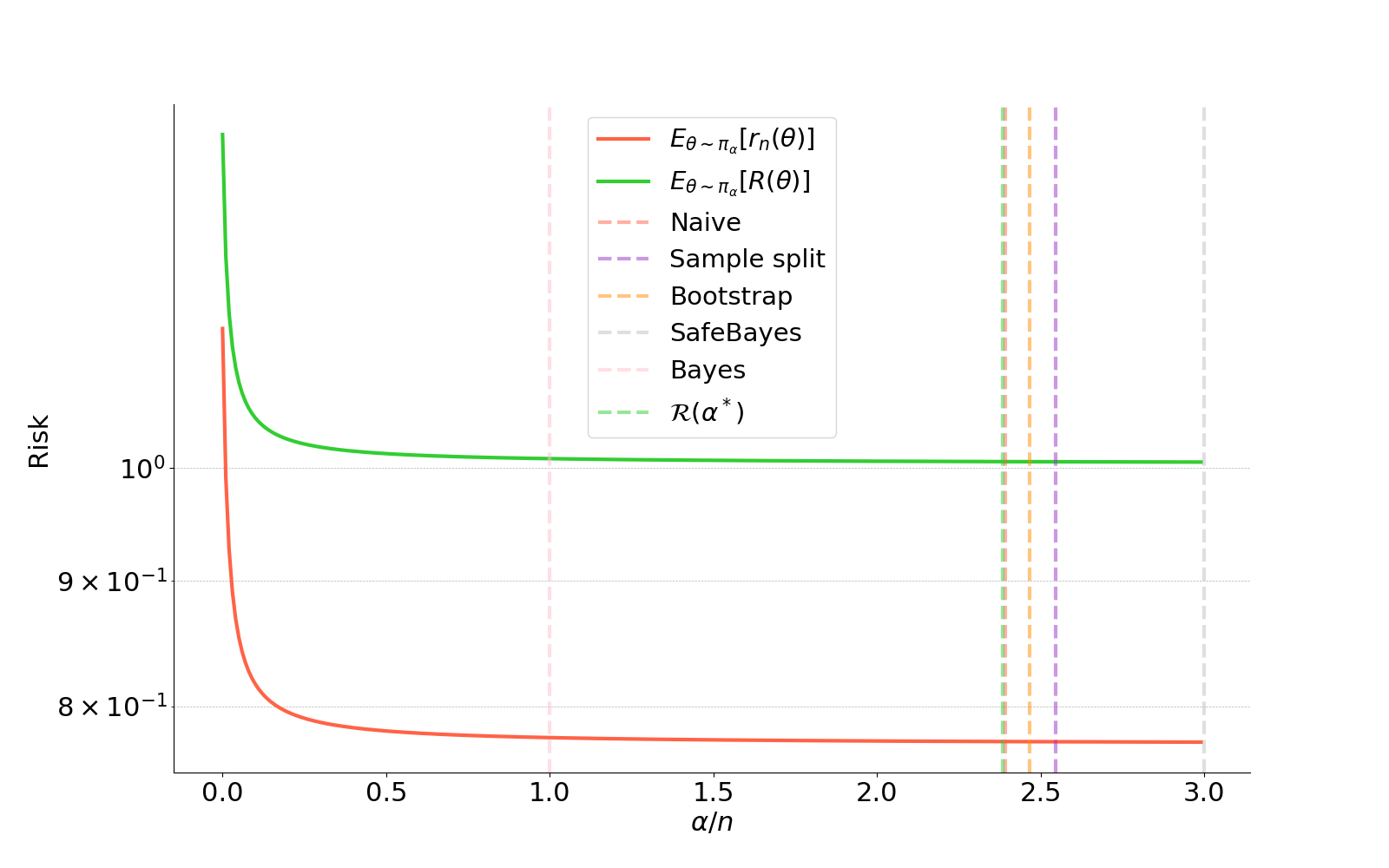}
\end{minipage}
    \caption{Comparison of the strategies for the Gaussian mean estimation when $n=40$ and $\sigma^2=4$. In this simple model, no overfitting occurs and and all strategies, except Bayes, choose a value of $\alpha$ close to the maximum allowed on the right.}
\label{fig:gaussian_plots}
\end{figure}

The Gaussian mean setting has a small fixed dimensionality of only $d=1$, and hence hardly overfits. Most of the time, all the strategies take the maximum allowed value of $\alpha$ on the right and give the same result. A typical behaviour is displayed in figure \ref{fig:gaussian_plots}. Sometimes, a small overfitting may occur after a slightly lower value of $\alpha$ than the maximum, explaining why all the boxplots do not look exactly the same. However, there is not one strategy that achieves significantly lower risk than the others. The minimal prediction error is also closer to the boxplots, indicating a good quality of results in this very simple setup. By dramatically increasing the noise, one would be able to reproduce results similar to figure \ref{fig:linreg1_additional_plots2}.

\paragraph{Polynomial regression}

The third setting we are testing for the linear regression model is the polynomial regression, where the model tries to fit a degree $d$ polynomial curve to $n$ noisy observations from a function $f(\cdot)$. More details about the general polynomial setup are found in appendix \ref{polynomial_setup}. We here choose to focus on a particular case of this setup, where the noise variance is misspecified in the model, which assumes the variance to be too small compared to the actual one. Typically, the actual variance is 10 to 100 times larger than the model's assumed known variance. The model hence tends to believe that the fluctuations in the observed data may be part of the shape of the function $f$ and not due to the noise. This may lead to overfitting. To counteract this phenomenon, we have chosen a prior where the diagonal elements of covariance matrix are decreasing powers of $2$. Thus, the higher the index of the diagonal element, the higher the associated power of the polynomial and the smaller its weight. The choice of $\alpha/n$ hence boils down to tuning the dampening of the high powers of the fitting polynomial.\\

\begin{figure}[ht]
\centering
\makebox[\textwidth]{
\begin{subfigure}{.55\textwidth}
  \centering
  \includegraphics[width=\linewidth]{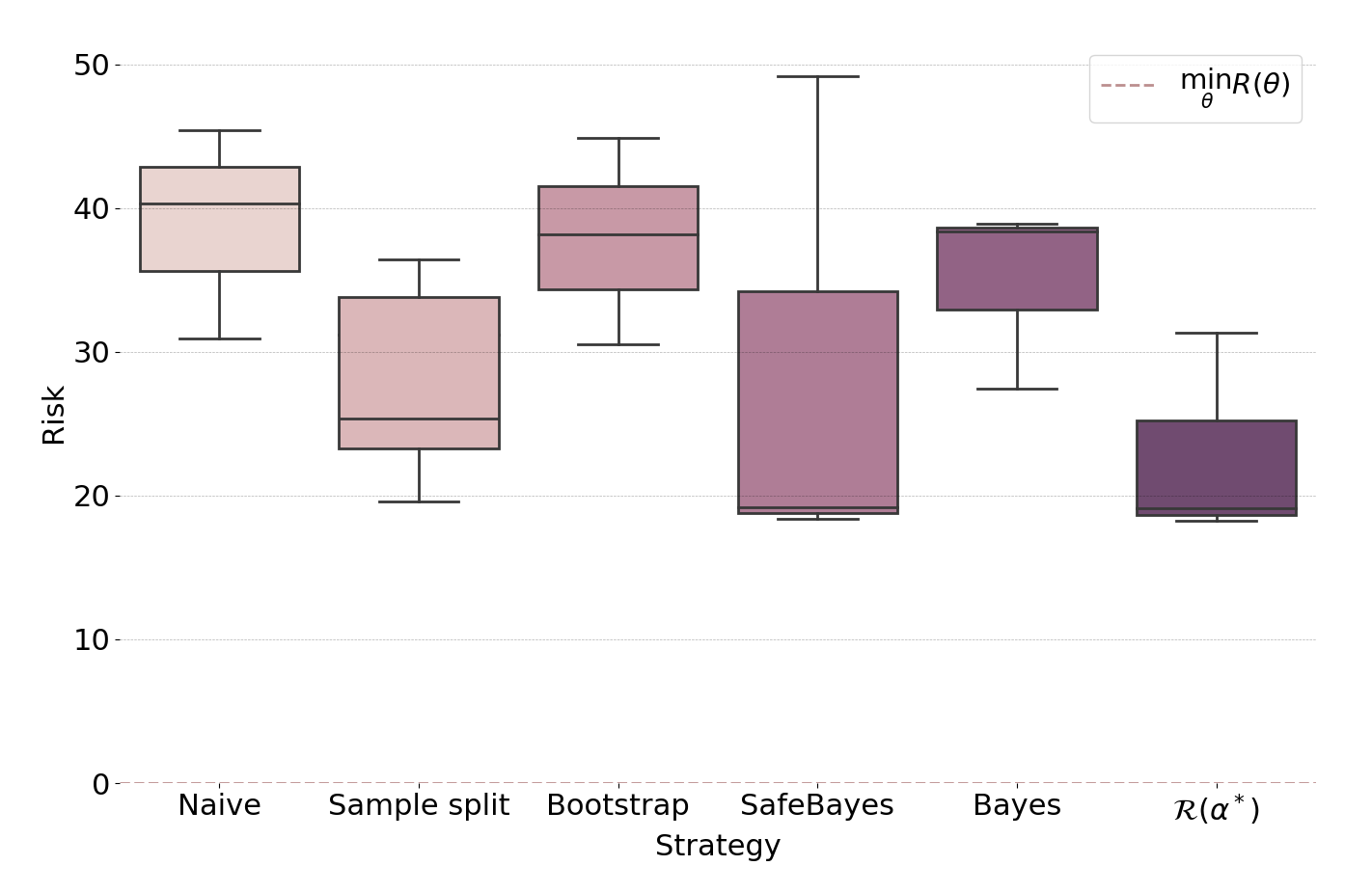}
  \caption{Boxplots of the strategies.}
\end{subfigure}%
\begin{subfigure}{.55\textwidth}
  \centering
  \includegraphics[width=\linewidth]{figures/polyreg1.png}
  \caption{Sample splitting predictive where $\alpha/n \approx 0.1$.}
\end{subfigure}
}
\makebox[\textwidth]{
\begin{subfigure}{.55\textwidth}
  \centering
  \includegraphics[width=\linewidth]{figures/polyreg2.png}
  \caption{Bayes predictive where $\alpha/n=1$.}
\end{subfigure}%
\begin{subfigure}{.55\textwidth}
  \centering
  \includegraphics[width=\linewidth]{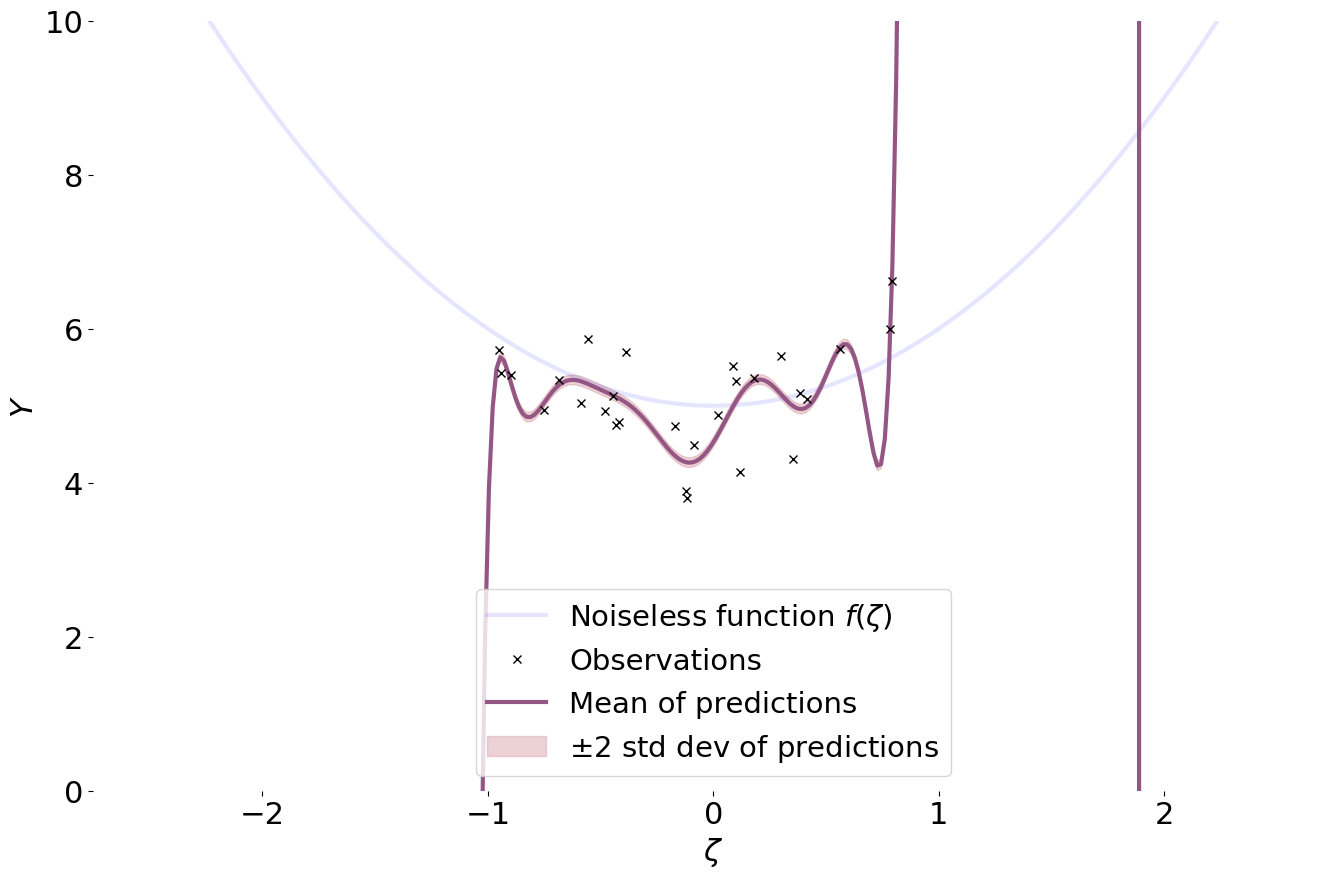}
  \caption{MLE predictive where $\alpha/n \rightarrow \infty$.}
\end{subfigure}
}
\caption{Boxplots and posterior predictive for the polynomial regression over the function $f(\zeta) = \zeta^2+5$, with $n=30$ and $d=12$, for different values of $\alpha/n$. The model assumes $\sigma^2 = 0.01$ yet the actual variance of the noise is $\sigma^2 = 0.5.$}
\label{fig:polyreg_plots}
\end{figure}

The polynomial regression has nice visualization properties since it allows plotting a regression curve with any number of parameters $d$ on two axes only, contrarily to the general linear regression which is limited to $d=2$. As a visualization, we compute the $\alpha$-posterior predictive for a set of $n=30$ observations, that we fit with a degree $d=12$ polynomial curve. The predictive is a Gaussian distribution: its mean represents the average polynomial curve that is fitting to the data and its double standard deviation, the $95\%$ credibility interval, describes the range of uncertainty about the behaviour of the function. We ideally would like the $\alpha$-posterior predictive to be smooth and accurate where observations are available, and uncertain where no observation is available. We compare three values of $\alpha/n$: the sample splitting's choice $\alpha/n=0.1$, the Bayesian choice $\alpha/n=1$, and the MLE value $\alpha/n \rightarrow \infty$. The boxplots of the strategies, as well as the three cited predictive distributions, are plotted in figure \ref{fig:polyreg_plots}. The observations are generated in the interval $\zeta_i \in [-1,1]$ but the predictive distributions are shown for a larger interval $\zeta_i \in [-2.5, 2.5]$ in order to observe their behaviour both in the central area where observations are available, and in the side areas where no observations are available.\\

First, the sample splitting strategy chooses a value of $\alpha/n = 0.1$ which is close to 0, the prior, leading to a smooth function that does not overfit. The predictive mean is close to the actual function $f$ and resembles a parabola. Additionally, the $95\%$ credibility interval shows confidence in the central area where observations are available and shows an accurate uncertainty in the regions on the right and left where no observations are available. The risks achieved by the sample splitting strategy are close to the generalization error's minimum. Second, Bayes chooses a larger $\alpha/n = 1$ and slightly overfits the data. Indeed, it gives less importance to the prior and hence more weight to high powers of the polynomial, allowing more detail in the fitted curve. The predictive mean is yet not such a bad estimate of the function $f$. In contrast, Bayes' predictive is too confident about the function's behaviour in the left area, and displays a small credibility interval. Third, the MLE chooses the maximum value of $\alpha/n$ and totally overfits the data, giving a bad estimate of the function $f$. It is also very confident in its predictions in all areas, as the posterior's variance goes to $0$, and the noise variance $\sigma^2$ is believed to be $0.01$. It is hence a poor predictor of the function $f$.

\subsection{Linear regression with unknown variance}
\subsubsection{Model setup}
We now generalize the linear regression to the case where the variance is assumed unknown. The likelihood is written as 
\begin{align*}
     \mathcal{L}(\bm{\theta}, \sigma^2, \bm{X}) = \left( \frac{1}{\sqrt{2 \pi \sigma^2}} \right)^{|\bm{X}|} \exp \left\{ -\frac{1}{2\sigma^2}  (\bm{Y} - \bm{Z} \bm{\theta})^\top (\bm{Y} - \bm{Z} \bm{\theta}) \right\},
\end{align*}
and the loss function as
\begin{align*}
    \ell (\bm{\theta}, \sigma^2, X_i) = \frac{(Y_i - Z_i^\top \bm{\theta})^2}{2 \sigma^2}  + \frac{|\bm{X}|}{2} \log 2 \pi \sigma^2 .
\end{align*}
The variance of the noise is now itself treated as a random variable, typically chosen to follow an inverse Gamma distribution, while the data is modelized by a Gaussian distribution with this very variance \cite{gelman2013bayesian, susan, murphy_NIG}. We choose a prior of the form
\begin{align*}
    \pi( d \bm{\theta},  d\sigma^2) \sim \text{NIG}(\bm{\mu}_0, \bm{S}_0, a_0, b_0) = \mathcal{N}(\bm{\theta} | \bm{\mu}_0, \bm{S}_0 \cdot \sigma^2) \cdot \Gamma^{-1}(\sigma^2 | a_0, b_0)
\end{align*}
where NIG denotes the Normal-Inverse-Gamma distribution, the combination of an inverse Gamma variance with a Gaussian distribution. Note that this distribution outputs 2 values. We typically choose the following prior values
\begin{align*}
    \bm{\mu}_0 &= \bm{0} \\
    \bm{S}_0 &= \bm{I}_d \\
    a_0 &= 2 \\
    b_0 &= 2 .
\end{align*}
Finally, we can compute a closed-form $\alpha$-posterior by modifying the NIG Bayesian posterior \cite{drugowitsch2013variational, denison2002bayesian} and obtain
\begin{align}\label{NIG-posterior}
     \pi_{\alpha}(d \bm{\theta}, d \sigma^2) &\sim \text{NIG} ( \bm{\mu}_P, \bm{S}_P, a_P, b_P)
\end{align}
where the parameters $\Omega_P$ are
\begin{align*}
         \bm{\mu}_P &= \bm{S}_P \left( \bm{S}_0^{-1} \bm{\mu}_0 + \frac{\alpha}{| \bm{X} |} \bm{Z}^{ \top} \bm{Y} \right) \\
     \bm{S}_P &= \left( \frac{\alpha}{| \bm{X} |} \bm{Z}^{ \top} \bm{Z} + \bm{S}_0^{-1} \right)^{-1} \\
     a_P &= a_0 + \frac{ \alpha }{2} \\
     b_P &= b_0 + \frac{1}{2} \left( \bm{\mu}_0^\top \bm{S}_0^{-1} \bm{\mu}_0 - \bm{\mu}_P^{ \top} \bm{S}_P^{ -1} \bm{\mu}_P + \frac{\alpha}{| \bm{X} |} \bm{Y}^{ \top} \bm{Y} \right).
\end{align*}
Again, when using the sample splitting or SafeBayes strategies, one should use the according data batch $\bm{X}^{(1)}$ or $\bm{X}^{(t)}$ instead of $\bm{X}$. All the derivations can be found in the appendix \ref{appendix_linreg2}.

\subsubsection{Datasets generation}
We generate the data using several settings which are more or less misspecified with the linear regression model:
\begin{enumerate}
\item Well specified case of the linear regression
    \begin{align*}
        Y_i =  Z_i^\top \bm{\theta} + \varepsilon_i, \quad \varepsilon_i \sim \mathcal{N}(0, \sigma^2)
    \end{align*}
    \item Heteroscedatic Gaussian mixture model (GMM) noise
\begin{align*}
     Y_i =  Z_i^\top \bm{\theta} + \varepsilon_i, \quad \varepsilon_i \sim p \cdot \mathcal{N}(0, \delta^2) + (1-p) \cdot \mathcal{N}(0, \sigma^2), \quad \delta << \sigma
\end{align*}
    \item Uniform noise
    \begin{align*}
         Y_i =  Z_i^\top \bm{\theta} + \varepsilon_i, \quad \varepsilon_i \sim \mathcal{U}(-1, 1)
    \end{align*}
\end{enumerate}

\subsubsection{Strategies performances}

\begin{figure}[ht]
\centering
\begin{subfigure}{.5\textwidth}
  \centering
  \includegraphics[width=\linewidth]{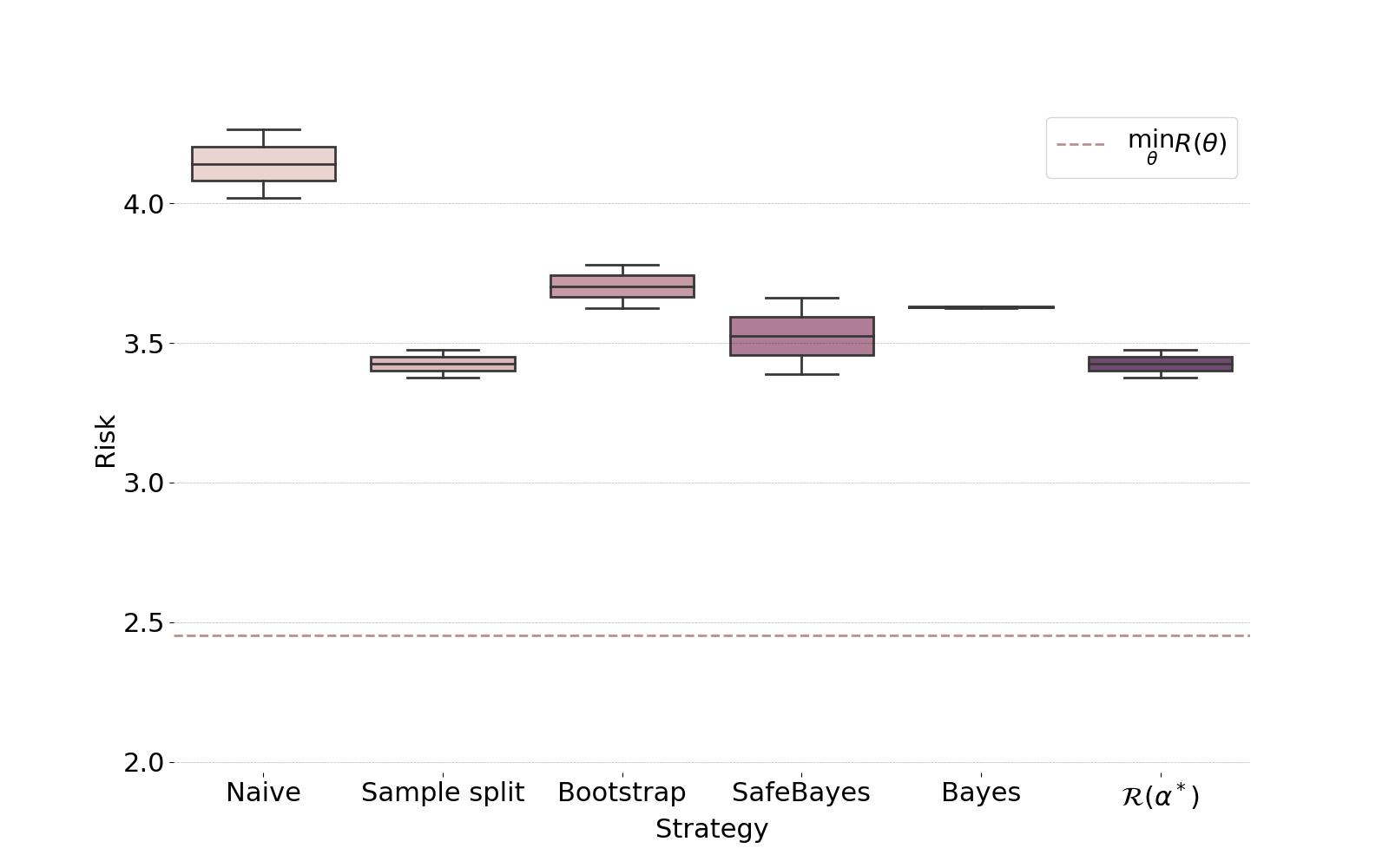}
  \caption{Well specified noise.}
\end{subfigure}%
\begin{subfigure}{.5\textwidth}
  \centering
  \includegraphics[width=\linewidth]{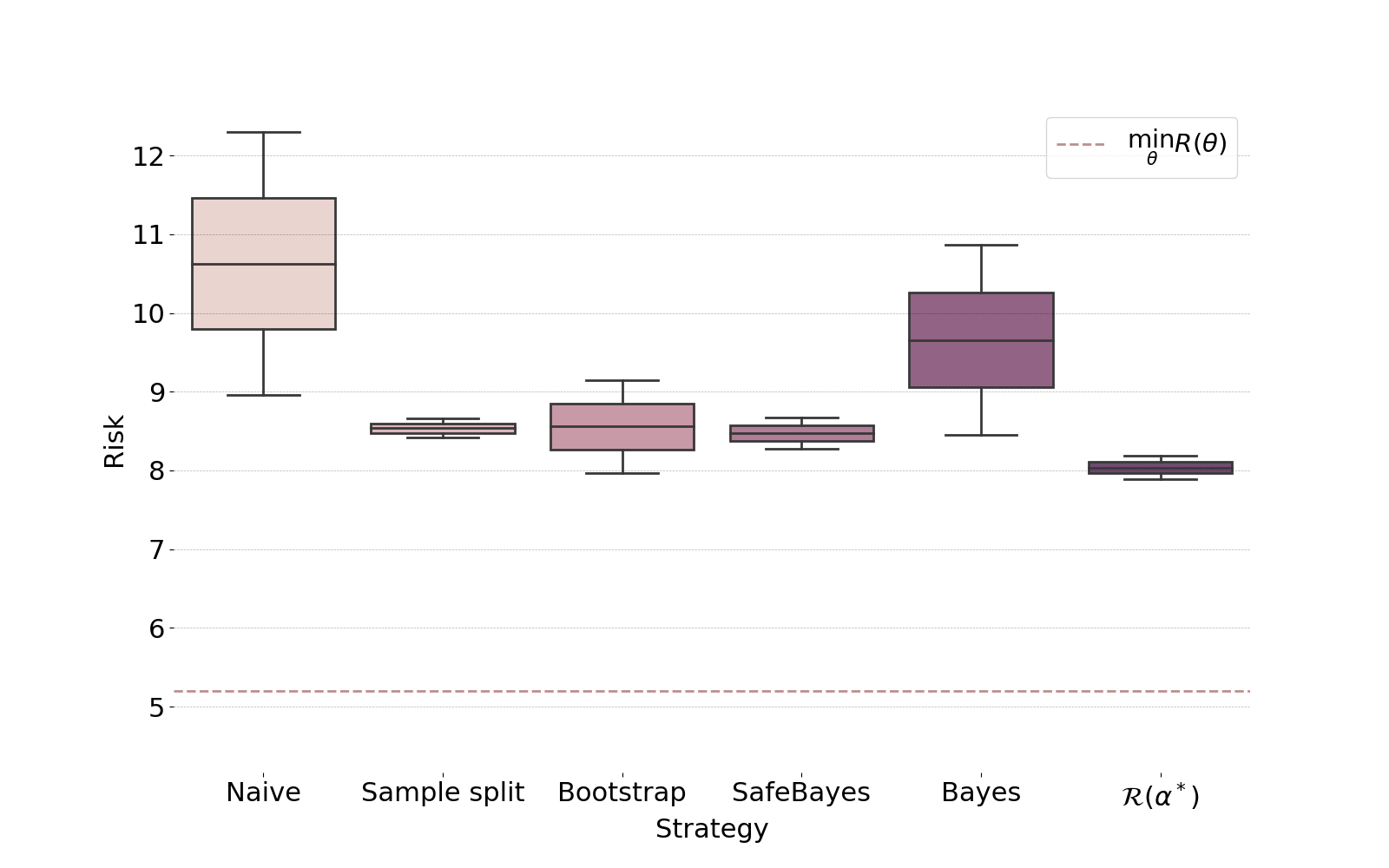}
\caption{GMM noise.}
\end{subfigure}

\begin{subfigure}{.5\textwidth}
  \centering
  \includegraphics[width=\linewidth]{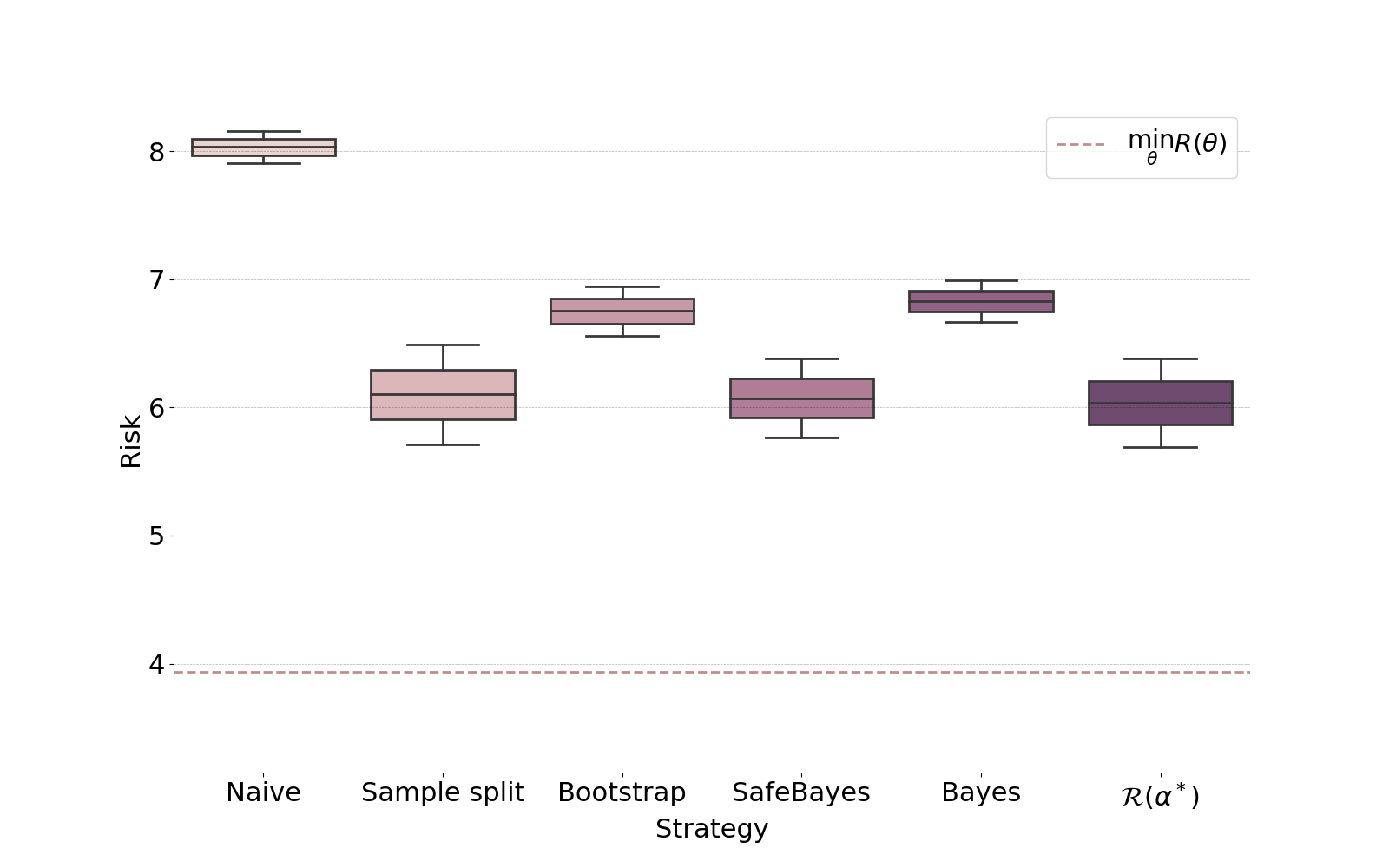}
\caption{Uniform noise.}
    \end{subfigure}%
\caption{Boxplots for different settings, where $n=60$ and $d=40$.}
\label{linreg2_all_graphs}
\end{figure}

Overall, the linear regression with unknown variance displays a similar behaviour than the linear regression with known variance, which steeper results when the misspecification becomes more important. Boxplots for the three settings are displayed in figure \ref{linreg2_all_graphs}. \\

\paragraph{Well specified variance} The first case is the well specified case, where the model makes correct assumptions about how the data is generated. The variance is here $\sigma^2=3.5$ but is not communicated to the model. In this case, the strategies have a similar behaviour as in the general linear regression with known variance. When enough observations are available compared to the noise of the variance and the number of dimensions, all strategies perform similarly well. When the noise or the number of dimensions increase, Bayes performs worse and the sample splitting and SafeBayes strategies should be preferred.

\paragraph{GMM noise variance} The second case, GMM noise variance, is misspecified in the sense that the model assumes a homoscedastic noise variance whereas it is heteroscedastic with $\sigma^2=12$, $\delta^2=0.1$, and $p=0.5$. The distribution of the data however, is correctly assumed to be Gaussian. The model makes no assumption about the size of the variance, being hence flexible against the changing variance size, and the performances of the strategies are hence not really worse than the well specified case when using $\frac{\sigma^2+\delta^2}{2}$ as the variance. We observe a better performance of the sample splitting, bootstrapping and SafeBayes strategies over standard Bayes.

\paragraph{Uniform noise} The third case, uniform noise, is this time misspecified about the distribution of the variance itself. Indeed, the model assumes the noise to be Gaussian, while it is uniform in $[-4.5, 4.5]$. The variance of the uniform distribution is $6.75$, and hence the boxplots are located in the region of this value. The misspecification being more important in this setting, the performances of the strategies are also more contrasted, with sample splitting and SafeBayes performing almost as good as $\mathcal{R}(\alpha^*)$ and sensibly better than Bayes. The naive strategy instead performs notably worse than the other strategies.  

\subsection{Logistic regression}
So far, most related works on $\alpha$-posterior have focused on models where the posterior is known exactly. Recent publications like \cite{alquier2017concentration, yang2017variational} extended the research to models where an exact $\alpha$-posterior is not available, and a variational approximation is used instead. The logistic regression is such a model, as neither the posterior nor the generalization error are available in closed-form. We now aim to verify if our strategies, and more specifically, the proposition \eqref{proposition_covariance} still give sensible results. We compute $\Tilde{\alpha}$ which minimizes the estimate of the generalization error in each strategy, using the proposition as if an exact $\alpha$-posterior was used. Mathematically, we have no guarantee that
\begin{align*}
    \Tilde{\alpha} \stackrel{?}{=} \alpha^* = \argmin_\alpha \mathcal{R}(\alpha) .
\end{align*}
However, we suggest that a good enough variational approximation of the $\alpha$-posterior may work with the proposition. Hence, rather than exploring different settings as we did for the previous models, we rather compare two variational Gaussian models on similar well-defined settings with a limited number of observations, and verify how well the strategies perform when using the proposition. \\

First, the logistic likelihood with outputs $Y_i \in \{0, 1 \}$ is written as 
\begin{align*}
\mathcal{L}(\bm{\theta}, \bm{X}) = P(Y_i = y | Z_i, \bm{\theta}) &= \sigma( \bm{\theta}^\top Z_i )^y \cdot  \{ 1 - \sigma( \bm{\theta}^\top Z_i) \}^{(1-y)} \\
&= e^{ \bm{\theta}^\top Z_i y } \sigma ( -\bm{\theta}^\top Z_i )
\end{align*}
where $\sigma(a) = \frac{1}{1 + e^{-a}}$ is the logistic sigmoid function. The loss function becomes
\begin{align*}
    \ell (\bm{\theta}, X_i) &= - Y_i \log \left( \sigma \left(\bm{\theta}^\top Z_i \right)  \right) -  \left(1- Y_i \right) \log  \left(1 - \sigma  \left(\bm{\theta}^\top Z_i \right) \right) \\
    &= - \bm{\theta}^\top Z_i Y_i - \log  \left( \sigma \left(- \bm{\theta}^\top Z_i \right) \right).
\end{align*}
We now describe the two models independently, and then compare them on logistic datasets.

\subsubsection{Jaakkola model setup}
This model was proposed in 1996 and was better described in 2001 by Jaakkola and Jordan in \cite{jaakkola2001} as a closed-form variational Gaussian posterior, whose parameters can be optimized using an expectation-maximization (EM) algorithm. We choose the prior to be Gaussian
\begin{align*}
    \pi( d \bm{\theta}) \sim \mathcal{N}(\bm{\mu}_0, \bm{S}_0 ) ,
\end{align*}
with typical hyperparameters values being $\bm{\mu}_0 = \bm{0}$ and $\bm{S}_0 = \bm{I}_d$. We modify the closed-form Jaakkola variational approximation of the posterior into an $\alpha$-variational posterior as
\begin{align}\label{jaakkola_posterior}
    \pi_\alpha (d \bm{\theta}) \sim \mathcal{N}( \bm{\theta} | \bm{\mu}_P, \bm{S}_P)
\end{align}
where the parameters $\Omega_P$ are
\begin{align*}
    \bm{\mu}_P &= \bm{S}_P \left( \bm{S}_0^{-1} \bm{\mu}_0 +  \frac{\alpha}{| \bm{X} |} \sum_{i=1}^{n} \left( Y_i - \frac{1}{2} \right) Z_i \right) \\
    \bm{S}_P &= \left( \bm{S}_0^{-1} + 2  \frac{\alpha}{| \bm{X} |} \sum_{i=1}^{n} \lambda(v_i) Z_i Z_i^\top \right)^{-1}
\end{align*}
and 
\begin{align*}
    \lambda(v_i) &= \frac{1}{2 v_i} \left[ \sigma(v_i) - \frac{1}{2} \right]  \\
     v_i &= \left( Z_i^\top (\bm{S}_P + \bm{\mu}_P \bm{\mu}_P^{ \top}) Z_i \right)^{1/2} .
\end{align*}
We then alternatively compute the value of $\lambda(v_i)$, and update the values of the parameters $\Omega_P$, where initial values of $\lambda(v_i)$ are set randomly. This optimization can alternatively be seen as a variational coordinate descent of the parameters. All the detailed computations can be found in appendix \ref{jaakkola-appendix}. As usual, one should replace the data batches according to the strategy used.

\subsubsection{Bayes by Backprop model setup}
This second model is based on Black-box variational inference and was more recently introduced in \cite{deepmind2015weight}. This model is called \textit{Bayes by Backprop}. It proposes a reparametrization of the parameter $\bm{\theta}$ as well as a closed-form expression for the gradient of the ELBO. As a result, we obtain stochastic update rules for the posterior's parameters $\Omega_P$.\\

The prior is chosen to be a normalized Gaussian
\begin{align*}
    \pi_0 ( d \bm{\theta}) \sim \mathcal{N}(\bm{0}, \bm{I}_d),
\end{align*}
since this model becomes a lot more complex for general values of the prior's hyperparameters. Using the proposition described in their method, we obtain a closed-form expression proportional to the negative ELBO and then use Autograd \cite{autograd2015} to compute its gradients with respect to the posterior's parameters. The gradients are finally used in a SGD algorithm to optimize the parameters $\Omega_P$ in turn. Hence, we obtain a Gaussian $\alpha$-posterior
\begin{align}\label{SVI_posterior}
    \pi_\alpha (d \bm{\theta}) \sim \mathcal{N}( \bm{\theta} | \bm{\mu}_P, \bm{S}_P)
\end{align}
where no closed-form is available for the parameters $\Omega_P$, that are updated with SGD. The detailed computations can be found in appendix \ref{SVI-appendix}.

\subsubsection{Strategies performances comparison on the models}

We compare the two models using the same well specified dataset, where $n=50$ and $d=30$. The Bayes by Backprop model gives a more accurate Gaussian $\alpha$-posterior than the Jaakkola model. Indeed, \cite{SVI_better} show that the Jaakkola model is a biased estimate and performs worse than unbiased models when a formula for the gradient is available. Bayes by Backprop is an unbiased estimate, which uses a reparametrization trick to improve the performance of the gradient estimate. In our case, we use the proposition \eqref{proposition_covariance} as the gradient formula, even though it is not exact with variational posteriors. The risks are lower for the Bayes by Backprop model than the Jaakkola model, confirming what has been shown in the above paper. The computations however are much slower as they require at least 200 iterations for the SGD to converge whereas the Jaakkola model only typically needs 5 iterations. \\

\begin{figure}[ht]
\centering
    \begin{minipage}{.5\textwidth}
  \centering
  \includegraphics[width=\linewidth]{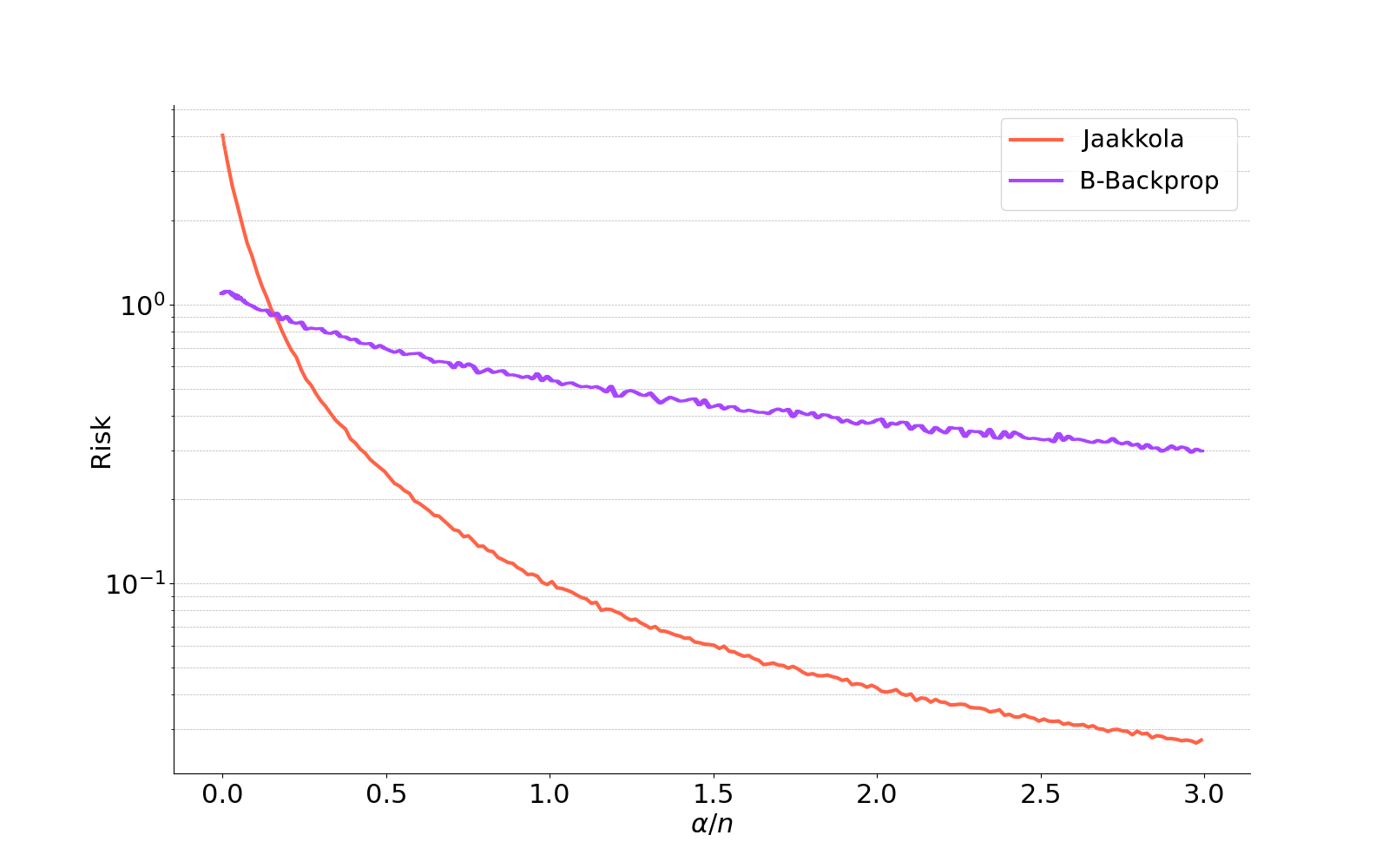}
\end{minipage}%
\begin{minipage}{.5\textwidth}
  \centering
  \includegraphics[width=\linewidth]{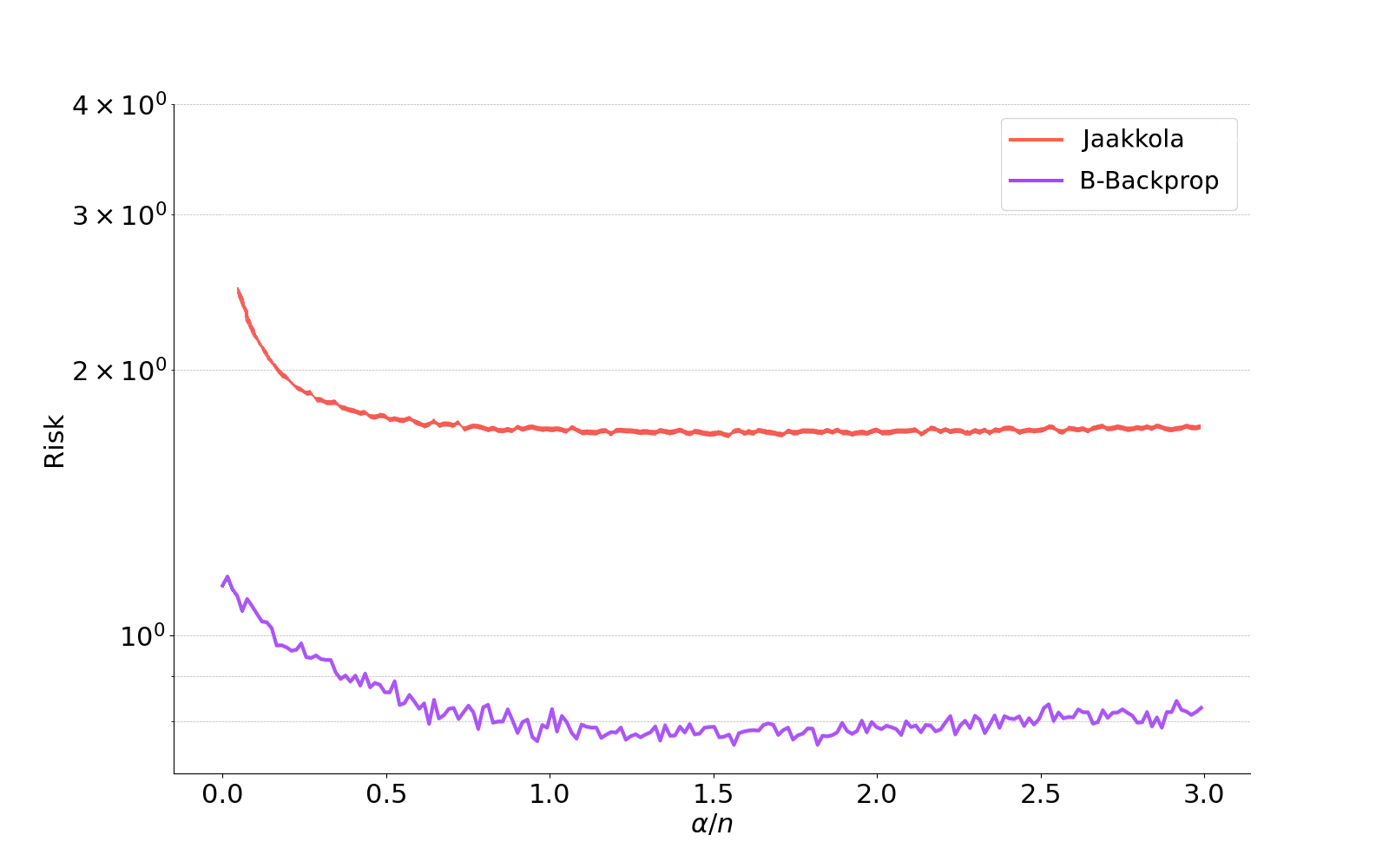}
\end{minipage}
\caption{Comparison of the empirical (left) and generalization (right) error curves, between the Jaakkola and the Bayes by Backprop models, when given the same dataset with $n=50$ and $d=30$. One can notice noise in the curves due to MC approximation (2000 samples were used).}
\label{fig:SVI_vs_EM}
\end{figure}

Figure \ref{fig:SVI_vs_EM} compares the empirical and generalization errors of the two models using the same dataset. One can observe that the Jaakkola $\alpha$-posterior obtains a lower empirical error than the Bayes by Backprop $\alpha$-posterior, which is more conscious about its uncertainty. However, the Jaakkola $\alpha$-posterior shows a higher generalization error than the Bayes by Backprop $\alpha$-posterior, making the Bayes by Backprop a more accurate $\alpha$-posterior. Additionally, Bayes by Backprop shows a distinct minimum for the generalization error, whereas Jaakkola's is very flat and less informative. Note however that both the Jaakkola's and Bayes by Backprop generalization errors are minimized by a similar value of $\alpha^*/n \approx 1.7$, and are hence consistent with each other. In summary, although the two models have a similar $\alpha^*$, Bayes by Backprop is preferred as it achieves a lower generalization error.\\

\begin{figure}[ht]
    \centering
    \begin{minipage}{.5\textwidth}
  \centering
  \includegraphics[width=\linewidth]{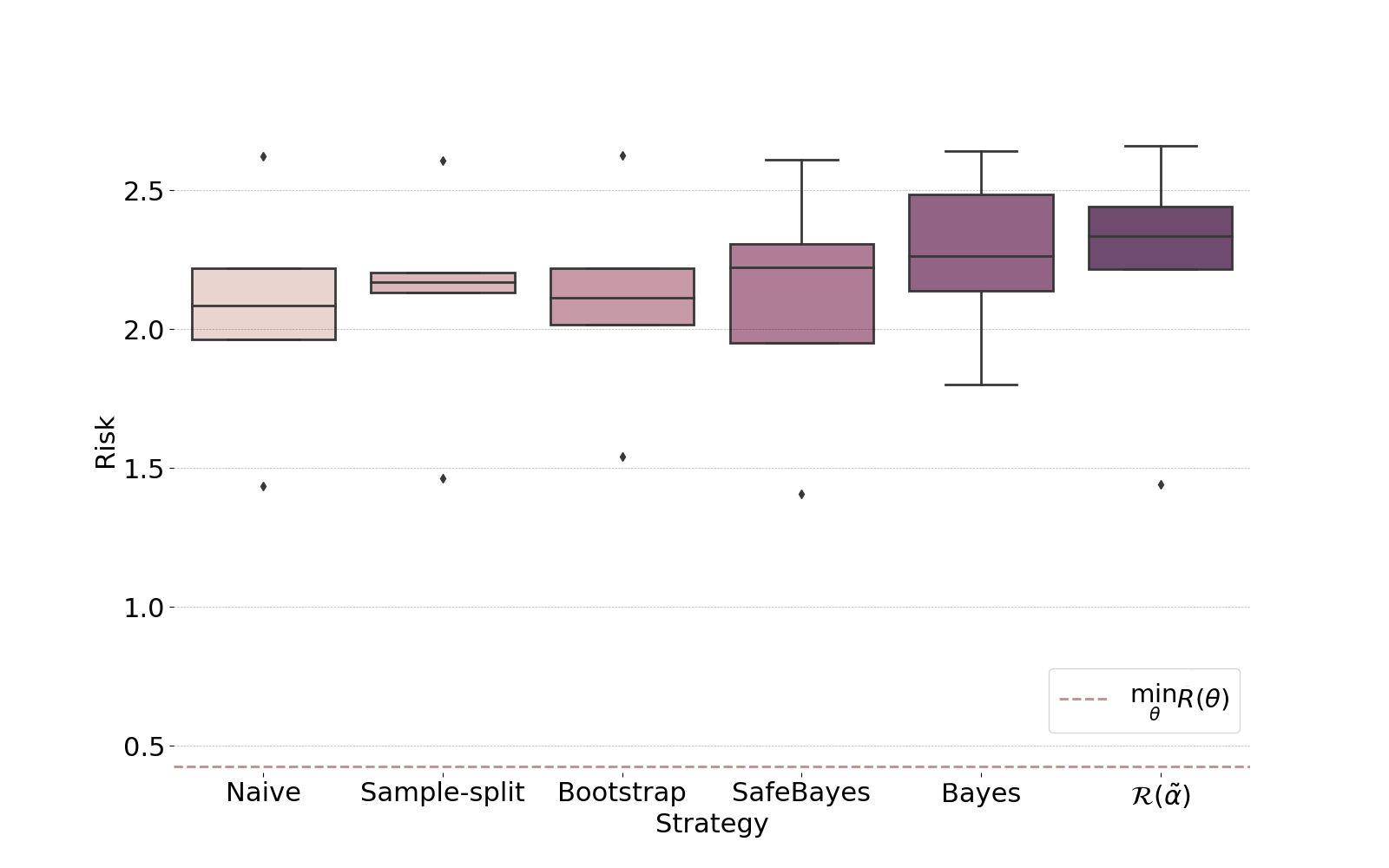}
\end{minipage}%
\begin{minipage}{.5\textwidth}
  \centering
  \includegraphics[width=\linewidth]{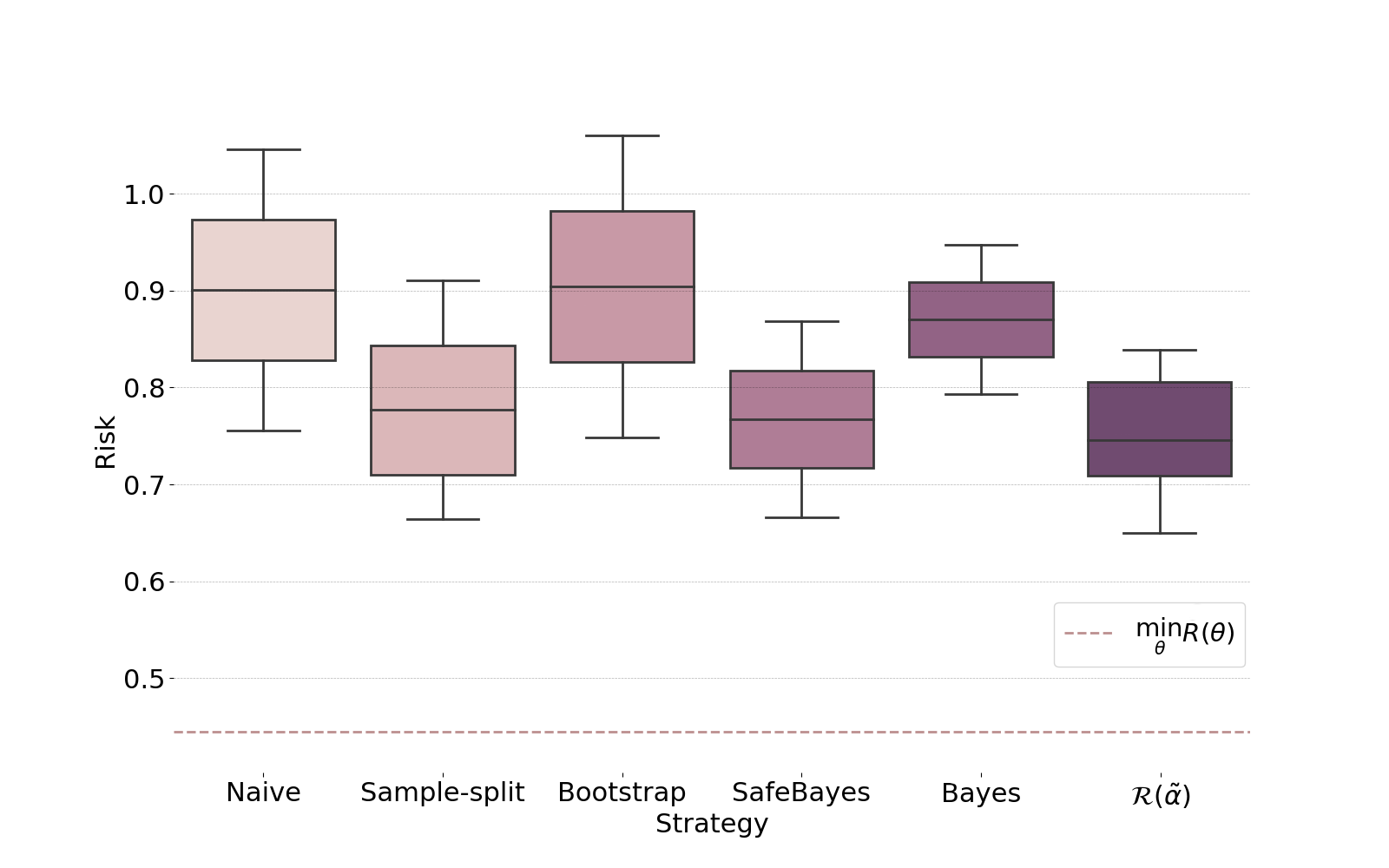}
\end{minipage}
    \caption{Example boxplots for the Jaakkola (left) and the Bayes by Backprop (right) models, when $n=50, d=30$. In the Jaakkola model, the strategies perform better than $\mathcal{R}(\Tilde{\alpha})$, which makes $\Tilde{\alpha}$ a bad estimator of $\alpha^*$. Contrarily, the Bayes by Backprop model shows sensible results.}
    \label{fig:logreg_boxplots}
\end{figure}

We now use \eqref{proposition_covariance} as our gradient formula and compute $\Tilde{\alpha}$, as well as each strategy's $\alpha$ on both models. In the Jaakkola model, the proposition does not achieve minimizing the generalization error and produces an $\Tilde{\alpha}$ that is not a good estimate of $\alpha^*$. One can observe this in the left plot of figure \ref{fig:logreg_boxplots}, where the value $\mathcal{R}(\Tilde{\alpha})$ is higher than the strategies' returned $\alpha$s. This would be a contradiction if $\Tilde{\alpha}$ was a good estimate of $\alpha^*$. Similarly, the strategies' $\alpha$ are not good estimates of $\alpha^*$. Alternative optimizers such as grid search can be applied to minimize the strategy risks, but run very slowly and still are lower bounded by the minimal prediction error, which is higher than Bayes by Backprop. It is hence not interesting to dig into this model, and we instead focus on the Bayes by Backprop model.\\

The Bayes by Backprop model empirically shows sensible results when the proposition is used as the gradient. The sample splitting and SafeBayes strategies achieve the lowest risks, meanwhile the bootstrapping, Bayes, and naive strategies are less effective. Unlike the Jaakkola model, the strategies' risks are close to the minimal prediction error. One additionally remarks that the minimal prediction error is the same in both plots, as it depends on the data only and not the model. The strategies can hence be effective on the Bayes by Backprop approximation in the logistic regression model.

\newpage
\thispagestyle{empty} 
\mbox{}

%% file: chapters/conclusion.tex
\section{Conclusion}
As presented in this work, we have explored a solution to handle model inconsistency and to make better predictions about future observations of a process than the standard Bayesian posterior. To that end, we have proposed two new strategies, sample splitting and bootstrapping, for tempering the PAC-Bayesian $\alpha$-posterior. We analyzed three statistical models, where for each model, we derived a closed-form distribution for the $\alpha$-posterior, adapted from the Bayesian posterior. We also proposed a closed-form formula for the gradient of the estimate of the generalization error which we used to optimize the strategies with SGD. We then tested our strategies on each model and compared their performance with standard Bayes, naive, and SafeBayes. \\

Our first strategy, the sample splitting, runs fast, achieves good results on both the exact and variational $\alpha$-posteriors we tested and outperforms standard Bayes in complex and misspecified models. Our second strategy, bootstrapping, runs slower and achieves more mixed results and works well only when a large number of observations is available. When the model is simple or well specified, all the strategies are quite equivalent, although Bayes is faster to compute. It is worth using the sample splitting and SafeBayes strategies when the model is known to be complex or misspecified, and to use regular Bayes in other cases. The bootstrapping strategy is not worth using. \\

We should finally say a word on comparison between SafeBayes and sample splitting. On all our experiments, these two strategies perform relatively similarly, and sample splitting is much faster to compute. Still, SafeBayes comes with theoretical guarantees \cite{grnwald2016safe} that are not yet established for sample splitting. We believe that the investigations on theoretical guarantees for sample splitting and further comparison of these two strategies is a very important topic.\\

Further work on the subject includes analyzing the performance of the strategies on larger real datasets, as well as on more complex variational models, such as Bayesian Neural Networks as in \cite{NIPS2019_8681}, or classification models.

\newpage
\thispagestyle{empty} 
\mbox{}

%% file: chapters/appendix.tex
\appendix
\section{Proofs}

\subsection{Proof of proposition \eqref{proposition_covariance}}\label{proof_of_covariance}
Define 
$$ \mathcal{\hat{R}}(\alpha) := \mathbb{E}_{\bm{\theta}\sim \pi_\alpha^{(1)}} \left[ r_n^{(2)}(\bm{\theta}) \right], $$ then for the exact $\alpha$-posterior $\pi_\alpha^{(1)}$, we have
$$\frac{{\partial} }{{\partial}\alpha} \mathbb{E}_{\bm{\theta}\sim \pi_\alpha^{(1)}} \left[ r_n^{(2)}(\bm{\theta}) \right] = -\text{Cov}_{\bm{\theta}\sim \pi_\alpha^{(1)}}[r_n^{(1)}(\bm{\theta}),r_n^{(2)}(\bm{\theta}) ] . $$

\begin{proof}
Let us first remind that the $\alpha$-posterior can be written as
\begin{align}
     \pi_\alpha^{(1)} ( d \bm{\theta}) = \frac{ \exp \left[ -\alpha r_n^{(1)} (\bm{\theta}) \right] \pi_0 (\bm{\theta}) }{ \int \exp \left[ -\alpha r_n^{(1)} (\bm{\tau}) \right] \pi_0 (\bm{\tau}) d\bm{\tau} } \label{posterior_rewrite}.
\end{align}
Note that the variables $\bm{\theta}$ and $\bm{\tau}$ can be used interchangeably and are only meant to help with the readability. We have
\begin{align}
    \mathcal{\hat{R}}( \alpha) &:= \mathbb{E}_{\bm{\theta}\sim \pi_\alpha^{(1)}}[r_n^{(2)}(\bm{\theta})] \nonumber \\
    &= \int r_n^{(2)} (\bm{\theta}) \pi_\alpha^{(1)} d \bm{\theta} \nonumber \\
    &\stackrel{\eqref{posterior_rewrite}}{=}  \int r_n^{(2)} (\bm{\theta}) \frac{ \exp \left[ -\alpha r_n^{(1)} (\bm{\theta}) \right] \pi_0 (\bm{\theta}) }{ \int \exp \left[ -\alpha r_n^{(1)} (\bm{\tau}) \right] \pi_0 (\bm{\tau}) d\bm{\tau} } d \bm{\theta} \nonumber \\
    &=  \frac{ \int r_n^{(2)} (\bm{\theta}) \exp \left[ -\alpha r_n^{(1)} (\bm{\theta}) \right] \pi_0 (\bm{\theta}) d \bm{\theta} }{ \int \exp \left[ -\alpha r_n^{(1)} (\bm{\tau}) \right] \pi_0 (\bm{\tau}) d\bm{\tau} } := \frac{F(\alpha)}{G(\alpha)} . \label{d_a_posterior}
\end{align}

We then use Leibniz's integral rule to compute the derivatives wrt. $\alpha$ of both terms in the fraction, giving
\begin{align}
    G'(\alpha) &= \frac{\partial}{\partial \alpha}  \int \exp \left[ -\alpha r_n^{(1)} (\bm{\tau}) \right] \pi_0 (\bm{\tau}) d\bm{\tau} \nonumber \\
    &= \int -r_n^{(1)} (\bm{\tau})  \exp \left[ -\alpha r_n^{(1)} (\bm{\tau}) \right] \pi_0 (\bm{\tau}) d\bm{\tau} \label{top_derivative}
\end{align}
and
\begin{align}
    F'(\alpha) &= \frac{\partial}{\partial \alpha}  \int r_n^{(2)} (\bm{\theta}) \exp \left[ -\alpha r_n^{(1)} (\bm{\theta}) \right] \pi_0 (\bm{\theta}) d\bm{\theta} \nonumber \\
    &= \int -r_n^{(1)} (\bm{\theta}) r_n^{(2)} (\bm{\theta})  \exp \left[ -\alpha r_n^{(1)} (\bm{\theta}) \right] \pi_0 (\bm{\theta}) d\bm{\theta} . \label{bottom_derivative}
\end{align}

We can then compute the derivative of \eqref{d_a_posterior} by substituting in the results of \eqref{top_derivative} and \eqref{bottom_derivative} to obtain
\begin{align}
    \frac{\partial}{\partial \alpha} \mathcal{\hat{R}}(\alpha) &= \frac{F'(\alpha) G(\alpha) - F(\alpha) G'(\alpha)}{G^2(\alpha)} \nonumber \\ 
    &= \frac{ \int -r_n^{(1)} (\bm{\theta}) r_n^{(2)} (\bm{\theta})  \exp \left[ -\alpha r_n^{(1)} (\bm{\theta}) \right] \pi_0 (\bm{\theta}) d\bm{\theta} \cdot \int \exp \left[ -\alpha r_n^{(1)} (\bm{\tau}) \right] \pi_0 (\bm{\tau}) d\bm{\tau}}{ \left( \int \exp \left[ -\alpha r_n^{(1)} (\bm{\tau}) \right] \pi_0 (\bm{\tau}) d\bm{\tau}  \right)^2 } \nonumber \\
    &\qquad \qquad + \frac{\int r_n^{(2)} (\bm{\theta}) \exp \left[ -\alpha r_n^{(1)} (\bm{\theta}) \right] \pi_0 (\bm{\theta}) d \bm{\theta} \cdot \int r_n^{(1)} (\bm{\tau})  \exp \left[ -\alpha r_n^{(1)} (\bm{\tau}) \right] \pi_0 (\bm{\tau}) d\bm{\tau} }{ \left( \int \exp \left[ -\alpha r_n^{(1)} (\bm{\tau}) \right] \pi_0 (\bm{\tau}) d\bm{\tau}  \right)^2 } \nonumber \\
    &\stackrel{\eqref{posterior_rewrite}}{=} - \int r_n^{(1)} (\bm{\theta}) r_n^{(2)} (\bm{\theta}) \pi_\alpha^{(1)} (\bm{\theta}) d\bm{\theta} + \int r_n^{(2)} (\bm{\theta}) \pi_\alpha^{(1)} (\bm{\theta}) d \bm{\theta} \cdot \int r_n^{(1)} (\bm{\tau})   \pi_\alpha^{(1)} (\bm{\tau}) d\bm{\tau} \nonumber \\
    &= - \mathbb{E}_{\bm{\theta}\sim \pi_\alpha^{(1)}}[ r_n^{(1)}(\bm{\theta}) r_n^{(2)}(\bm{\theta}) ] + \mathbb{E}_{\bm{\theta}\sim \pi_\alpha^{(1)}}[ r_n^{(1)}(\bm{\theta}) ] \cdot \mathbb{E}_{\bm{\theta}\sim \pi_\alpha^{(1)}}[ r_n^{(2)}(\bm{\theta}) ] \nonumber \\
    &= - \text{Cov}_{\bm{\theta}\sim \pi_\alpha^{(1)}}[r_n^{(1)}(\bm{\theta}) , r_n^{(2)}(\bm{\theta}) ]
\end{align}
where again, we have used the rewriting of the posterior from \eqref{posterior_rewrite}.
\end{proof}

\section{Derivations for the linear regression with known variance}\label{appendix_linreg1}
Here, the computations pertain to the sample splitting strategy. When using another strategy, one should replace the data batches accordingly.

\subsection{Derivation of the $\alpha$-posterior \eqref{linear_regression_posterior_gaussian} }\label{proof_of_posterior_linreg1}
We first compute the empirical error as
\begin{align*}
 r_n^{(1)}(\bm{\theta}) & = \frac{1}{|\bm{X}^{(1)}|} \sum_{i=1}^{n/2} \ell (\bm{\theta}, X_i)\\
    &= \frac{1}{2 \sigma^2 |\bm{X}^{(1)}|} \sum_{i=1}^{n/2} (Y_i - Z_i^\top \bm{\theta} )^2 \\
 r_n^{(2)}(\bm{\theta}) &= \frac{1}{2 \sigma^2 |\bm{X}^{(2)}|} \sum_{i=1+n/2}^{n} (Y_i - Z_i^\top \bm{\theta} )^2 ,
\end{align*}
and using the conjugate property of the prior, the $\alpha$-posterior becomes
\begin{align*}
    \pi_{\alpha}^{(1)} ( d \bm{\theta}) &\propto \exp\left[-\alpha r_n^{(1)}(\bm{\theta}) \right] \pi( d \bm{\theta} ) \\
     &\propto  \underbrace{ \exp \left[ - \frac{ \alpha}{ |\bm{X}^{(1)} | } \frac{1}{2\sigma^2}  (\bm{Y}^{(1)} - \bm{Z}^{(1)} \bm{\theta})^\top (\bm{Y}^{(1)} - \bm{Z}^{(1)} \bm{\theta}) \right]}_{\text{tempered-likelihood}} \underbrace{ \exp \left[- \frac{ (\bm{\theta}-\bm{\mu}_0)^\top \bm{S}_0^{-1} (\bm{\theta}-\bm{\mu}_0) }{2} \right] }_{\text{prior}} \\
    &\propto \exp \left[- \frac{1}{2} \left( \bm{\theta} - \bm{\mu}_P \right)^\top \bm{S}_P^{-1} \left( \bm{\theta} - \bm{\mu}_P \right)    \right]
\end{align*}
where we have completed the squares.\\
$ $\\
The $\alpha$-posterior can be rewritten as a Gaussian distribution as follows,
\begin{align*}
    \pi_{\alpha}^{(1)} ( d \bm{\theta}) \sim \mathcal{N} \left( \bm{\mu}_P^{(1)}, \bm{S}_P^{(1)} \right),
\end{align*}
where 
\begin{align*}
     \bm{S}_P^{(1)} &= \left( \frac{\alpha}{\sigma^2 |\bm{X}^{(1)}|}  \bm{Z}^{(1) \top} \bm{Z}^{(1)} + \bm{S}_0^{-1} \right)^{-1} \\
     \bm{\mu}_P^{(1)} &= \bm{S}_P^{(1)} \left( \frac{ \alpha}{\sigma^2 |\bm{X}^{(1)}|} \bm{Z}^{(1) \top} \bm{Y}^{(1)} + \bm{S}_0^{-1} \bm{\mu}_0 \right).
\end{align*}
Note that $\bm{\mu}_P^{(1)}$ has dimension $d \times 1$ and $\bm{S}_P^{(1)}$ has dimension $d \times d$. \\

It is interesting to check the values of the $\alpha$-posterior for limit values of $\alpha$. 
\begin{align*}
    \bm{\mu}_P^{(1)} &\rightarrow_0 \bm{\mu}_0 && \bm{\mu}_P^{(1)} \rightarrow_\infty (\bm{Z}^{(1) \top} \bm{Z}^{(1)})^{-1} \bm{Z}^{(1) \top } \bm{Y}^{(1)} \\
    \bm{S}_P^{(1)} &\rightarrow_0 \bm{S}_0 && \bm{S}_P^{(1)} \rightarrow_\infty \bm{0}
\end{align*}
We see that when $\alpha \rightarrow 0$, the $\alpha$-posterior simply becomes the prior. When $\alpha \rightarrow \infty$, the $\alpha$-posterior becomes the solution of the least squares.

\subsection{Derivation of the generalization error}
The generalization error for datasets $\bm{\bm{X}}^{(\lambda)}$ and $\bm{\bm{X}}^{(\nu)}$ is computed as
\begin{align*}
    \mathbb{E}_{\bm{\theta}\sim \pi_\alpha^{(\lambda)}}[r_n^{(\nu)}(\bm{\theta})] \nonumber &= \frac{1}{2 \sigma^2 |\bm{X}^{(\nu)}|}  \mathbb{E}_{\bm{\theta}\sim \pi_\alpha^{(\lambda)}} \left[   (\bm{Y}^{(\nu)} - \bm{Z}^{(\nu)} \bm{\theta})^\top (\bm{Y}^{(\nu)} - \bm{Z}^{(\nu)} \bm{\theta})  \right]\\
     &= \frac{1}{2 \sigma^2 |\bm{X}^{(\nu)}|}  \Bigg\{  \bm{Y}^{(\nu) \top} \bm{Y}^{(\nu)} - \bm{Y}^{(\nu) \top} \bm{Z}^{(\nu)} \mathbb{E}_{\bm{\theta}\sim \pi_\alpha^{(\lambda)}} \left[ \bm{\theta} \right] \\
     &\qquad - \mathbb{E}_{\bm{\theta}\sim \pi_\alpha^{(\lambda)}} \left[ \bm{\theta}^\top \right] \bm{Z}^{(\nu) \top}  \bm{Y}^{(\nu)} + \mathbb{E}_{\bm{\theta}\sim \pi_\alpha^{(\lambda)}} \left[ \bm{\theta}^\top \bm{Z}^{(\nu) \top} \bm{Z}^{(\nu)} \bm{\theta}  \right] \Bigg\} \\
     &\stackrel{(a)}{=} \frac{1}{2 \sigma^2 |\bm{X}^{(\nu)}|}  \Bigg\{  \bm{Y}^{(\nu) \top} \bm{Y}^{(\nu)} - 2 \bm{Y}^{(\nu) \top} \bm{Z}^{(\nu)} \bm{\nu}_P^{(\lambda)} \\
     &\qquad + \Tr \left( \bm{Z}^{(\nu) \top} \bm{Z}^{(\nu)} \bm{S}_P^{(\lambda)} \right) + \bm{\nu}_P^{{(\lambda)}\top} \bm{Z}^{(\nu) \top} \bm{Z}^{(\nu)} \bm{\nu}_P^{(\lambda)}   \Bigg\}
\end{align*}
where in (a) we have used the moment rules for Gaussian expectations \cite{IMM2012-03274}, using the mean and the variance of $\pi_{\alpha}^{(\lambda)}$ found in equation \eqref{linear_regression_posterior_gaussian}. \\

For the bootstrap strategy, we cannot minimize the noisy generalization error, hence we nust compute its gradient and run a SGD over it. To that end, we first compute the derivatives of the posterior parameters,
\begin{align*}
    \frac{\partial \bm{S}_P}{\partial \alpha} &\stackrel{(b)}{=}- \bm{S}_P \frac{\partial \bm{S}_P^{ -1}}{\partial \alpha} \bm{S}_P = - \frac{1}{ \sigma^2 |\bm{X}|} \bm{S}_P \bm{Z}^{ \top} \bm{Z} \bm{S}_P \\
    \frac{\partial \bm{\mu}_P}{\partial \alpha} &=  \frac{ 1}{\sigma^2 |\bm{X}|} \bm{S}_P \bm{Z}^{ \top} \left( \bm{Y} - \bm{Z} \bm{\mu}_P \right) 
\end{align*}
where in (b), the derivative of an inverse matrix has been used (see \cite{IMM2012-03274} eq. 59), and we finally obtain the derivative of the generalization error estimate
\begin{align*}
    \frac{\partial}{\partial \alpha} \mathcal{\hat{R}}( \alpha) &= \frac{1}{2 \sigma^2 |\bm{X}^{(\mu)}|} \left\{ - 2  \frac{\partial}{\partial \alpha} \bm{Y}^{(\mu) \top} \bm{Z}^{(\mu)}  \bm{\mu}_P^{(\lambda)} +  \frac{\partial}{\partial \alpha} \Tr \left( \bm{Z}^{(\mu) \top} \bm{Z}^{(\mu)} \bm{S}_P^{(\lambda)} \right) +  \frac{\partial}{\partial \alpha} \bm{\mu}_P^{(\lambda) \top} \bm{Z}^{(\mu) \top} \bm{Z}^{(\mu)} \bm{\mu}_P^{(\lambda)}   \right\} \\
    &= \frac{1}{2 \sigma^2 |\bm{X}^{(\mu)}|} \left\{ - 2  \bm{Y}^{(\mu) \top} \bm{Z}^{(\mu)}  \frac{\partial  \bm{\mu}_P^{(\lambda)}}{\partial \alpha} +  \Tr \left( \bm{Z}^{(\mu) \top} \bm{Z}^{(\mu)} \frac{\partial \bm{S}_P^{(\lambda)}}{\partial \alpha} \right) \stackrel{(c)}{+}  2  \bm{\mu}_P^{ (\lambda) \top} \bm{Z}^{(\mu) \top} \bm{Z}^{(\mu)}   \frac{ \partial \bm{\mu}_P^{(\lambda)} }{ \partial \alpha} \right\} .
\end{align*}

In (c), we have used the property $\frac{ \partial g(U)}{\partial x} = \Tr \left( \left( \frac{ \partial g(U)}{ \partial U} \right)^\top \frac{ \partial U}{ \partial x} \right) $ where $x$ is a scalar, $U$ is a matrix whose entries depend on $x$, and $g$ in a function taking a matrix and returning a scalar. Note that the trace operator can be removed as its inside is a scalar already.\\

Finally, we average together the above derivative for many bootstrap datasets, to be used as the gradient in a SGD algorithm. Replacing $\bm{X}^{(\lambda)}$ with each of the bootstrap dataset $\bm{X}^{(b)}$ in turn, and $\bm{X}^{(\mu)}$ with $\bm{X}$, we obtain 
\begin{align*}
    \Psi  = \frac{1}{boot} \sum_{b=1}^{boot}
    \frac{1}{2 \sigma^2 |\bm{X}^{(b)}|} \Bigg\{ &- 2  \bm{Y}^{(b) \top} \bm{Z}^{(b)}  \frac{\partial  \bm{\mu}_P}{\partial \alpha} +  \Tr \left( \bm{Z}^{(b) \top} \bm{Z}^{(b)} \frac{\partial \bm{S}_P}{\partial \alpha} \right) \\
    &+ 2  \bm{\mu}_P^{  \top} \bm{Z}^{(b) \top} \bm{Z}^{(b)}   \frac{ \partial \bm{\mu}_P }{ \partial \alpha} \Bigg\} .
\end{align*}

\subsection{Derivation of the SafeBayes error term}
We first compute the expected loss
\begin{align*}
\mathcal{E}(\alpha, t) &= \mathbb{E}_{\bm{\theta}\sim \pi_\alpha^{(t)} } \left[ \ell(\bm{\theta}, X_{t+1} ) \right] \\
&= \mathbb{E}_{\bm{\theta}\sim \pi_\alpha^{(t)} } \left[ \frac{(Y_{t+1} - Z_{t+1}^\top \bm{\theta})^2}{2 \sigma^2} \right] \\
&= \frac{1}{2 \sigma^2} \Big(   Y_{t+1}^2 + \Tr ( Z_{t+1} Z_{t+1}^\top \bm{S}_P^{(t)}) + \bm{\mu}_P^{ (t) \top} Z_{t+1} Z_{t+1}^\top \bm{\mu}_P^{(t)} - 2 Y_{t+1} Z_{t+1}^\top \bm{\mu}_P^{(t)} \Big)
\end{align*}
and second the PEPRL term
\begin{align*}
    \mathcal{S}(\alpha) &= \sum_{t=1}^{n-1} \mathcal{E}( \alpha, t) \\
    &= \frac{1}{2 \sigma^2} \sum_{t=1}^{n-1}  \Big(   Y_{t+1}^2 + \Tr ( Z_{t+1} Z_{t+1}^\top \bm{S}_P^{(t)}) + \bm{\mu}_P^{ (t) \top} Z_{t+1} Z_{t+1}^\top \bm{\mu}_P^{(t)} - 2 Y_{t+1} Z_{t+1}^\top \bm{\mu}_P^{(t)} \Big) .
\end{align*}
A simple minimization algorithm is then applied over this function.

\subsection{Gaussian mean estimation model setup}\label{gaussian_setup}
The Gaussian mean estimation model is a special case of the linear regression with known variance. In this model, one wants to estimate the scalar mean $\theta$ of a one-dimensional Gaussian distribution with known variance. The model is written as
\begin{align*}
     X_i \sim \mathcal{N} (\theta, 1).
\end{align*}
The likelihood is written as the mean-squares formula
\begin{align*}
    \mathcal{L}(\theta, \bm{X}) \propto \exp \left( \frac{ (\bm{X} - \theta)^\top (\bm{X} - \theta) }{2} \right)
\end{align*}
and the loss function is defined as the negative log likelihood
\begin{align*}
    \ell (\theta, X_i) = - \frac{1}{2} (X_i - \theta)^2 .
\end{align*}
One can rewrite this model as the linear regression with known variance 
\begin{align*}
    Y_i = Z_i \theta + \varepsilon_i, \qquad \varepsilon_i \sim \mathcal{N}(0,\sigma^2)
\end{align*}
by setting 
\begin{align}
    Y_i &=X_i, \forall i \nonumber \\
    Z_i &= 1 , \forall i  \nonumber \\ 
        d &=1 \nonumber \\
    \sigma^2 &=1 . \label{linreg_into_gaussian}
\end{align}

The prior is again chosen Gaussian $\pi (d \theta) \sim \mathcal{N} (\mu_0, s_0)$, where both $\mu_0$ and $s_0$ are scalars. By modifying the linear regression with known variance's $\alpha$-posterior \eqref{linear_regression_posterior_gaussian} with the edits proposed in \eqref{linreg_into_gaussian}, we rewrite the Gaussian $\alpha$-posterior
\begin{align}\label{gaussian_mean_posterior_gaussian}
    \pi_{\alpha}^{(1)} ( d \theta) \sim \mathcal{N} \left( \mu_P^{(1)}, s_P^{(1)} \right)
\end{align}
where the parameters $\Omega_P$ are
\begin{align*}
    s_P^{(1)} &= \frac{ s_0 }{ 1+ \alpha  s_0} \\
    \mu_P^{(1)} &=  s_P^{(1)} \left( \frac{ \mu_0 }{ s_0 } +  \frac{ \alpha }{ | X^{(1)} |}  \sum_{i=1}^{n/2} X_i \right) .
\end{align*}

\subsection{Generalization error for the Gaussian mean estimation}
We replace \eqref{linreg_into_gaussian} in the generalization error term of the linear regression with known variance, and obtain
\begin{align*}
    \mathbb{E}_{\theta\sim \pi_\alpha^{(\lambda)}} \left[ r_n^{(\nu)}(\theta) \right] =
  &\frac{\Vert \bm{X}^{(\nu)} \Vert^2_2}{2 | \bm{X}^{(\nu)}|}  -  \sum_{i \in (\nu)} \frac{X_i \mu_P^{(\lambda)}}{| \bm{X}^{(\nu)}|} + \frac{1}{2} s_P^{(\lambda)}  + \frac{1}{2} (\mu_P^{{(\lambda)}})^2 
\end{align*}
where $\Vert \cdot \Vert$ denotes the $L^2$ norm. All the following computations can be trivially obtained by applying the same replacements and are hence omitted.

\subsection{Polynomial regression model setup}\label{polynomial_setup}
Polynomial regression is also a special case of linear regression where a non-linear function is to be estimated with a polynomial of fixed degree. The data is generated from the non-linear function $f$ as 
$$ Y_i = f(\zeta_i) + \varepsilon_i$$
where $\varepsilon_i \sim \mathcal{N}(0, \sigma^2)$, $\sigma^2$ is a constant, and $f$ is a non-linear function, for instance $f(\cdot):= \exp(\cdot)$. The model then tries to fit the data using a design matrix $\bm{Z}$ of size $n \times d$ which is not arbitrary, but actually generated from a vector $\bm{\zeta}$ of size $n \times 1$, that we expand into a Vandermonde design matrix, using a polynomial basis
$$ \bm{\zeta} = \begin{bmatrix} \zeta_1 & \zeta_2 & \dots & \zeta_n \end{bmatrix}^\top \quad \longrightarrow \quad \bm{Z} = \begin{bmatrix} 1 & \zeta_1 & \zeta_1^2 & \dots & \zeta_1^{d-1} \\ 1 &  \zeta_2 & \zeta_2^2 & \dots & \zeta_2^{d-1} \\  1 & \zeta_3 & \zeta_3^2 & \dots & \zeta_3^{d-1} \\   \vdots & \vdots & \vdots & \ddots & \vdots \\ 1 & \zeta_n & \zeta_n^2 & \dots & \zeta_n^{d-1} \end{bmatrix} , $$
hence reducing the input data size to only $n \times 1$ observations instead of $n \times d$ in the general linear regression. The linear regression model can be rewritten as
$$Y_i =   Z_i^\top \bm{\theta} + \varepsilon_i = \sum_{k=0}^{d-1} \bm{\theta}_k \zeta_i^k + \varepsilon_i . $$

The prior is chosen to be 
$$\pi( d \bm{\theta}_k) \sim \mathcal{N} \left( 0, 1/2^k \right), k \in (0, \dots, d-1), $$
so that the low powers of the polynomial are given larger weights and high powers small weights and hence overfitting is limited. The likelihood is computed the same way as in the linear regression with known variance model.\\

In order to predict a new observation $y$ of the noisy function $f(\cdot)$, we compute the $\alpha$-posterior predictive \cite{bishop_book} as
\begin{align*}
    p(y | \bm{Z}, \bm{Y}, \sigma^2) = \mathcal{N} \left( \bm{Z} \bm{\mu}_P, \sigma^2 + \bm{Z} \bm{S}_P \bm{Z}^\top \right).
\end{align*}
We remark that when $\alpha$ approaches infinity, then $\sigma^2 + \bm{Z} \bm{S}_P \bm{Z}^\top \rightarrow_\infty \sigma^2$, the variance of the predictions tends to the variance of the noise of the observations as the posterior's variance goes to 0. The same phenomenon occurs when $n$ grows to infinity \cite{Qazaz1997}. 

\newpage
\thispagestyle{empty} 
\mbox{}

\section{Derivations for the linear regression with unknown variance}\label{appendix_linreg2}
These computations are done for the sample splitting strategy, using the according data batches. When using the other strategies, one should replace the batches accordingly.

\subsection{Derivation of the $\alpha$-posterior \eqref{NIG-posterior}}

The empirical error is computed as
\begin{align*}
 r_n^{(1)}(\bm{\theta}, \sigma^2) & = \frac{1}{|\bm{X}^{(1)}|} \sum_{i=1}^{n/2} \ell (\bm{\theta}, \sigma^2, X_i)\\
    &= \frac{1}{2 \sigma^2 |\bm{X}^{(1)}|} \sum_{i=1}^{n/2} (Y_i - Z_i^\top \bm{\theta} )^2 + \frac{1}{2} \log 2 \pi \sigma^2  \\
 r_n^{(2)}(\bm{\theta}, \sigma^2) &= \frac{1}{2 \sigma^2 |\bm{X}^{(2)}|} \sum_{i=1+n/2}^{n} (Y_i - Z_i^\top \bm{\theta} )^2 + \frac{1}{2} \log 2 \pi \sigma^2  .
\end{align*}
The empirical error formulae are very similar to those in the known variance case, except that now $\sigma^2$ is a parameter as well, and not a constant anymore.\\

The $\alpha$-posterior is now a joint distribution of 2 random variables and can be computed in closed-form using a similar derivation as for the known variance case, except that now the joint conjugate posterior is a Normal-Inverse-Gamma distribution as well.

\begin{align*}
    \pi_{\alpha}^{(1)}(d \bm{\theta}, d \sigma^2) &\propto \exp\left[-\alpha r_n^{(1)}(\bm{\theta}, \sigma^2) \right] \times \pi({ d}\bm{\theta}) \\
    &\propto \underbrace{ \left( \frac{1}{\sigma^2} \right)^{ \alpha/2 } \exp \left\{ - \frac{\alpha}{2 \sigma^2  |\bm{X}^{(1)}| } (\bm{Y} - \bm{Z} \bm{\theta})^\top (\bm{Y} - \bm{Z} \bm{\theta}) \right\} }_{\text{tempered likelihood}} \\
    &\qquad \qquad \times \underbrace{ \left( \frac{1}{\sigma^2} \right)^{ a_0 + d/2 + 1} \exp \left\{ - \frac{1}{\sigma^2} \left[ b_0 + \frac{1}{2} (\bm{\theta} - \bm{\mu}_0)^\top \bm{S}_0^{-1} (\bm{\theta} - \bm{\mu}_0) \right] \right\} }_{\text{NIG prior}}\\
    &\propto \left( \frac{1}{\sigma^2} \right)^{a_0 + \alpha/2  + d/2 + 1 } 
    \times \exp \bigg\{ - \frac{1}{\sigma^2} \Big[ b_0 + \frac{1}{2} (\bm{\theta} - \bm{\mu}_0)^\top  \bm{S}_0^{-1} (\bm{\theta}-\bm{\mu}_0) \\
    &\qquad \qquad + \frac{1}{2} \frac{\alpha}{| \bm{X}^{(1)} |} (\bm{Y}- \bm{Z} \bm{\theta})^\top (\bm{Y} - \bm{Z} \bm{\theta}) \Big] \bigg\} \\
    &\propto \left( \frac{1}{\sigma^2} \right)^{ a_P + d/2 + 1 } \times \exp \left\{ -\frac{1}{\sigma^2} \left[    b_P + \frac{1}{2}( \bm{\theta}- \bm{\mu}_P)^\top \bm{S}_P^{-1} (\bm{\theta} - \bm{\mu}_P) \right] \right\}\\
    &\sim \text{NIG} ( \bm{\mu}_P^{(1)}, \bm{S}_P^{(1)}, a_P^{(1)}, b_P^{(1)})
\end{align*}
where 
\begin{align*}
     \bm{\mu}_P^{(1)} &= \bm{S}_P \left( \bm{S}_0^{-1} \bm{\mu}_0 + \frac{\alpha}{| \bm{X}^{(1)} |} \bm{Z}^{(1) \top} \bm{Y}^{(1)} \right) \\
     \bm{S}_P^{(1)} &= \left( \frac{\alpha}{| \bm{X}^{(1)} |} \bm{Z}^{(1) \top} \bm{Z}^{(1)} + \bm{S}_0^{-1} \right)^{-1} \\
     a_P^{(1)} &= a_0 + \frac{ \alpha }{2} \\
     b_P^{(1)} &= b_0 + \frac{1}{2} \left( \bm{\mu}_0^\top \bm{S}_0^{-1} \bm{\mu}_0 - \bm{\mu}_P^{(1) \top} \bm{S}_P^{(1) -1} \bm{\mu}_P^{(1)} + \frac{\alpha}{| \bm{X}^{(1)} |} \bm{Y}^{(1) \top} \bm{Y}^{(1)} \right).
\end{align*}
The mean vector $\bm{\mu}_P^{(1)}$ has dimensions $d \times 1$, covariance matrix $\bm{S}_P^{(1)}$ has dimensions $d \times d$, and Gamma parameters $a_P^{(1)}$ and $b_P^{(1)}$ both are scalars.\\ 

We can compute the limit values of the $\alpha$-posterior parameters. When $\alpha \rightarrow 0$, only the prior influences the joint posterior, hence the posterior hyperparameters should tend towards the prior hyperparameters. Conversely, when $\alpha \rightarrow \infty$, only the likelihood influences the posterior, and the posterior should be a least-squares MLE. We can verify this by observing that
\begin{align*}
    \bm{\mu}_P^{(1)} &\rightarrow_0 \bm{\mu}_0 &&  \bm{\mu}_P^{(1)} \rightarrow_\infty (\bm{Z}^{(1) \top} \bm{Z}^{(1)})^{-1} \bm{Z}^{(1) \top } \bm{Y}^{(1)} \\
    \bm{S}_P^{(1)} &\rightarrow_0 \bm{S}_0 && \bm{S}_P^{(1)} \rightarrow_\infty 0 \\
    a_P^{(1)} &\rightarrow_0 a_0 && a_P^{(1)} \rightarrow_\infty ( \infty ) \\
    b_P^{(1)} &\rightarrow_0 b_0 && b_P^{(1)} \rightarrow_\infty ( \infty ) .
\end{align*}
Note that the two last infinite terms on the right are not an issue as they do no contribute to the distribution since $\bm{S}_P$ becomes $0$ and cancels the influence of the Gamma distribution.

\subsection{Derivation of the generalization error}
The generalization error is here described when trained on a batch $\bm{X}^{(\lambda)}$ and tested on a batch $\bm{X}^{(\nu)}$. For the following, it is going to be useful to list the following values:
{\small
\begin{align*}
    \mathbb{E}_{ \sigma^2 \sim \pi_\alpha^{(\lambda)}} \left[ \frac{1}{\sigma^2} \right] &= \frac{a_P^{(\lambda)}}{b_P^{(\lambda)}} \\
    \mathbb{E}_{ \sigma^2 \sim \pi_\alpha^{(\lambda)}} \left[ \log \sigma^2 \right] &= \log (b_P^{(\lambda)}) - \psi(a_P^{(\lambda)}) \\
    \mathbb{E}_{(\bm{\theta}, \sigma^2) \sim \pi_\alpha^{(\lambda)}} \left[ \frac{\bm{\theta}}{\sigma^2} \right] &\stackrel{(a)}{=} \int \frac{1}{ |\bm{S}_P^{(\lambda)}|^{1/2} (2 \pi )^{d/2} } \frac{(b_P^{(\lambda)})^{a_P^{(\lambda)}}}{\Gamma(a_P^{(\lambda)})} \left( \frac{1}{\sigma^2} \right)^{a_P^{(\lambda)} + d/2 + 1} \\
    &\cdot \exp{ \Bigg\{ -\frac{b_P^{(\lambda)}}{\sigma^2} \int \bm{u} \exp{ \Bigg\{ - \frac{1}{2} \left( \bm{u} - \frac{ \bm{\mu}_P^{(\lambda)}}{\sigma} \right)^\top (\bm{S}_P^{(\lambda)})^{-1} \left( \bm{u} - \frac{ \bm{\mu}_P^{(\lambda)}}{\sigma} \right)  \Bigg\} } d \bm{u} \Bigg\}}  d \sigma^2    \\
    &= \bm{\mu}_P^{(\lambda)} \frac{(b_P^{(\lambda)})^{a_P^{(\lambda)}}}{\Gamma(a_P^{(\lambda)})} \frac{ \Gamma \left(a_P^{(\lambda)} + (d-1)/2 \right) }{ (b_P^{(\lambda)})^{a_P^{(\lambda)} + (d-1)/2} } = \frac{\bm{\mu}_P^{(\lambda)} }{\Gamma(a_P^{(\lambda)})} \frac{ \Gamma \left(a_P^{(\lambda)} + (d-1)/2 \right) }{ (b_P^{(\lambda)})^{(d-1)/2} } \\
     \mathbb{E}_{(\bm{\theta}, \sigma^2) \sim \pi_\alpha^{(\lambda)}} \left[ \frac{1}{\sigma^2} \bm{\theta}^\top \bm{Z}^{ (\nu) \top} \bm{Z}^{(\nu)} \bm{\theta}  \right] &\stackrel{(a)}{=} \int \frac{1}{ |\bm{S}_P^{(\lambda)}|^{1/2} (2 \pi )^{d/2} } \frac{(b_P^{(\lambda)})^{a_P^{(\lambda)}}}{\Gamma(a_P^{(\lambda)})} \left( \frac{1}{\sigma^2} \right)^{a_P^{(\lambda)} + d/2 + 1/2} \cdot \exp \Bigg\{ -\frac{b_P^{(\lambda)}}{\sigma^2}  \\
    &\cdot \int \bm{u}^\top \bm{Z}^{(\nu)} \bm{Z}^{ (\nu) \top} \bm{u} \exp \Bigg\{ - \frac{1}{2} \left( \bm{u} - \frac{ \bm{\mu}_P^{(\lambda)}}{\sigma} \right)^\top (\bm{S}_P^{(\lambda)})^{-1} \left( \bm{u} - \frac{ \bm{\mu}_P^{(\lambda)}}{\sigma} \right)  \Bigg\}  d \bm{u} \Bigg\}   d \sigma^2    \\
    &= \int \frac{1}{ |\bm{S}_P^{(\lambda)}|^{1/2} (2 \pi )^{d/2} } \frac{(b_P^{(\lambda)})^{a_P^{(\lambda)}}}{\Gamma(a_P^{(\lambda)})} \left( \frac{1}{\sigma^2} \right)^{a_P^{(\lambda)} + d/2 + 1/2} \cdot \exp \Bigg\{ -\frac{b_P^{(\lambda)}}{\sigma^2} |\bm{S}_P^{(\lambda)}|^{1/2} (2 \pi)^{d/2} \\
    &\cdot \left( \Tr \left(\bm{Z}^{(\nu)}  \bm{S}_P^{(\lambda)} \bm{Z}^{ {(\nu)} \top} + \frac{1}{\sigma^2} \bm{\mu}_P^{^{(\lambda)} \top} \bm{Z}^{(\nu) \top} \right) \bm{Z}^{(\nu)}  \bm{\mu}_P^{(\lambda)} \right) \Bigg\}   d \sigma^2    \\
    &= \frac{(b_P^{(\lambda)})^{a_P^{(\lambda)}} }{\Gamma(a_P^{(\lambda)})} \Bigg( \frac{ \Gamma( a_P^{(\lambda)} + d/2 - 1/2) }{ (b_P^{(\lambda)})^{a_P^{(\lambda)} + d/2 - 1/2} } \Tr( \bm{Z}^{(\nu)}  \bm{S}_P^{(\lambda)} \bm{Z}^{ (\nu) \top} ) \\
    &+ \frac{ \Gamma( a_P^{(\lambda)} + d/2 -3/2 ) }{ (b_P^{(\lambda)})^{a_P^{(\lambda)} + d/2 -3/2 }} \bm{\mu}_P^{ ^{(\lambda)}\top} \bm{Z}^{(\nu)  \top} \bm{Z}^{(\nu)}  \bm{\mu}_P^{(\lambda)} \Bigg)
\end{align*}
}%
where $\psi(\cdot)$ is the digamma function, and where in (a), the change of variable $\bm{u}=\frac{\bm{\theta}}{\sigma^2}$ was applied.\\

We now compute 
\begin{align*}
\mathcal{\hat{R}}(\alpha) &= \mathbb{E}_{(\bm{\theta}, \sigma^2) \sim \pi_\alpha^{(\lambda)}}[r_n(\bm{\theta}, \sigma^2)]\\
 &\propto \mathbb{E}_{(\bm{\theta}, \sigma^2) \sim \pi_\alpha^{(\lambda)}} \left[ \frac{1}{2 \sigma^2 |\bm{X}^{ (\nu) }|}   (\bm{Y}^{ (\nu) } - \bm{Z}^{ (\nu) } \bm{\theta})^\top (\bm{Y}^{ (\nu) } - \bm{Z}^{ (\nu) } \bm{\theta}) + \frac{1}{2} \log \sigma^2  \right]\\
 &= \frac{1}{2 |\bm{X}^{ (\nu) }|}  \Bigg\{ \mathbb{E}_{\sigma^2 \sim \pi_\alpha^{(\lambda)}} \left[ \frac{1}{\sigma^2} \right] \bm{Y}^{(\nu) \top} \bm{Y}^{ (\nu) } - \bm{Y}^{(\nu) \top} \bm{Z}^{ (\nu) } \cdot \mathbb{E}_{(\bm{\theta}, \sigma^2) \sim \pi_\alpha^{(\lambda)}} \left[ \frac{\bm{\theta}}{\sigma^2} \right]  \\
 &- \mathbb{E}_{(\bm{\theta}, \sigma^2) \sim \pi_\alpha^{(\lambda)}} \left[ \frac{\bm{\theta}^\top}{\sigma^2} \right] \bm{Z}^{ (\nu) \top}  \bm{Y}^{ (\nu) } + \mathbb{E}_{(\bm{\theta}, \sigma^2) \sim \pi_\alpha^{(\lambda)}} \left[ \frac{1}{\sigma^2} \bm{\theta}^\top \bm{Z}^{ (\nu) \top} \bm{Z}^{ (\nu) } \bm{\theta}  \right] \Bigg\} + \frac{1}{2} \mathbb{E}_{\sigma^2 \sim \pi_\alpha^{(\lambda)}} \left[ \log \sigma^2 \right]
\end{align*}
and subsequently plug in the previously computed values.\\

For the bootstrap strategy, the derivative of $\mathcal{\hat{R}}$ is needed for optimization. Although the derivative is theoretically tractable, the computation becomes very heavy and costly. Furthermore, popular gradient computers such as Autograd do not include complex distributions like NIG. It hence becomes easier and computationally faster to simply estimated it with MC and to run an SGD algorithm using it.

\subsection{Derivation of the SafeBayes error term}
First, the local error term is
\begin{align*}
\mathcal{E}(\alpha, t) &= \mathbb{E}_{(\bm{\theta}, \sigma^2) \sim \pi_\alpha^{(t)} } \left[ \ell(\bm{\theta}, \sigma^2, X_{t+1} ) \right] \\
&\propto \mathbb{E}_{(\bm{\theta}, \sigma^2) \sim \pi_\alpha^{(t)} } \left[ \frac{(Y_{t+1} - Z_{t+1}^\top \bm{\theta})^2}{2 \sigma^2} + \frac{|X_{t+1}|}{2} \log \sigma^2 \right] \\
&= \frac{a_P^{(1)}}{2 b_P^{(1)}} \Big(   Y_{t+1}^2 + \Tr ( Z_{t+1} Z_{t+1}^\top \bm{S}_P^{(t)}) + \bm{\mu}_P^{ (t) \top} Z_{t+1} Z_{t+1}^\top \bm{\mu}_P^{(t)} - 2 Y_{t+1} Z_{t+1}^\top \bm{\mu}_P^{(t)} \Big) \\
&\qquad \qquad + \frac{1}{2} \left( \log (b_P^{(t)}) - \psi(a_P^{(t)}) \right)
\end{align*}
and second, the global error term is
\begin{align*}
    \mathcal{S}(\alpha) &= \sum_{t=1}^{n-1} \mathcal{E}( \alpha, t) .
\end{align*}

\section{Derivations for the logistic regression}

\subsection{Derivation of the Jaakkola variational $\alpha$-posterior \eqref{jaakkola_posterior}}\label{jaakkola-appendix}
In the Jaakkola setup, we want to obtain closed-form expressions for the $\alpha$-posterior parameters, as well as a variational lower bound $v$. Both the parameters and $v$ depend on each other, and can be updated in turn. The computations are here done for the sample splitting strategy, and the data batches must be replaced accordingly when another strategy is used. We compute the the empirical error functions
\begin{align*}
 r_n^{(1)}(\bm{\theta}) & = \frac{1}{|\bm{X}^{(1)}|} \sum_{i=1}^{n/2}  - \bm{\theta}^\top Z_i Y_i - \log  \left( \sigma \left(- \bm{\theta}^\top Z_i \right) \right) , \\
 r_n^{(2)}(\bm{\theta}) & = \frac{1}{|\bm{X}^{(2)}|} \sum_{i=1+n/2}^{n}  - \bm{\theta}^\top Z_i Y_i - \log  \left( \sigma \left(- \bm{\theta}^\top Z_i \right) \right).
\end{align*}
In the following, we rewrite $\alpha' := \alpha / |\bm{X}^{(1)}|$ for clarity of notation. The $\alpha$-posterior is then written as
\begin{align}
\pi_{\alpha}^{(1)}  ( d \bm{\theta}) &= \exp\left[-\alpha r_n^{(1)}(\bm{\theta}) \right] \pi({ d}\bm{\theta}) \nonumber \\ 
 &= \underbrace{ \prod_{i=1}^{n/2} \left(e^{ \bm{\theta}^\top Z_i Y_i } \sigma ( -\bm{\theta}^\top Z_i ) \right)^{\alpha'} }_{\text{tempered likelihood}} \cdot \underbrace{ \pi( d \bm{\theta}) }_{\text{prior}} \nonumber \\
  &= \prod_{i=1}^{n/2} e^{ \alpha'  \bm{\theta}^\top Z_i Y_i } \cdot \sigma ( -\bm{\theta}^\top Z_i )^{\alpha' }  \cdot \pi( d \bm{\theta}) \label{logistic_intractable}
\end{align}

The exact posterior of logistic regression is intractable, as the marginal likelihood is too complex to be computed. Instead, using the variational Bayes method from \cite{jaakkola2001}, the posterior can be estimated as a Gaussian (see \cite{bishop_book} p.514). We first remind the variational lower bound for the sigmoid function: for scalars $u$ and $v$, 
$$ \sigma(u) \geq \sigma(v) \exp \left\{ (u - v)/2 - \lambda(v) (u^2 - v^2)  \right\}$$
where 
$$ \lambda(v) = \frac{1}{2 v} \left[ \sigma(v) - \frac{1}{2} \right] . $$
We next remark that as the sigmoid function and the exponential function are non-negative, and the exponentiation function $f(w) = w^\alpha$ is increasing for non-negative values of $\alpha$, we can rewrite
\begin{align}\label{sigmoid_lower_bound}
    \sigma(u)^\alpha \geq \sigma(v)^\alpha \exp \left\{ \alpha (u - v)/2 - \alpha \lambda(v) (u^2 - v^2)  \right\} . \end{align}
Replacing the sigmoid function in the formula \eqref{logistic_intractable} with its lower bound from \eqref{sigmoid_lower_bound} then leads to
$$ \pi_\alpha^{(1)} \geq \prod_{i=1}^{n/2} \sigma(v_i)^{\alpha'} \exp \left\{ \alpha' \bm{\theta}^\top Z_i Y_i - \alpha' (\bm{\theta}^\top Z_i + v_i) / 2 - \alpha' \lambda(v_i) ([ \bm{\theta}^\top Z_i]^2 - v_i^2) \right\}  \cdot \pi( d \bm{\theta}) . $$ 
One can then take the log of this expression while keeping only the terms depending on $\bm{\theta}$, giving
$$ \log  \pi_\alpha^{(1)} \geq -\frac{1}{2}(\bm{\theta} - \mu_0)^\top S_0^{-1} (\bm{\theta} - \mu_0) + \alpha' \sum_{i=1}^{n/2} \left\{  \bm{\theta}^\top Z_i (Y_i - 1/2 ) - \lambda(v_i) \bm{\theta}^\top (Z_i Z_i^\top) \bm{\theta}  \right\} + \text{cst} , $$
and subsequently complete the squares to obtain a Gaussian variational posterior:
$$ \pi_\alpha^{(1)} (d \bm{\theta}) \sim \mathcal{N}( \bm{\theta} | \bm{\mu}_P^{(1)}, \bm{S}_P^{(1)} ) $$
where
\begin{align*}
    \bm{\mu}_P^{(1)} &= \bm{S}_P^{(1)} \left( \bm{S}_0^{-1} \bm{\mu}_0 + \alpha' \sum_{i=1}^{n/2} \left( Y_i - \frac{1}{2} \right) Z_i \right) \\
    \bm{S}_P^{(1)} &= \left( \bm{S}_0^{-1} + 2 \alpha' \sum_{i=1}^{n/2} \lambda(v_i) Z_i Z_i^\top \right)^{-1}
\end{align*}
and the vector $v$ is still to be computed.\\

A closed-form solution exists for $v$, and it depends on the Gaussian posterior's parameters:
$$ v_i = \left( Z_i^\top (\bm{S}_P^{(1)} + \bm{\mu}_P^{(1)} \bm{\mu}_P^{(1) \top}) Z_i \right)^{1/2},$$
or in vector form
$$ v = \left( \text{diag} \left(\bm{Z}^{(1)} (\bm{S}_P^{(1)} + \bm{\mu}_P^{(1)} \bm{\mu}_P^{(1) \top)} \bm{Z}^{(1)\top} \right) \right)^{1/2} . $$
We now want to maximize the variational posterior. As the parameter $v$ and the posterior's parameters all depend on each other, an expectation-maximization (EM) algorithm can be used as follows to compute $v$ and $\bm{\mu}_P, \bm{S}_P$ iteratively. We first choose some arbitrary values for $v$ (initialization step), then compute the posterior distribution using this $v$ (expectation step), then compute a new value for $v$ using the new values of the parameters (maximization step), and so on. Usually, several ($<10$) iterations are sufficient to reach a good enough approximation.\\

One can try to check limit values for $\alpha$:
\begin{align*}
    \bm{\mu}_P^{(1)} &\rightarrow_0 \bm{\mu}_0 &&  \bm{\mu}_P^{(1)} \rightarrow_\infty \left( 2 \sum_{i=1}^{n/2} \lambda(v_i) Z_i Z_i^\top \right)^{-1} \sum_{i=1}^{n/2} (Y_i - 1/2) Z_i  \\
    \bm{S}_P^{(1)} &\rightarrow_0  \bm{S}_0 && \bm{S}_P^{(1)} \rightarrow_\infty 0 .
\end{align*}
The loss and empirical error contain the sigmoid function and it is intractable to compute the expected value terms. Thus, all strategies are approximated with MC and minimized with a SGD algorithm.

\subsection{Derivation of the Bayes by Backprop variational $\alpha$-posterior \eqref{SVI_posterior}}\label{SVI-appendix}
In the Bayes by Backprop setup, we need to derive update equations for the $\alpha$-posterior parameters. We start from the minimization view of the $\alpha$-posterior from \eqref{bayes_kl} and rewrite it as
\begin{align*}
    \pi_\alpha( d \bm{\theta}) &\propto \argmin_{\rho \in \mathcal{S}(\Theta)} \Big\{  \alpha \cdot \mathbb{E}_{\bm{\theta} \sim \rho} \left[ r_n (\bm{\theta} ) \right] + \mathcal{KL}(\rho || \pi_0) \Big\} \\
    &= \argmin_{\rho \in \mathcal{S}(\Theta)} \left\{ \mathbb{E}_{\bm{\theta} \sim \rho} \left[ \alpha  r_n (\bm{\theta} ) \right] + \mathbb{E}_{\bm{\theta} \sim \rho} \left[ \log \frac{\rho}{\pi_0} \right] \right\} \\
    &= \argmin_{\rho \in \mathcal{S}(\Theta)} \left\{ \underbrace{ \mathbb{E}_{\bm{\theta} \sim \rho} \left[ \log \rho - \log \pi_0 + \alpha  r_n (\bm{\theta} )  \right] }_{\text{negative ELBO}} \right\}  \\
    &:= \argmin_{\rho \in \mathcal{S}(\Theta)} \mathbb{E}_{\bm{\theta} \sim \rho} \left[ f(\bm{\theta}, \Omega_P) \right]   .
\end{align*}
We define the function $f$ as the term inside the expectation in the negative ELBO. We now want to minimize the negative ELBO using a gradient descent algorithm. To that end, we need to compute the derivative of the negative ELBO with respect to the posterior's parameters. We use a mean-field variational approximation, where only the diagonal elements of the covariance matrix are used. For parameters $\bm{\mu}_P$, the vector of means of the posterior, and $\bm{\sigma}_P$, the vector of standard deviations of the diagonal of the covariance matrix $\bm{S}_P$, \cite{deepmind2015weight} propose the reparametrization 
\begin{align*}
    \bm{\hat{\theta}} := \bm{\mu}_P + \log ( 1 + \exp (\bm{\rho}_P)) \cdot \bm{\varepsilon}
\end{align*}
where $\bm{\rho}_P = \log (\exp(\bm{\sigma}_P) -1)$ and $\bm{\varepsilon} \sim \mathcal{N}(0, \bm{I}_d)$. Note that the product is done elementwise. One can then compute the function $f$ in explicit form
\begin{align*}
    f(\bm{\theta}, \Omega_P) &= \log \rho - \log \pi_0 + \alpha  r_n (\bm{\theta} )  \\
    &\propto  -\frac{1}{2} \log | \bm{S}_P | - \frac{1}{2} (\bm{\theta} - \bm{\mu}_P)^\top \bm{S}_P^{-1} (\bm{\theta} - \bm{\mu}_P) -\frac{1}{2} \bm{\theta}^\top \bm{\theta}  + \alpha r_n (\bm{\theta}) .
\end{align*}
We then use the following gradient derivative trick to compute the gradient of the negative ELBO
\begin{align*}
    - \nabla_{\Omega_P} \text{ELBO} &= \nabla_{\Omega_P} \mathbb{E}_{\bm{\hat{\theta}} \sim \rho} \left[ f(\bm{\hat{\theta}}, \Omega_P) \right] \\
    &= \mathbb{E}_{\bm{\varepsilon} \sim \mathcal{N}(0,1) } \left[ \nabla_{\bm{\hat{\theta}}} f(\bm{\hat{\theta}}, \Omega_P) \cdot \nabla_{\Omega_P} \bm{\hat{\theta}}  + \nabla_{\Omega_P} f(\bm{\hat{\theta}}, \Omega_P) \right] \\
    &\stackrel{\text{MC}}{\approx} \frac{1}{mc} \sum_{i=1}^{mc} \left[ \nabla_{\bm{\hat{\theta}}_i} f(\bm{\hat{\theta}}_i, \Omega_P) \cdot \nabla_{\Omega_P} \bm{\hat{\theta}}_i  + \nabla_{\Omega_P} f(\bm{\hat{\theta}}_i, \Omega_P) \right], \qquad \varepsilon_i \sim \mathcal{N}(0,1) 
\end{align*}
where $mc$ can be chosen to be equal to 1. In that case, the gradients with respect to each parameter translate to first sampling $\bm{\varepsilon} \sim \mathcal{N}(0,1) $ and then computing
\begin{align*}
    \nabla_{\bm{\mu}_P} f(\theta, \Omega_P) &= \nabla_{\bm{\hat{\theta}}} f(\bm{\hat{\theta}}, \Omega_P)  + \nabla_{\bm{\mu}_P} f(\bm{\hat{\theta}}, \Omega_P)  \\
    \nabla_{\bm{\rho}_P} f(\theta, \Omega_P) &= \nabla_{\bm{\hat{\theta}}} f(\bm{\hat{\theta}}, \Omega_P) \cdot \frac{\bm{\varepsilon}}{1 + \exp(- \bm{\rho}_P)} + \nabla_{\bm{\rho}_P} f(\bm{\hat{\theta}}, \Omega_P)  
\end{align*}
using Autograd \cite{autograd2015}. We finally update the posterior values using a SGD with learning rate $\lambda$ by alternatively computing the gradients and updating the parameters
\begin{align*}
    \bm{\mu}_P &\leftarrow \lambda \cdot  \nabla_{\bm{\mu}_P} \\
    \bm{\rho}_P &\leftarrow \lambda \cdot \nabla_{\bm{\rho}_P},
\end{align*}
and last reparametrize $\bm{\rho}_P$ back into $\bm{\sigma}_P$. We plug the parameters into a Gaussian distribution to obtain
\begin{align*}
    \pi_\alpha (d \bm{\theta}) \sim \mathcal{N}( \bm{\theta} | \bm{\mu}_P, \text{diag} ( \bm{\sigma}^2_P) ).
\end{align*}

\begin{figure}[ht]
\centering
\begin{minipage}{.5\textwidth}
  \centering
    \includegraphics[width=\linewidth]{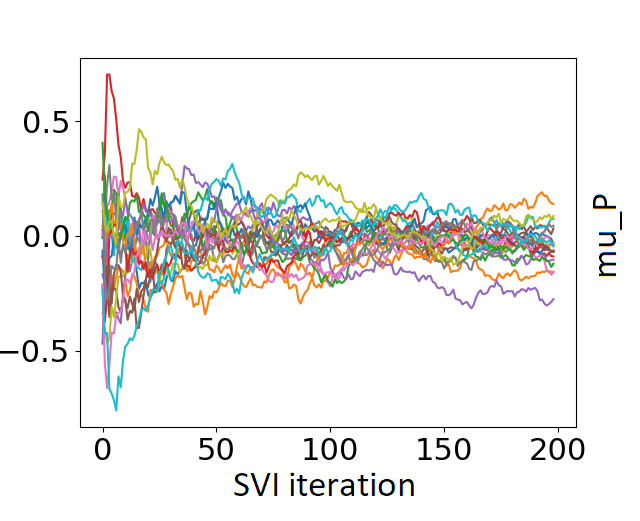}
\end{minipage}%
\begin{minipage}{.5\textwidth}
  \centering
  \includegraphics[width=\linewidth]{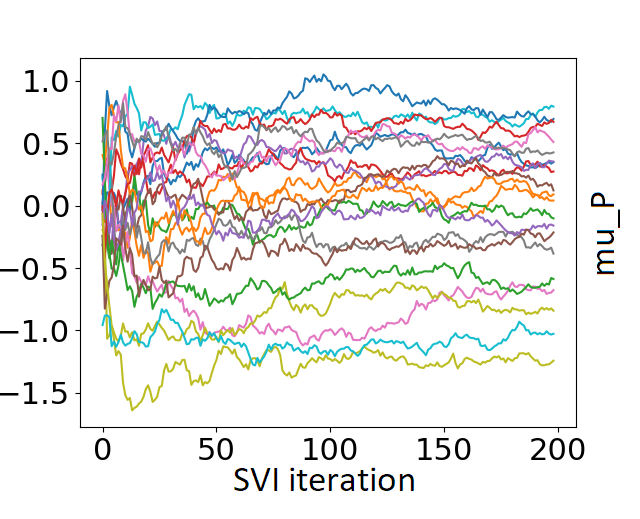}
\end{minipage}
    \caption{Bayes by Backprop convergence of the mean parameter $\bm{\mu}_P$'s values for fixed $\alpha/n=0$ on the left, and $\alpha/n=1$ on the right, when $n=100$ and $d=20$.}
    \label{fig:alpha100}
\end{figure}

In practice, the number of SGD iterations is typically set to 200 with inverse square root learning rate. One can observe the difference in the convergence of the $\alpha$-posterior's mean parameter $\bm{\mu}_P$ for different values of $\alpha$ in figure \ref{fig:alpha100}. Although no general closed-form expression of the parameters is available for limit values of $\alpha$, we can numerically remark the same phenomenon as for the other models. The value of $\alpha/n=0$ corresponds to giving all the weight to the prior $\pi_0 \sim \mathcal{N}(0, \bm{I}_d)$ and hence the mean parameter converges to an all-zero vector. As $\alpha$ increases, the values of the mean vector spread more evenly. Similarly, the values of the covariance matrix should converge to zero for a very large value of $\alpha$.

\newpage
\thispagestyle{empty} 
\mbox{}